\documentclass[acmlarge]{acmart}
\makeatletter
\newcommand{\myconfshort}{\acmConference@shortname}
\newcommand{\myconffull}{\acmConference@name}
\newcommand{\myconfdate}{\acmConference@date}
\newcommand{\myconfloc}{\acmConference@venue}
\AtBeginDocument{
  \fancypagestyle{firstpagestyle}{
    \fancyhead{}%
    \fancyfoot[C]{}%
  }
  \fancyhf{}
  \fancyhead[LO]{\@headfootfont\shorttitle}%
  \fancyhead[RE]{\@headfootfont\@shortauthors}%
  \fancyhead[LE]{\@headfootfont\footnotesize \myconfshort, \myconfdate, \myconfloc}%
  \fancyhead[RO]{\@headfootfont\footnotesize \myconfshort, \myconfdate, \myconfloc}%
  \fancyfoot[C]{}%
}
\makeatother
\acmBooktitle{\conffull\@ (\confshort), \confdate, \confloc}

\AtBeginDocument{%
  }

\copyrightyear{2026}
\acmYear{2026}
\setcopyright{cc}
\setcctype{by}
\acmConference[FAccT '26]{The 2026 ACM Conference on Fairness, Accountability, and Transparency}{June 25--28, 2026}{Montreal, QC, Canada}
\acmBooktitle{The 2026 ACM Conference on Fairness, Accountability, and Transparency (FAccT '26), June 25--28, 2026, Montreal, QC, Canada}
\acmDOI{10.1145/3805689.3812291}
\acmISBN{979-8-4007-2596-8/2026/06}


\usepackage[utf8]{inputenc} 
\usepackage[T1]{fontenc}    
\usepackage{hyperref}       
\usepackage{url}            
\usepackage{booktabs}       
\usepackage{amsfonts}       
\usepackage{nicefrac}       
\usepackage{microtype}      
\usepackage{xcolor}         
\usepackage{multicol}
\usepackage{multirow}
\usepackage{xspace}
\usepackage{tikz}
 \usepackage{subfig}
 \usepackage{graphicx}
\usepackage{wrapfig}
\usepackage{comment}
 \newcommand{\mar}{\texttt{Mar}\xspace}
\newcommand{\sex}{\texttt{Sex}\xspace}
\newcommand{\race}{\texttt{Race}\xspace}
\newcommand{\asian}{\texttt{Asian}\xspace}
\newcommand{\female}{\texttt{Female}\xspace}

\newcommand{\hair}{\texttt{Hair Color}\xspace}
\newcommand{\hairblack}{\texttt{Black Hair}\xspace}
\newcommand{\hairblond}{\texttt{Blond Hair}\xspace}
\newcommand{\black}{\texttt{African-American}\xspace}
\newcommand{\income}{\texttt{ACS Income}\xspace}
\newcommand{\dutch}{\texttt{Dutch Census}\xspace}
\newcommand{\celeba}{\texttt{CelebA}\xspace}

\usepackage{amssymb} 
\usepackage{enumitem}

\newcommand{\att}{\textsc{Attribute-Bias}\xspace}
\newcommand{\val}{\textsc{Value-Bias}\xspace}
\newcommand{\homogeneous}{\textsc{Homogeneous Bias}\xspace}

\definecolor{attcolor}{RGB}{55,126,184}  
\definecolor{valcolor}{RGB}{200,120,0}
\definecolor{homcolor}{RGB}{0,158,115}

\newcommand{\fairdataset}{{FeDa4Fair}\xspace}

\begin{document}

\title{\fairdataset: Client-Level Federated Datasets for Fairness Evaluation}

  
\author{Xenia Heilmann}
\affiliation{%
  \institution{Institute of Computer Science, Johannes Gutenberg University}
  \city{Mainz}
  \country{Germany}}
\email{xenia.heilmann@uni-mainz.de}
\author{Luca Corbucci}
\authornote{Work was partially done while at the University of Pisa.}
\affiliation{%
  \institution{Fondazione Bruno Kessler}
  \city{Trento}
  \country{Italy}
}
\email{lcorbucci@fbk.eu}
\author{Mattia Cerrato}
\affiliation{%
  \institution{Institute of Computer Science, Johannes Gutenberg University}
  \city{Mainz}
  \country{Germany}}
\email{mcerrato@uni-mainz.de}
\author{Anna Monreale}
\affiliation{%
  \institution{University of Pisa}
  \city{Pisa}
  \country{Italy}
}
\email{anna.monreale@unipi.it}

\begin{abstract}
Federated Learning (FL) enables collaborative training while preserving privacy, yet it introduces a critical challenge: the ``illusion of fairness''. A global model, usually evaluated on the server, appears fair on average while keeping persistent unfairness at the client level. Current fairness-enhancing FL solutions often fall short, as they typically mitigate biases for a single, usually binary, sensitive attribute, while ignoring two realistic and conflicting scenarios: \att (where clients are unfair toward different sensitive attributes) and \val (where clients exhibit conflicting biases toward different values of the same attribute).
To support more robust and reproducible fairness research in FL, we introduce \fairdataset, the first benchmarking framework designed to stress-test FL fairness methods under these heterogeneous conditions. Our contributions are three-fold: \textbf{(1)} We introduce \fairdataset, a library designed to create datasets tailored to evaluating fair FL methods under heterogeneous client bias; \textbf{(2)} we release a benchmark suite generated by the \fairdataset library to standardize the evaluation of fair FL methods; \textbf{(3)} we provide ready-to-use functions for evaluating fairness outcomes for these datasets.
\end{abstract}
\maketitle

\section{Introduction}

With the increasing use of Machine Learning (ML) in critical economic and societal sectors, the demand for its responsible use is growing. This shift has led to the introduction of several Artificial Intelligence (AI) regulations~\citep{roberts2021chinese, biden2023executive, trustworthyAIEU, madiega2021artificial} and the emergence of new research fields such as explainability~\citep{bodria2023benchmarking}, fairness~\citep{caton_2024_survey}, and user privacy~\citep{survey_privacy}.

To mitigate privacy risks in decentralized settings, Federated Learning (FL)~\citep{mcmahan2023communicationefficient} has emerged as a standard solution enabling collaborative model training without requiring users, commonly called clients, to share raw data. However, a significant challenge in FL is the inevitable presence of non-independent and identically distributed (non-i.i.d.) data. While substantial progress has been made in addressing the utility degradation caused by non-i.i.d. conditions~\citep{surveyFLnoniid, 10492865, 10.1145/3625558, karami2025harmony}, efforts to mitigate the resulting unfairness remain limited. Existing fair FL approaches~\citep{papadaki2022minimax, puffle, abay2020mitigating, pentyala2022privfairfl, selialia2024mitigating, rychener2025, mmdm, badar24} typically reduce unfairness for underrepresented groups using group-level metrics such as Demographic Disparity or Equalized Odds Difference~\citep{barocas2017nips}, and in doing so rely on a critical simplifying assumption: that bias is uniform across the federation, both in terms of sensitive attributes and affected subpopulations. 

In practice, this assumption rarely holds. Real-world federated settings are inherently heterogeneous. Clients operate within distinct legal, cultural, and socioeconomic contexts that induce conflicting biases. For instance, one hospital (client) may exhibit data bias against a specific ethnic group due to local demographics, whereas another may exhibit bias against a gender group due to differences in data collection pipelines.
Under such heterogeneity, evaluating a global model on the server using a single fairness metric can create an illusion of fairness: the model may appear fair on average while remaining substantially unfair for individual clients or intersectional groups~\citep{ourpaper}. This has real consequences: Clients who are harmed by the global model have little incentive to further participate in the federation, bias propagates across clients~\citep{biaspropagation}, and the resulting global model can be both less fair and less accurate than models trained locally~\citep{ourpaper, taik2025fairness}, defeating the very purpose of federated collaboration.

\begin{figure}[t]
    \centering
    \includegraphics[width=0.9\linewidth]{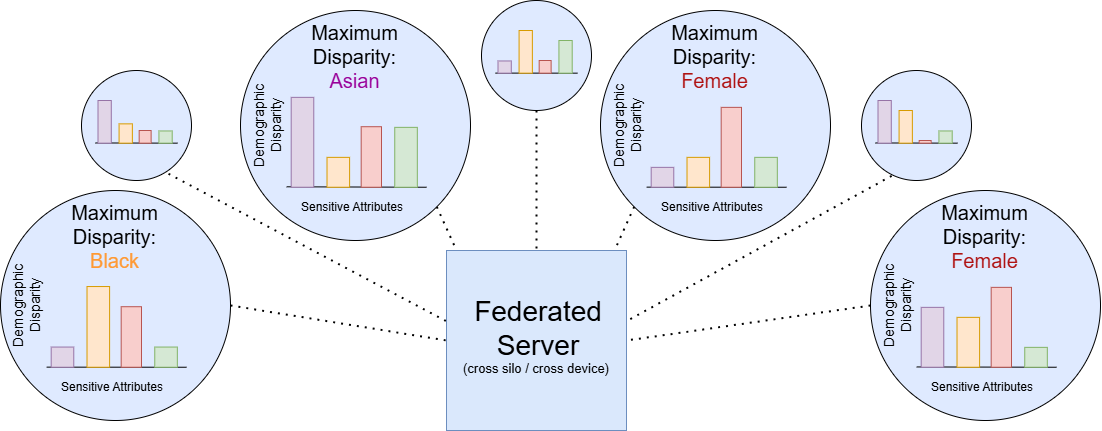}
    \caption{A pictorial representation of the FL scenarios tackled by \fairdataset. Four simulated clients exhibit varying levels of maximal bias, depicted as Demographic Disparity values. Two clients are unfair toward different values of the same sensitive attribute (\val: one toward \black, one toward \asian within \race). Across all four clients, they are unfair toward different sensitive attributes altogether (\att: two toward \race, two toward \sex).}
    \label{fig:summary}
\end{figure}

Despite growing recognition of these issues~\citep{ourpaper, taik2025fairness, heilmann25a}, the field lacks the infrastructure to study them systematically. In this work, we argue that a more granular client-level evaluation framework is necessary to address the current limitations. We identify two specific heterogenous bias scenarios that current benchmarks fail to capture, together covering the primary axes along which client bias diverges in practice: \textbf{(I)} \val, where clients are biased toward conflicting \emph{values} of the same sensitive attribute (e.g., one client is unfair toward \black, while another toward \asian within the attribute \race); and \textbf{(II)} \att, where clients exhibit bias toward entirely different sensitive \emph{attributes} (e.g., the two already mentioned clients are unfair toward \race while the two other clients are unfair toward \sex). We visualize these scenarios in Figure~\ref{fig:summary} with four clients and highlight their highest Demographic Disparity. To illustrate why \val fairness is particularly challenging, consider a federated facial recognition model trained across multiple smartphones. Since users typically store photos of themselves, each client's local dataset is naturally skewed toward their own demographic. A client whose data is biased toward \asian faces expects the global model to perform well on that subgroup; a client skewed toward \black faces has the same expectation for theirs. These objectives are inherently in conflict.
Standard fair FL methods typically address unfairness only where it is most pronounced globally. Concretely, they implement fairness constraints for the demographic group (the attribute value) exhibiting maximal unfairness at the federation level. This means \val fairness constraints are often left unsatisfied. Worse, even when server-side evaluation reports that fairness constraints are met overall, the model may remain unfair for clients whose data is concentrated on specific minority values. Consequently, conflicting client interests can make it difficult, if not impossible, to achieve meaningful fairness in the federated model, even when federation-level mitigation is applied~\cite{ourpaper}.

\att raises analogous but distinct challenges. When clients' data are biased with respect to entirely different attributes (\sex for some, \race for others), standard mitigation strategies that target a single attribute are misaligned with the problem. Federation-level interventions become ineffective or even harmful to participants whose fairness concerns are not represented in the global objective.

Empirically validating these failure modes requires real-world federated datasets in which both scenarios naturally occur across clients. However, such datasets are exceedingly rare. 
What is needed instead is a controlled simulation environment that can systematically stress-test fair FL solutions under worst-case bias heterogeneity before real-world deployment. Recent work by~\citet{taik2025fairness} and~\citet{ourpaper} explicitly calls for such datasets, highlighting their importance for evaluating FL models in settings where clients may hold diverse or even conflicting fairness objectives. To the best of our knowledge, however, no such resource currently exists.

We address this gap with ``Client-Level \textbf{Fe}derated \textbf{Da}tasets \textbf{for} \textbf{Fair}ness Evaluation'' (\textbf{\fairdataset}), a library and benchmarking suite designed to systematically stress-test fair FL solutions under heterogeneous and potentially conflicting bias conditions. While prior work has primarily identified these failure modes theoretically or through limited empirical studies~\citep{ourpaper, taik2025fairness}, \fairdataset provides the infrastructure needed for controlled, systematic, and reproducible experimentation. In doing so, it enables researchers to move beyond federation-averaged fairness metrics and evaluate mitigation strategies under the complex and heterogeneous conditions that real-world deployments are likely to face. Our contributions are as follows:

\begin{itemize}[label=\textcolor{attcolor}{$\blacktriangleright$}]
    \item \textbf{Federated dataset generation library.} We introduce \fairdataset \footnote{\scriptsize{\url{https://github.com/xheilmann/FeDa4Fair}}}, a library and dashboard for generating datasets with customizable, heterogeneous bias distributions (e.g., \att or \val). By providing programmatic mechanisms to simulate client-level biases, it streamlines the evaluation of fair FL in complex scenarios. To support reproducibility, \fairdataset includes automated datasheet generation that documents dataset parameters and injected bias configurations.
    \item \textbf{Benchmark suite for heterogeneous fairness evaluation.} We release a benchmarking suite spanning 8 tabular (\income, \dutch) and 4 image (\celeba) datasets, each configured to instantiate both \val and \att conditions, reflecting realistic bias disparities across modalities and application contexts.
    \item \textbf{Empirical evaluation of fairness mitigation under heterogeneity.} We provide a comparative analysis of client-level fairness evaluation against standard federated aggregation (FedAvg) as well as pre-process and in-process fair-FL solutions under \val and \att conditions. Our results highlight the limitations of current ``one-size-fits-all'' mitigation strategies and underscore the need for federated fairness techniques that account for bias heterogeneity.
\end{itemize}

\section{Background and Related Work}\label{sec:bg}

\subsection{Federated Learning} \label{sec:backgroundFL}
FL~\citep{mcmahan2023communicationefficient} is a distributed training framework that enables $K$ clients to train a shared ML model without exposing their private datasets $\mathcal{D}_k$. A central server $S$ orchestrates training by selecting a subset $\chi$ of available clients for $r \in [0, R]$ rounds and aggregating their model updates. At round $r=0$, the server shares a model $\theta_0$ with random weights. The training procedure depends on the chosen aggregation method.
A frequently used approach is Federated Averaging (FedAvg)~\citep{mcmahan2023communicationefficient}. In this case, clients update $\theta_r$ locally for $E$ epochs before sending updates to $S$, which aggregates them as $\theta_{r+1} \leftarrow \sum_{k=1}^{K} \frac{n_k}{n} \theta_r^{k}$, where $\theta_r^k$ are the parameters updated locally by client $k$, $n_k$ the dataset size of client $k$ and $n = \sum_{k=1}^K n_k$. FedAvg minimizes communication overhead while preserving performance, making it widely applied in FL. 
Based on the number and the availability of the clients involved in the training, we can distinguish between cross-silo and cross-device FL. In the cross-silo scenario, there are typically tens to hundreds of clients, such as hospitals or companies, that are always available during the training and possess large volumes of data. On the contrary, the cross-device scenario involves a larger number of clients, each holding a small number of data samples. In this case, the clients are only available under specific circumstances, e.g., a client could be a smartphone and could be available only while charging.
Several frameworks are available to simulate model training using FL~\citep{felebrities}. One of the most well-known frameworks is Flower~\citep{beutel2022flowerfriendlyfederatedlearning}, which we adopted for all the FL experiments presented in this paper. Additionally, the \fairdataset library we introduce is fully compatible with Flower, ensuring a seamless integration for researchers and practitioners. 

\subsection{Fairness in Machine Learning}

Fairness in ML refers to the principle that models' predictions should not systematically disadvantage individuals or groups based on sensitive attributes such as \sex or \race. 
To quantify how much an ML model is biased, multiple fairness metrics are available in the literature~\citep{mehrabi2021survey, caton_2024_survey}. 

\noindent In this paper, we measure fairness using Demographic Disparity (DD) and Equalized Odds Difference (EOD)~\citep{eqop}. 
The former builds upon Demographic Parity~\citep{barocas2017nips}, which requires that the likelihood of a particular prediction outcome must not depend on the membership of a sensitive group. Formally, Demographic Parity can be expressed as: 
\begin{equation}\mathbb P(\hat{Y}=y \mid Z=z) = \mathbb P(\hat{Y}=y \mid Z \neq z)\end{equation} 
where $y$ is one of the possible targets predicted by the model and $z$ is one of the possible values of the sensitive attribute. DD then evaluates the maximum difference between the Demographic Parity of the different sensitive groups: \begin{equation}
    \max_{y\in Y} \max_{z\in Z} \mathbb P(\hat{Y}=y \mid Z=z) - \mathbb P(\hat{Y}=y \mid Z \neq z).
\end{equation}
Equalized Odds instead demands equality of both the true positive and false positive rates across groups. Thus,  Equalized Odds Difference is defined as: 
\begin{equation}
\max_{y\in Y} \max_{z\in Z}\mathbb P(\hat{Y}=y \mid Y=y, Z=z) - \mathbb P(\hat{Y}=y \mid Y=y,  Z \neq z).
\end{equation}
For both EOD and DD, lower values indicate fairer models. 
Additionally, intersectional fairness captures forms of discrimination and societal effects that emerge from the intersection of multiple sensitive features~\citep{crenshaw2013demarginalizing}. For instance, \texttt{young} \black people may experience bias not because of \texttt{Age} or \race alone, but from their intersection.  While each group may not be disadvantaged independently, their intersection can be. Addressing intersectional fairness is challenging, as these subgroups are often small and difficult to identify within the data.

\subsection{Fairness in Federated Learning}

In the FL context, fairness can be interpreted in multiple ways. Early work primarily focused on \textit{Accuracy parity}, aiming to achieve consistent model performance across clients ~\cite{9746300,donahue2023models, li2021ditto, wang2021federated,mohri2019agnostic,li2020fair}, then on mitigating \textit{Group Fairness}~\cite{dwork2011fairness}, which seeks to mitigate disparities across demographic groups. In this paper, we focus on \textit{Group Fairness} mitigation. In the following, we discuss a selection of methods proposed to address group fairness and group bias in FL; we refer the reader to \cite{salazar2025surveygroupfairnessfederated,benarba2025bias} for more comprehensive reviews of the field.

Several families of approaches have been proposed to achieve group fairness in FL.
A first family of solutions consists of making the model fair using a \textit{pre-processing} approach, i.e., computing importance weights for different groups based on their data distributions~\cite{abay2020mitigating, selialia2024mitigating}. We include the former ~\cite{abay2020mitigating} in the benchmarking phase of this paper. Similarly,~\citet{pentyala2022privfairfl} introduced a pre-processing and a post-processing algorithm. Specifically, the post-processing algorithm debiases the predicted outcomes by finding optimal classification thresholds for each group.
Alternatively, unfairness in FL can be mitigated using \textit{in-processing} methods. These methods~\cite{rychener2025,puffle} introduce regularization terms during training to steer the model toward fairer outcomes. In the benchmarks of this paper, we adopted the former~\cite{puffle} to evaluate how it behaves on our datasets. 
A third family of approaches formulates federated training as a \textit{multi-objective optimization} problem~\cite{mmdm, badar24}, balancing multiple desiderata during learning. Typically, objectives include utility maximization, unfairness reduction, privacy preservation, and satisfaction of system-level constraints such as communication efficiency. However, exploring such multi-objective trade-offs is beyond the scope of this work.
In all these methods, fairness is typically framed as a global objective, prioritizing bias reduction in the aggregated model~\citep{salazar2025surveygroupfairnessfederated, state_FL_fairness_2023}. However, a critical limitation of current approaches is the assumption of a uniform fairness objective, i.e., that all clients wish to mitigate bias against the same sensitive attribute (e.g., \race) in the same direction (e.g., \black). Recent studies suggest this assumption is invalid in real-world scenarios, where clients exhibit heterogeneous fairness needs due to varying local demographics and regulations~\citep{taik2025fairness, ourpaper}. For instance, one client may prioritize gender fairness while another prioritizes racial fairness. These conflicting objectives can exacerbate bias propagation if ignored~\citep{biaspropagation}. Standard mitigation strategies often fail in these settings, as optimizing for a global average can inadvertently harm clients with minority fairness objectives~\citep{ourpaper}, making FL participation worthless or even harmful.

\subsection{Federated Learning Benchmarks and Datasets}

Despite the growing popularity of FL research, the field lacks standardized benchmarking practices. This is particularly evident in two areas: the absence of a gold standard in terms of datasets that should be experimented on\footnote{For a summary of federated learning datasets in ML/AI conference publications, see \href{https://flower.ai/blog/2024-12-02-federated-datasets-in-research/}{https://flower.ai/blog/2024-12-02-federated-datasets-in-research/}} and the inconsistent methodologies for data partitioning used to simulate realistic FL scenarios~\citep{g2024noniiddatafederatedlearning}. 

LEAF~\citep{caldas2019leafbenchmarkfederatedsettings} was the first attempt to establish a benchmarking dataset for FL. However, this benchmark only contains a dataset originally designed for the centralized context, which was then adapted to work for FL. Therefore, it does not effectively capture the different client-level distributions that could be present in an inherently federated dataset. To address this limitation, researchers have proposed various approaches for simulating non-i.i.d. data distributions across clients ~\citep{g2024noniiddatafederatedlearning}. The use of the Dirichlet distribution is a popular partitioning method to simulate a non-i.i.d. split. Solutions like NIID-Bench~\citep{li2022federated} proposed the first benchmarking suite to compare different approaches and evaluate the impact on the model quality. More recently, FedArtML~\citep{jimenez2024fedartml} proposed a similar solution to simulate and evaluate how data heterogeneity impacts the quality of the FL model. 

A similar problem appears when fairness is taken into account during the FL process. Most existing datasets used in fairness literature were originally designed for centralized contexts and lack the natural client-level distributions needed for federated environments~\citep{salazar2025surveygroupfairnessfederated}. Beyond the issue of data heterogeneity, these datasets also fail to capture the diverse bias settings that can emerge across clients in real-world FL deployments. Specifically, to the best of our knowledge, no consensus exists on benchmark datasets or libraries designed to create data distributions with controlled bias properties such as \att or \val.
Most of the bias patterns explored in the literature assume simplified settings that do not reflect the complex and conflicting bias distribution observed in practice. Real-world federated systems involve clients with different types and degrees of bias, yet such scenarios remain largely unexplored.
As also highlighted by~\citet{taik2025fairness}, the existence of datasets capable of simulating these conditions would benefit the evaluation of fair FL solutions, especially when dealing with clients with varying fairness preferences and objectives. A standardized benchmarking framework capable of reproducing such bias configurations would support both the development of new bias mitigation techniques and the rigorous evaluation of existing approaches in FL.

\section{\fairdataset}
Meaningful comparison of fair FL methods requires a shift in the evaluation pipeline: from evaluating single global models on the server in simplified client settings to individual-level evaluation in diverse, bias-heterogeneous client settings. To facilitate this and move beyond current fair FL evaluations, we introduce ``Client-Level \textbf{Fe}derated \textbf{Da}tasets \textbf{for} \textbf{Fair}ness Evaluation'' (\textbf{\fairdataset}), a library designed to create datasets for this purpose and help researchers to investigate fairness across a wide range of FL scenarios. 
Unlike existing evaluation pipelines for FL models' fairness evaluation, \fairdataset allows researchers to design complex scenarios in which clients have differing objectives and interests, such as different sensitive attributes on which to intervene. The library is designed with four core principles: modality agnosticism, configurable bias injection, fairness evaluation, and accessibility.

\subsection{Modality-Agnostic Framework} 
\fairdataset is built on top of the Flower framework~\citep{beutel2022flowerfriendlyfederatedlearning} and integrates seamlessly with the Hugging Face Hub~\citep{lhoest-etal-2021-datasets}. This allows \fairdataset to ingest any dataset hosted on the Hub, spanning tabular, textual, and image data, and transform it into a partitioned federated benchmark\footnote{The only prerequisite is the presence of a sensitive attribute (e.g., \sex, \race) to enable the library's bias injection mechanisms.} with optional bias injection. In addition to the open-ended opportunities afforded by \fairdataset, we release twelve benchmark datasets\footnote{\href{https://huggingface.co/datasets/lucacorbucci/FeDa4Fair}{https://huggingface.co/datasets/lucacorbucci/FeDa4Fair}}.
These are based on three well-known datasets, spanning two modalities:
\begin{itemize}
    \item \textbf{Tabular:} We use \income~\citep{ding2022retiringadultnewdatasets} (derived from US Census data) and the \texttt{\dutch} dataset~\citep{dutch}. The \income dataset offers natural geographic partitioning (by U.S. State) to simulate realistic cross-silo settings. The \texttt{\dutch} dataset is similar to the previous one but provides a non-US demographic perspective.
    \item \textbf{Image:} We use the \celeba dataset~\citep{liu2015faceattributes}, a large-scale face attributes dataset. We incorporate it to demonstrate the library's capability to handle tasks where bias is hidden in more complex feature correlations, rather than in tabular columns.
\end{itemize}

\subsection{Configurable Bias Injection} 
\label{sec:configurable_bias}
The core contribution of  \fairdataset is its ability to inject biases to simulate complex federated scenarios with fairness conflicts between clients. 
This ability allows practitioners to move beyond simple settings and define three distinct fairness scenarios:

\begin{itemize}
    \item \textbf{\homogeneous (Baseline):} The standard scenario found in current literature. The dataset is split into $K$ clients, each biased toward the same value of the same sensitive attribute (e.g., all clients are biased against the value \female of the attribute \sex). This is the classic scenario considered until now in the literature. 
    \item \textbf{\val:} A scenario where clients possess different levels of biases toward the values of the sensitive attribute they disadvantage. For example, the federation is split such that $K/2$ clients are more strongly biased against the sensitive attribute \race with value \black and $K/2$ toward \asian.
    \item \textbf{\att:} The most complex scenario, where clients differ in the sensitive attributes toward which they are unfair. For instance, half of the clients are unfair toward the sensitive attribute \sex while the other half exhibit bias toward \race.
\end{itemize}
To amplify or introduce these biases, \fairdataset adapts techniques from data robustness literature~\cite{inject_unfairness}, specifically, sample dropping to simulate under-representation and label flipping to simulate mislabeling, both applied to data instances with negative labels. While traditionally used as ``adversarial attacks'', we repurpose these mechanisms to model structural inequities in our controlled environment,  allowing researchers to systematically simulate specific demographic disparities and historical labeling errors rather than adversarial noise.
To ensure high customizability, the injection process offers granular control over bias injection. Concretely, practitioners indicate the magnitude of bias injection and the location of its injections. In this regard, \fairdataset allows the definition of distinct bias profiles for clients individually or for groups of clients. In the first case, the practitioner specifies the sample dropping probability and label flipping probability per client individually. In the case of groups of clients, instead, the sample dropping or label flipping probabilities for each client are drawn from a Gaussian Distribution. Therefore, the practitioner has to specify the mean $\mu$ and variance $\sigma$ of this distribution. This allows us to model varying degrees of severity across the federation. 

For the location where bias should be injected, practitioners can specify for which attributes and for which attribute values the manipulation should be applied. Concretely, they select the sensitive attributes which should be considered (e.g., \sex, \race), and the attribute values to which the manipulation should apply (e.g., \female, \black).
Furthermore, more granular control is possible by defining an additional attribute and value, e.g., only flip labels for \female instances within the \black group, to enable intersectional bias manipulation. 
Label flipping and sample dropping can be applied independently or combined: for instance, flipping of negative labels for the \black subgroup while dropping at a different probability across datapoints with negative labels for the \asian subgroup.

Formally, let $\mathcal{D}_k = \{(x_i, y_i, z_i^1,\dots,z_i^m)\}_{i=1}^{n_k}$ represent the local dataset of client $k$, where $y_i \in Y$ represents the ground-truth label, $z_i^1 \in Z^1,\dots,z_i^m\in Z^m$ values of the sensitive attributes and $n_k$ the dataset size of client $k$. Practitioners then specify the bias injection policy based on a dropping probability $P_{drop}$ and a flipping probability $P_{flip}$. They also decide on the sensitive attributes $Z^j$ and attribute values $z^j$ for which the bias should be injected. Finally, $P_{drop}$ and $P_{flip}$ are applied to the subset $\tilde{\mathcal{D}_k} = \{(x_i, y_i,z_i^1,\dots,z_i^m)| z_i^j = z^j , y_i =y^-\}_{i=1}^{n_k}$ with $y^-$ denoting the negative label. 

Finally, the library is compatible with both naturally partitioned and centralized datasets. Here, centralized datasets refer to datasets that are originally designed for centralized ML experiments (e.g., \dutch, \celeba). For these, the library splits the data and optionally injects bias. Naturally partitioned datasets, instead, are the ones that are grouped into partitions directly during the data collection process (e.g., \income). In this case, the library supports the natural partitioning into clients while optionally injecting additional biases, or can subdivide large clients into smaller sub-partitions to better simulate a cross-device scenario.

\subsection{Fairness Evaluation} 
\label{sec:fairness_specifications} 
\fairdataset provides built-in evaluation tools to evaluate client-level bias distribution. \fairdataset supports both individual and intersectional fairness evaluation, allowing bias assessment based on single or combinations of attributes. The library integrates Demographic Disparity (DD) and Equalized Odds Difference (EOD) as metrics for fairness measurement, calculated at three levels of granularity to uncover different types of heterogeneity:

\begin{figure*}[t]
    \centering
    \begin{minipage}[b]{0.48\textwidth}
        \centering
        \subfloat[Attribute-Silo]{
            \includegraphics[width=0.45\linewidth]{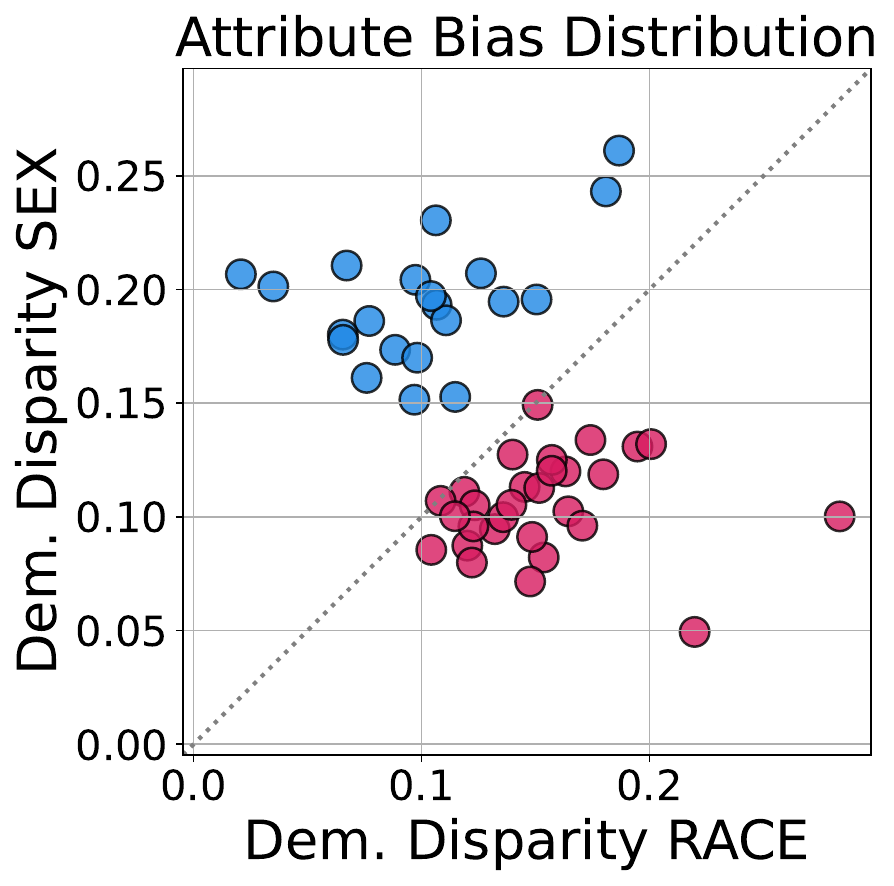}
            \label{fig:fd_cs_att_income}
        }
        \hfill
        \subfloat[Attribute-Device]{
            \includegraphics[width=0.45\linewidth]{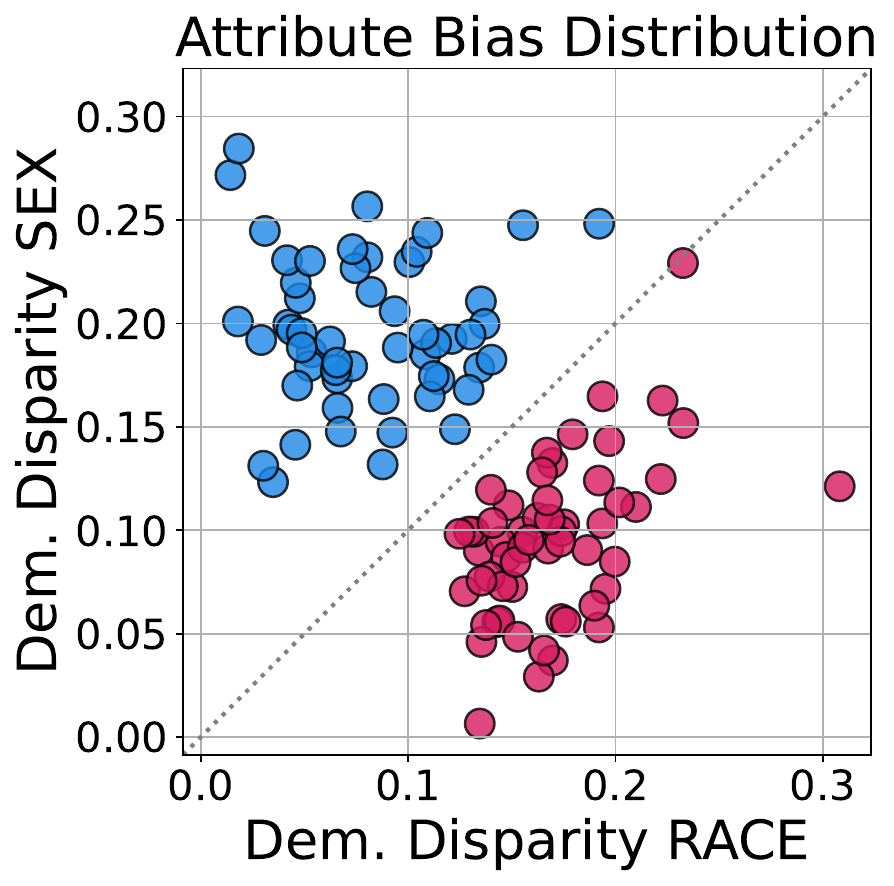}
            \label{fig:fd_cd_att_income}
        }
        \par\vspace{2mm}
        \includegraphics[width=0.8\linewidth]{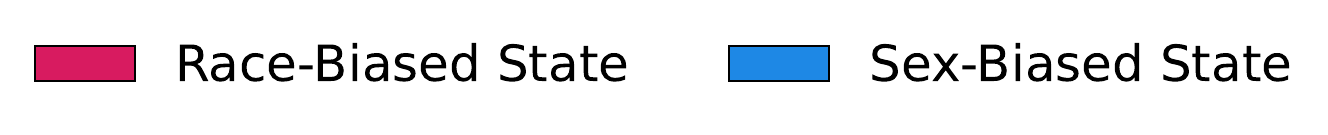}
        
        \caption{\att measured with DD on the local XGBoost models for attribute benchmark datasets using \income.}
        \label{fig:attribute_example_income}
    \end{minipage}
    \hfill 
    \begin{minipage}[b]{0.48\textwidth}
        \centering
        \subfloat[Value-Silo]{
            \includegraphics[width=0.45\linewidth]{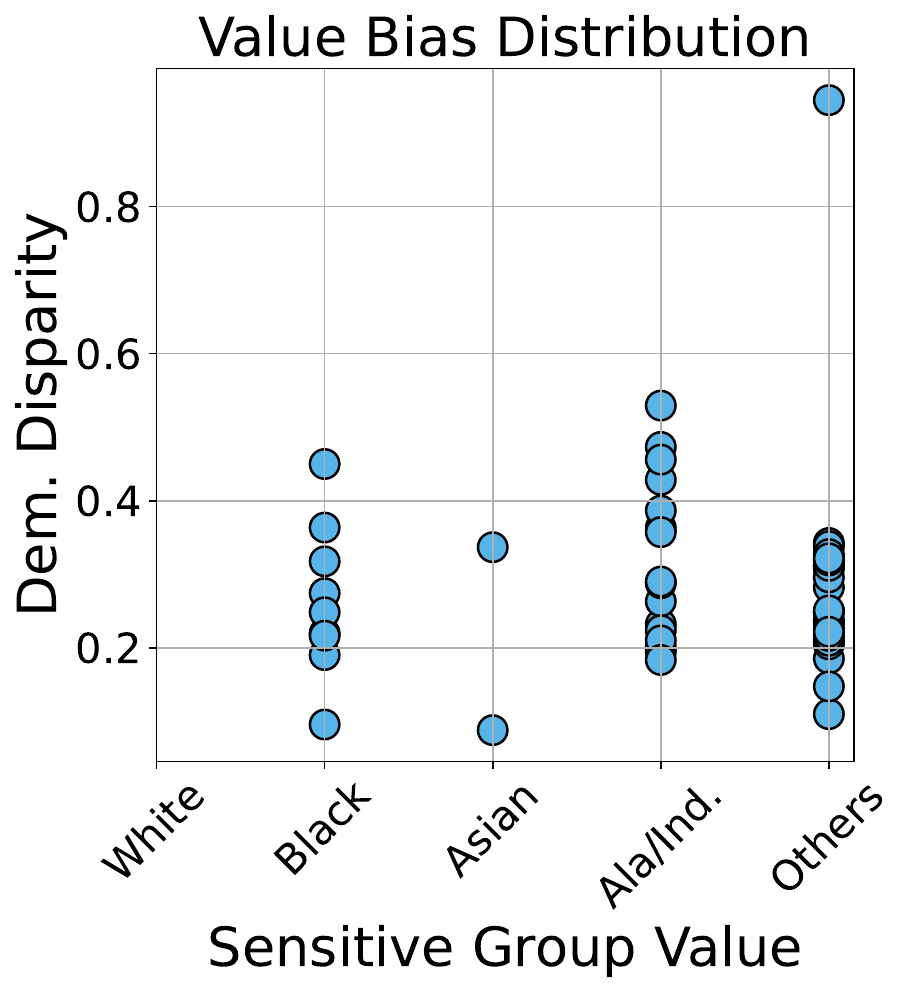}
            \label{fig:fd_cs_val_income}
        }
        \hfill
        \subfloat[Value-Device]{
            \includegraphics[width=0.45\linewidth]{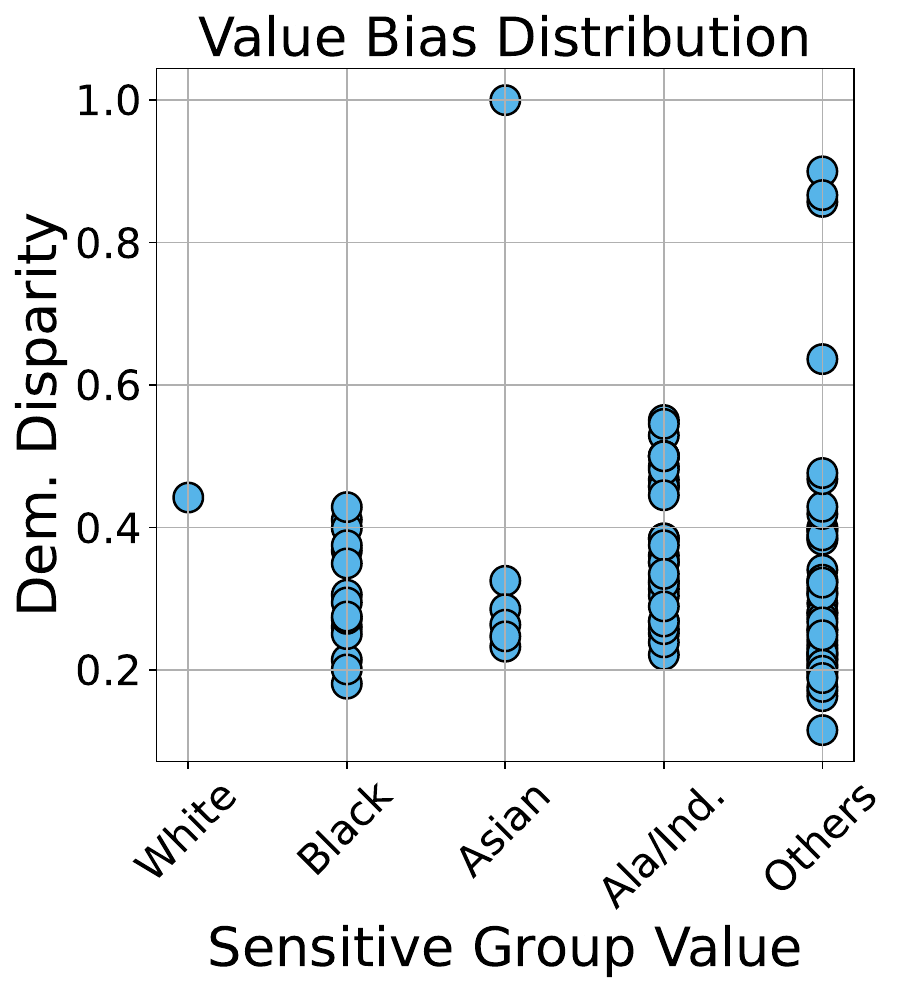}
            \label{fig:fd_cd_val_income}
        }
        \par\vspace{2mm}
        
        \caption{\val measured with DD on the local XGBoost models for value benchmark datasets using \income.}
        \label{fig:value_example_income}
    \end{minipage}
\end{figure*}

\begin{enumerate} 
\item \textbf{Attribute Level:} Reports the maximum unfairness for a specific sensitive attribute (e.g., ``How unfair are the clients toward the sensitive attribute \race?''). 
\item \textbf{Value Level:} Reports the unfairness of the clients toward specific values of a specific sensitive attribute (e.g., ``How unfair are the clients toward the value \asian of the sensitive attribute \race?''). 
\item \textbf{Attribute-Value Level:} A matrix of fairness metrics for all possible attribute-value combinations (e.g., ``How unfair are the clients toward all different values of the sensitive attribute \race?''). 
\end{enumerate}
Figure~\ref{fig:attribute_example_income} illustrates examples of \att in datasets. Each dot represents a local XGBoost model trained on a dataset created via \fairdataset using \income~\citep{ding2022retiringadultnewdatasets}. The plots highlight clients biased toward \race (in red) and \sex (in blue). Similarly, Figure~\ref{fig:value_example_income} shows the \val in datasets, specifically how different clients exhibit varying degrees of maximum DD across the values of the \race attribute. Each dot represents the DD for a specific sensitive group value in a model trained locally on a client.

\begin{figure}
\subfloat[\sex \att.]{\includegraphics[width=0.33\textwidth]{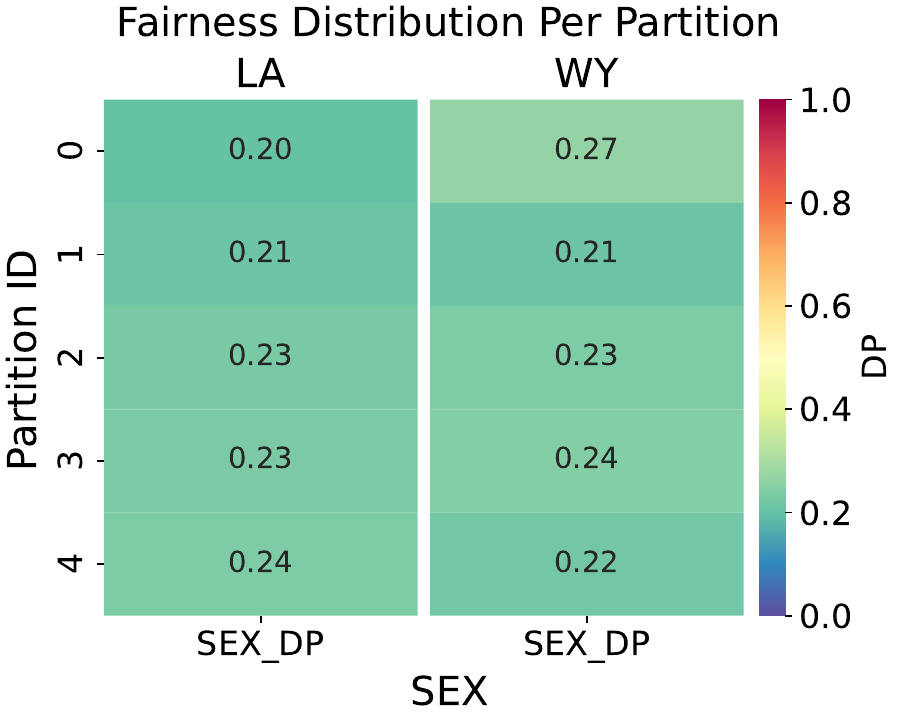}\label{fig:ex_att_sex}}
  \hfill
     \subfloat[\race \att.]{\includegraphics[width=0.33\textwidth]{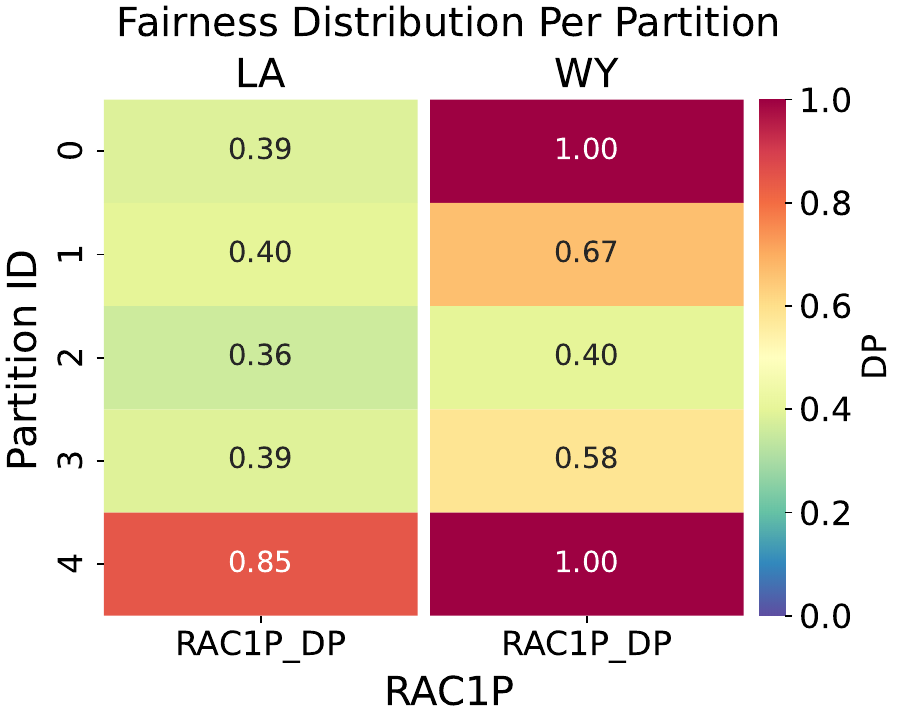}\label{fig:ex_att}}
  \hfill
  \subfloat[\race \val.]{\includegraphics[width=0.33\textwidth]{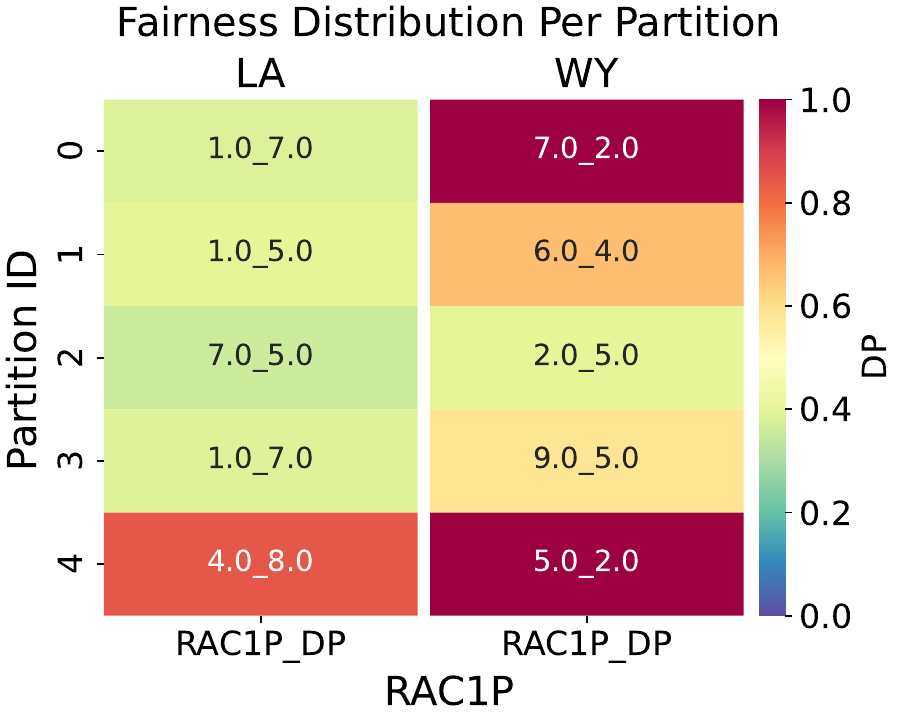}\label{fig:ex_val}}
  \caption{\att and \val measured with DD on the true labels and partitioning data from ``LA'' and ``WY'' of \income. These plots can be generated for any dataset created with \fairdataset.}
  \label{fig:example}
\end{figure}

Similarly, in Figure~\ref{fig:example}, we present a more detailed analysis of fairness distributions for two states, now also partitioned into 5 clients each. Comparing Figures~\ref{fig:ex_att_sex} and~\ref{fig:ex_att}, we observe distinct bias patterns across the \sex and \race attributes. Notably, \sex consistently shows lower bias than \race across the clients. Thus, we can easily identify the dominant bias toward specific attributes for each client.
As shown in Figure~\ref{fig:ex_val}, the highest level of DD spans a broad range of attribute values. For instance, in the state of ``LA'', the maximum DD for \race occurs between $z=4$ and $z=8$ where 4 and 8 are two of the possible values for the sensitive attribute \race. 

\subsection{Accessibility and Reproducibility} 

To lower the barrier for interdisciplinary researchers and auditors to apply our framework, \fairdataset includes a web-based Interactive Dashboard~\footnote{The dashboard is available here: \scriptsize{\url{https://feda4fair-submission.streamlit.app/}}}. This tool provides an interface to: 

\begin{enumerate} 
\item Load datasets from the Hugging Face Hub or local repositories. 
\item Inspect the distribution of sensitive attributes across clients via interactive charts before training begins, ensuring the injected biases match the desired research scenario. 
\item Export the generated dataset to use it in simulations and benchmarks. 
\end{enumerate} 
This workflow allows domain experts without deep programming expertise to design fairness benchmarks while ensuring reproducibility. We provide screenshots of the dashboard and details in Appendix~\ref{app:dashboard}.

Recognizing recent concerns regarding the availability and reliability of FL research datasets~\citep{taik2025fairness}, \fairdataset is designed such that all parameter choices to create the dataset are straightforward to disclose. 
To further support this goal, we have implemented a datasheet generation feature that semi-automatically provides documentation of the parameters employed to create any dataset, inheriting broader information about \fairdataset from a fixed template. 
With this, we aim for reproducibility and to ensure robust validation of fair FL research.

\section{Benchmark Datasets}\label{sec:fixed_datasets}

Besides releasing the \fairdataset library to build datasets for FL fairness evaluation, we also propose a new benchmark suite to standardize the evaluation of fair FL methods. These datasets enable a comprehensive evaluation of fair FL methods under diverse bias conditions. While the library supports generating any variation, we release pre-configured benchmarks for three datasets (\income, \dutch, and \celeba) spanning two modalities (image and tabular). 
For each dataset, we consider both the cross-silo and cross-device FL scenario under the \val and \att setting. Thus, in total, we release twelve datasets grouped into four categories: 
\begin{itemize}
    \item[(I)] \textbf{attribute-silo} datasets: \att dataset for the cross-silo setting; 
     \item[(II)]  \textbf{value-silo} datasets: \val dataset for the cross-silo setting; 
      \item[(III)] \textbf{attribute-device} datasets: \att dataset for the cross-device setting; 
       \item[(IV)] \textbf{value-device} datasets: \val dataset for the cross-device setting. 
\end{itemize}

\subsection{\income} 
The core of our benchmark relies on the 2018 \income dataset~\citep{ding2022retiringadultnewdatasets}. We provide four configurations of this dataset to test different bias-conflicting scenarios. All configurations are pre-partitioned to simulate the FL clients.

In the case of a \textbf{attribute-silo} setting, we leverage the natural distribution of data across the 50 U.S. states and Puerto Rico, focusing on the \race and \sex attributes. To align with binary-focused fair FL methods~\citep{puffle,papadaki2022minimax,abay2020mitigating}, we binarize \race into \texttt{White} and \texttt{Others}. The bias of the different clients was exacerbated to achieve a scenario in which all clients satisfy $DD>0.09$. To meet this target, we iteratively increase the sample drop rate of the more biased attribute between \race and \sex. Figure~\ref{fig:fd_cs_att_income} illustrates the resulting bias distributions after these modifications. In our evaluation, 22 states exhibit a higher DD value for \sex, while 29 states show a higher DD value for \race across both models. Further details, including the list of affected states, EOD statistics, and applied data modifications, are provided in Table~\ref{tab:fixds_att_bias_modific} and Table~\ref{tab:fixds_cs_att_bias_dis} in Appendix~\ref{app:fixed_datasets}. 

For the \textbf{attribute-device} scenario, instead, we derived the dataset from the attribute-silo dataset by splitting each state into six subsets. We then sample from these subsets to create datasets satisfying our bias constraints ($DD>0.09$ for each client). As a result, the modifications applied to these datasets are directly inherited from the corresponding parent state dataset in the attribute-silo setting (Table~\ref{tab:fixds_att_bias_modific}). Across these subsets, we observe that 55 states exhibit stronger bias toward \sex and 56 toward \race. These patterns are visualized in Figure~\ref{fig:fd_cd_att_income} and summarized in Table~\ref{tab:fixds_cd_att_bias_dis} in Appendix~\ref{app:fixed_datasets}.

As with the attribute-silo dataset, the \textbf{value-silo} dataset is built on the natural distribution of the \income dataset across 50 U.S. states and Puerto Rico. However, in this case, to better reflect real-world patterns and to induce different distributions of \val, we avoided binarizing the sensitive attribute. This choice is motivated by the observation that bias is often present in only a subset of the attributes' values. Moreover, introducing bias into groups that are historically not affected would be inappropriate and undesirable.
Instead, we considered multiple classes for the \race attribute (\texttt{White}, \black, \asian, \texttt{Alaska Native/American Indian}, \texttt{Others}) as a basis for analysis. Within these, we identify the attribute values exhibiting the highest DD values and apply data point dropping to amplify existing biases where $DD>0.09$ was not satisfied (see Table~\ref{tab:fixds_value_bias_modific} in Appendix~\ref{app:fixed_datasets}). The resulting \val distribution can be found in Figure~\ref{fig:fd_cs_val_income}. 
In our analysis, we observe the most biased group is \black in 9 states, \asian in 2 states, \texttt{Alaska/Indian} in 16 states, and \texttt{Others} in 24 states (see Table~\ref{tab:fixds_cs_value_bias_dis} in Appendix~\ref{app:fixed_datasets}).  

We derived the \textbf{value-device} dataset by partitioning the value-silo dataset into four subsets per state and sampling from those that satisfy our bias constraints. The resulting client-level bias distribution includes 1 state with predominant \texttt{White} bias, 15 states with \black bias, 6 with \asian bias, 31 with \texttt{Alaska Native/American Indian} bias, and 47 with \texttt{Others} bias, as shown in Figure~\ref{fig:fd_cd_val_income}. Further details can be found in Table~\ref{tab:fixds_value_bias_modific} and~\ref{tab:fixds_cd_value_bias_dis} in Appendix~\ref{app:fixed_datasets}. 

\subsection{\dutch}\label{sec:dutchdatasets}

\begin{figure*}[t]
    \centering
    \begin{minipage}[b]{0.48\textwidth}
        \centering
        \subfloat[Cross-silo]{
            \includegraphics[width=0.45\linewidth]{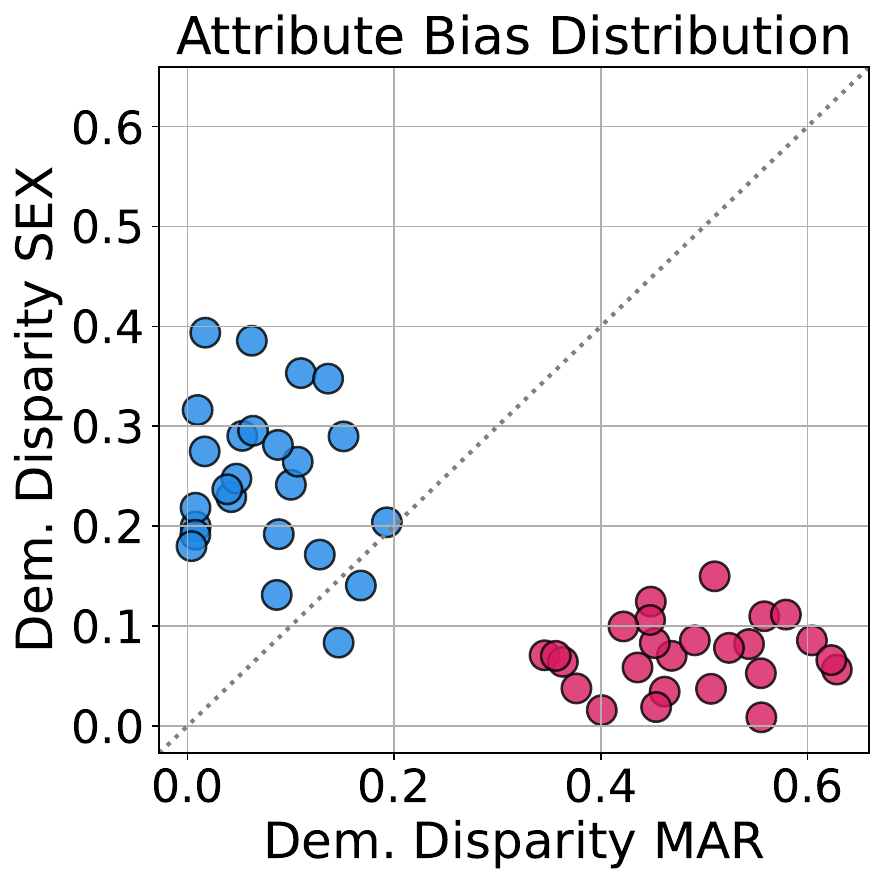}
            \label{fig:fd_cs_att_dutch}
        }
        \hfill
        \subfloat[Cross-device]{
            \includegraphics[width=0.45\linewidth]{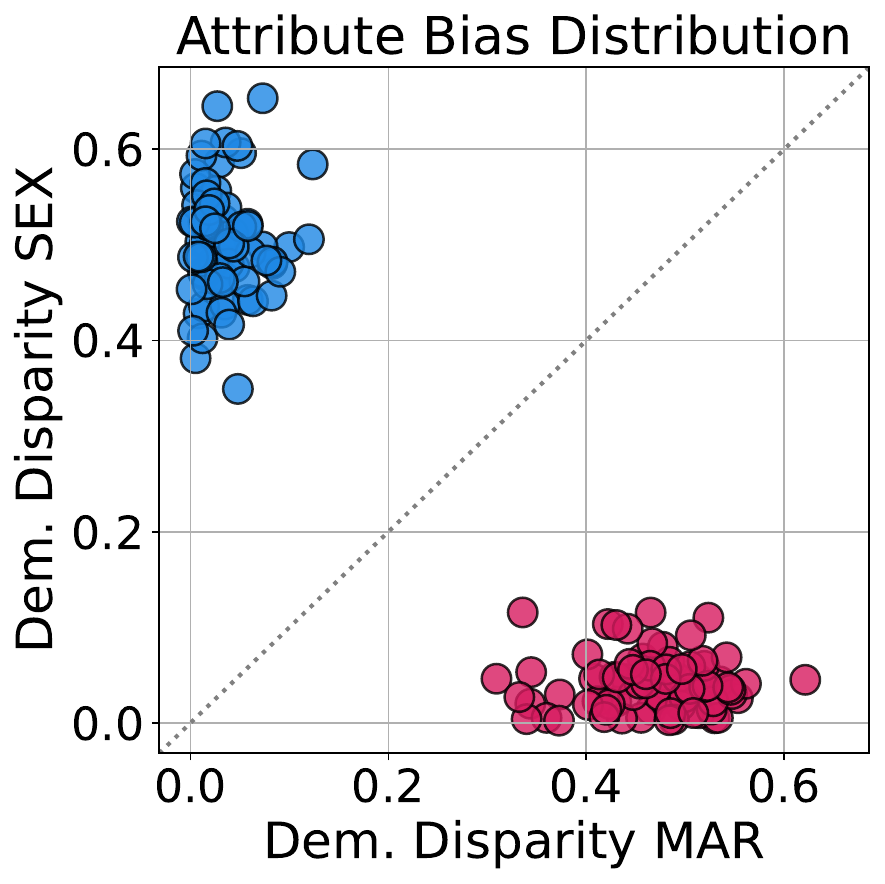}
            \label{fig:fd_cd_att_dutch}
        }
        \par\vspace{2mm}
        \includegraphics[width=0.8\linewidth]{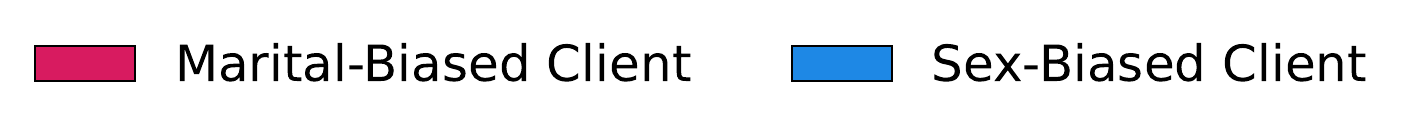}
        
        \caption{\att measured with DD on the local XGBoost models for attribute benchmark datasets using \dutch.}
        \label{fig:attribute_example}
    \end{minipage}
    \hfill 
    \begin{minipage}[b]{0.48\textwidth}
        \centering
        \subfloat[Cross-silo]{
            \includegraphics[width=0.45\linewidth]{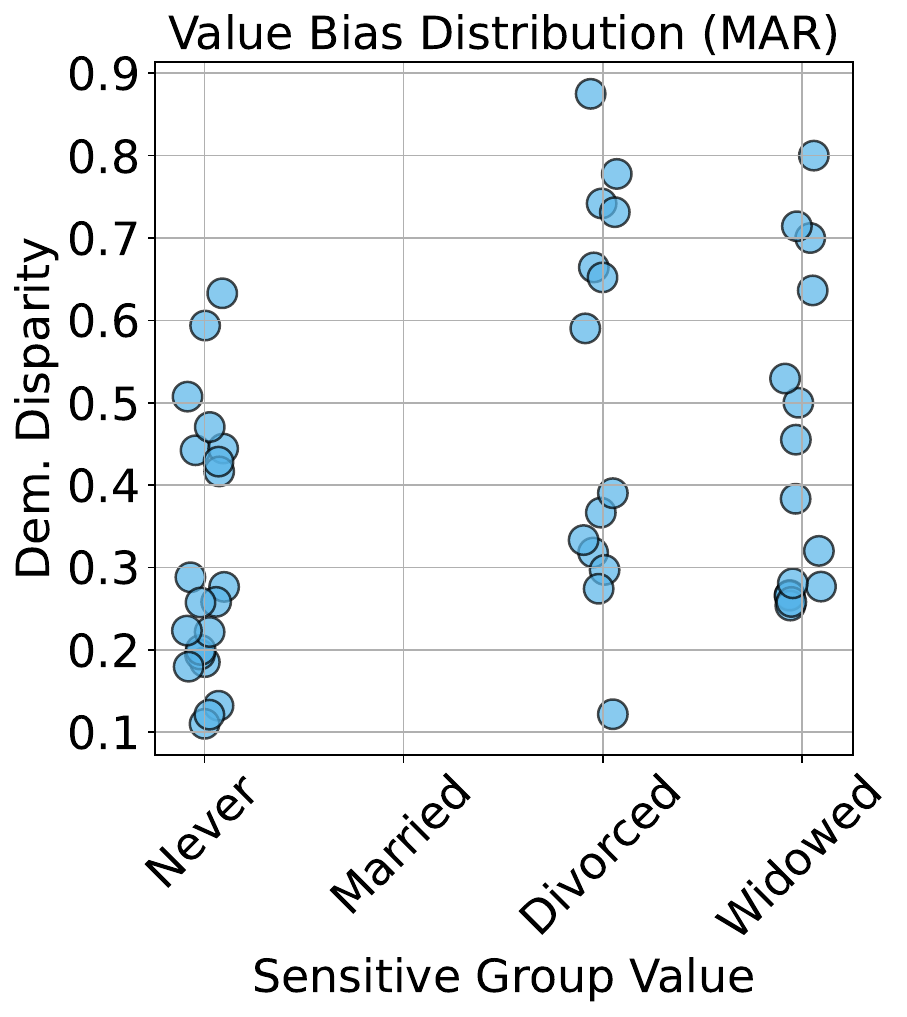}
            \label{fig:fd_cs_val_dutch}
        }
        \hfill
        \subfloat[Cross-device]{
            \includegraphics[width=0.45\linewidth]{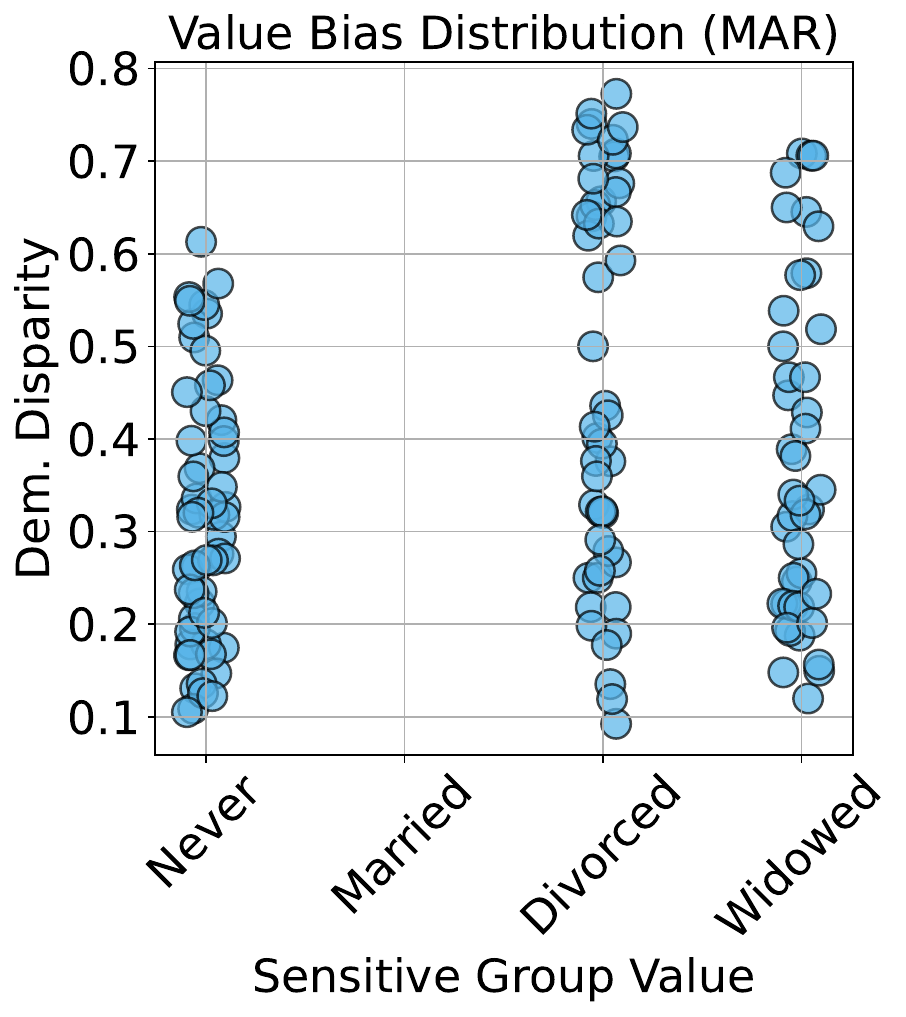}
            \label{fig:fd_cd_val_dutch}
        }
        \par\vspace{2mm}
        
        \caption{\val measured with DD on the local XGBoost models for value benchmark datasets using \dutch.}
        \label{fig:value_example}
    \end{minipage}
\end{figure*}

We release a second benchmark based on the \dutch~\citep{dutch} dataset. Unlike \income, this dataset is originally centralized; therefore, we apply a partitioning to generate the cross-silo scenarios with 50 Clients and the cross-device scenarios with 150 clients. 
Here, for the \textbf{attribute-silo} and \textbf{attribute-device} datasets, we create a conflict between the two sensitive attributes \sex and \mar (marital status). The latter, in this case, we binarized into \texttt{Married} and \texttt{Not Married}. To create both datasets, we apply the library feature that allows us to define groups of users to inject bias into. In the former case, we created two groups of clients; the first one shows higher bias toward the sensitive attribute \sex while it is less biased toward \mar; the latter group has the opposite configuration. The bias distribution of these datasets is visualized in Figure~\ref{fig:attribute_example}, which shows the DD distribution of clients that exhibit higher bias toward \sex in blue and clients that exhibit higher bias toward \mar in red.

For the \textbf{value-silo} and \textbf{value-device} datasets, we considered \mar as a sensitive attribute. As for \income, we do not binarize the sensitive attribute in these scenarios. Thus, we have four possible values: \texttt{Never Married}, \texttt{Married}, \texttt{Divorced}, and \texttt{Widowed}. Figures~\ref{fig:fd_cs_val_dutch} and~\ref{fig:fd_cd_val_dutch} show the \val distribution of the different clients. Here, we can see that no clients are unfair toward the value \texttt{Married}. For the value-silo dataset (Figure~\ref{fig:fd_cs_val_dutch}), clients are distributed among the other three values, while in the value-device scenario, most of the clients are unfair for \texttt{Never Married} and for \texttt{Divorced}. All details about the parameter configuration for these datasets are reported in Appendix~\ref{app:fixed_datasets}.

\subsection{\celeba}

Lastly, to show the compatibility of \fairdataset with image datasets, we also include in our benchmarking suite the four configurations derived from the \celeba dataset. Similar to \dutch, we have 50 clients in the cross-silo settings and 150 clients in the cross-device settings. \celeba, alongside images, also offers a tabular dataset containing information about the people in the images. For instance, for each image we know both the \sex and the \hair, two sensitive attributes that we build the bias injection procedure on. The configurations to obtain these datasets are group-based, as for the \dutch dataset; we report them in Appendix~\ref{app:fixed_datasets} Table~\ref{tab:celeba_settings}.

In particular, for the \textbf{attribute-silo} and \textbf{attribute-device} datasets, about half of the clients have a higher bias for the \sex sensitive attribute than for \hair; the other clients instead have the opposite setting. Due to space constraints, we report these results in Figure~\ref{fig:histogram_celeba} in Appendix~\ref{app:celeba}. In Figure~\ref{fig:distribution_celeba} in Appendix~\ref{app:celeba}, instead, we report the distribution of the maximum Demographic Disparity in the case of the \textbf{value-silo} and \textbf{value-device} datasets. In this case, we created a scenario in which the maximum Demographic Disparity measured on clients is, in about half of the cases, higher for \hairblack and in the other cases higher for \hairblond.

\section{Experimental Analysis}\label{sec:experiments}

We report here an analysis of the results of the FL simulations executed with the four \income datasets released in our benchmarking suite. Due to space constraints, we report the \dutch experiments in Appendix~\ref{app:experiments_dutch}. Unlike the previous two datasets, for \celeba, computational constraints prevented us from running extensive experiments across all dataset splits; we elaborate on this in Appendix~\ref{app:celeba}.

\subsection{Setup}

To contextualize our benchmark datasets, we stress-tested three representative FL paradigms. Our goal is not only to empirically show whether current Fair FL methods can handle the sociotechnical heterogeneity introduced in our scenarios, but also to show challenging FL settings, leaving open research questions for the future.
We compare locally trained client models (Logistic Regression and XGBoost) with three FL training strategies: a Baseline FL model and two Fair FL methods. 
For the baseline model, we apply the FedAvg algorithm~\cite{mcmahan2023communicationefficient}, simulating the FL training with Flower~\cite{beutel2022flowerfriendlyfederatedlearning}, as discussed in Section~\ref{sec:backgroundFL}. For a first Fair-FL benchmark, we use PUFFLE~\citep{puffle} on top of the FedAvg algorithm to reduce model unfairness measured with DD. Specifically for PUFFLE, during training, selected clients compute the gradient and model fairness on a given batch. The computed fairness metric is then incorporated as an additional regularization term in the model update to mitigate unfairness by summing and weighting using a hyperparameter $\lambda$ indicating the importance of the model's utility and its fairness. Here, a $\lambda$ closer to $1$ prioritizes fairness. In our experiments, we treat  $\lambda$ as a hyperparameter. More details about the hyperparameters and hardware underlying the experiments are reported in Appendix~\ref{app:hyp_tuning},~\ref{app:hardware}.

Lastly, we adopt a pre-processing approach for fairness mitigation, Reweighing~\cite{abay2020mitigating}, where each client computes weights based on its local demographic distribution. These weights are applied during the loss computation to penalize errors on underrepresented groups.

In the case of \income, we set for both PUFFLE and Reweighing the strategy to mitigate the unfairness for the \sex sensitive attribute. This choice allows us to evaluate two distinct scenarios: In the attribute-silo datasets, where \sex is one of the sensitive attributes, we assess how mitigation affects the disparity for both \sex-biased clients and differentially biased clients. In contrast, in the value-silo and value-device datasets, we show how mitigating unfairness for \sex influences \val shifts across the clients. 
For the experiments run using the PUFFLE method, a target value for the DD is needed to run the experiments. Specifically, we set the target DD=0.05, which represents the DD value that the global model should maximally exhibit toward \sex; however, there is no formal guarantee that this holds for each individual client. A similar approach is not included in the Reweighing method, which performs the simulation while reducing the unfairness as much as possible.


\begin{figure}[t]
    \subfloat[\textbf{Attribute-silo:} XGB vs. FedAvg]{\includegraphics[width=0.16\textwidth]{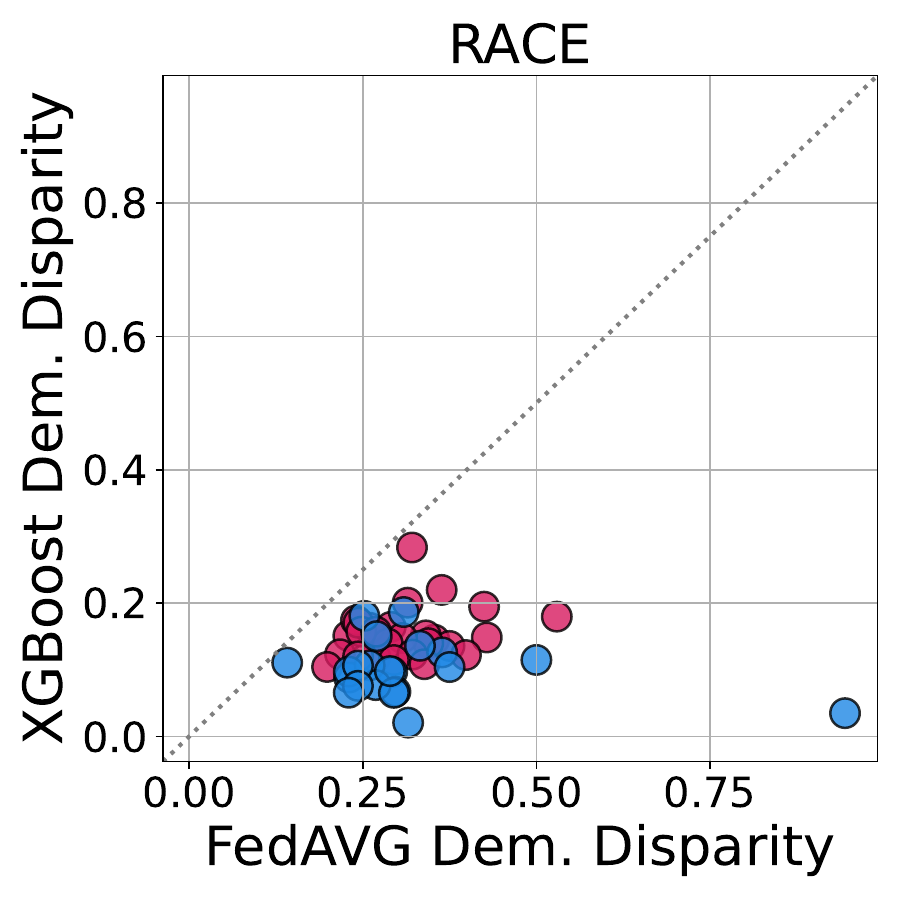}\includegraphics[width=0.16\textwidth]{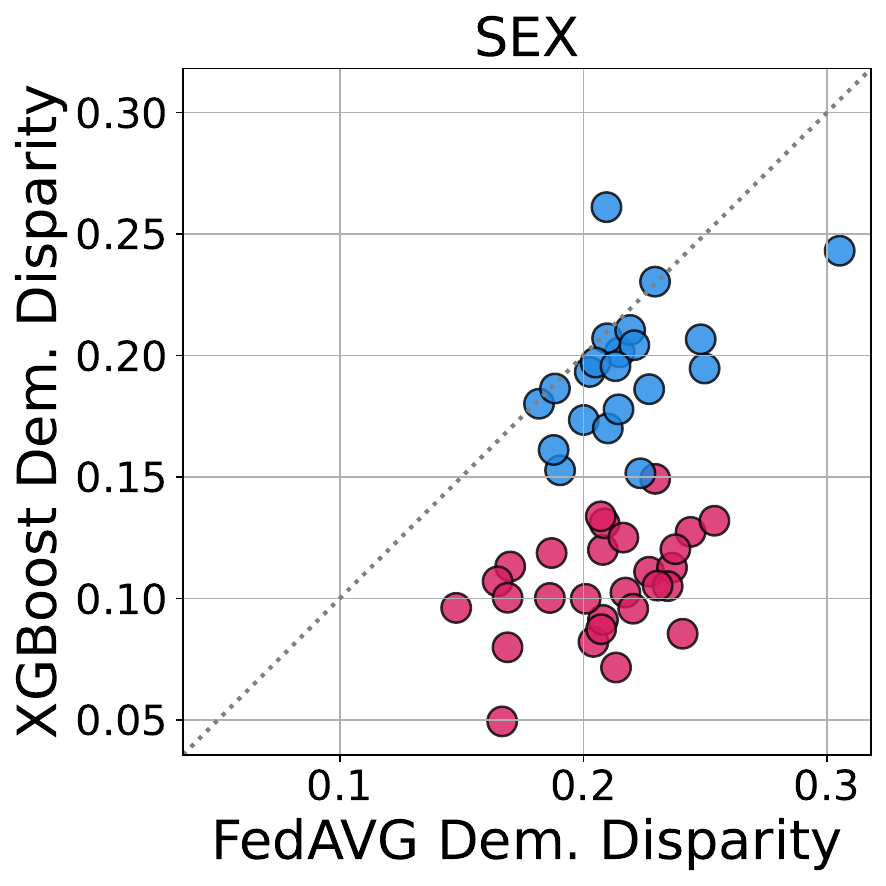}\label{fig:before_after_silo_attribute_xgb_fedavg}}
    \subfloat[\textbf{Attribute-silo:} XGB vs. PUFFLE]{\includegraphics[width=0.16\textwidth]{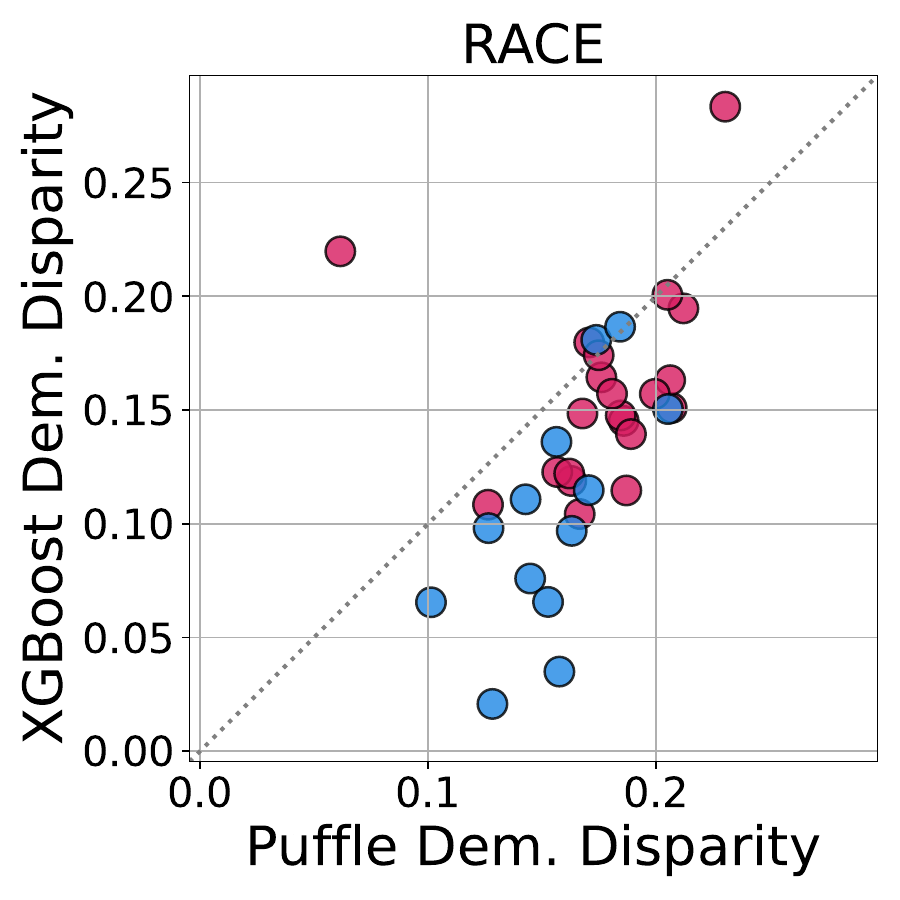}\includegraphics[width=0.16\textwidth]{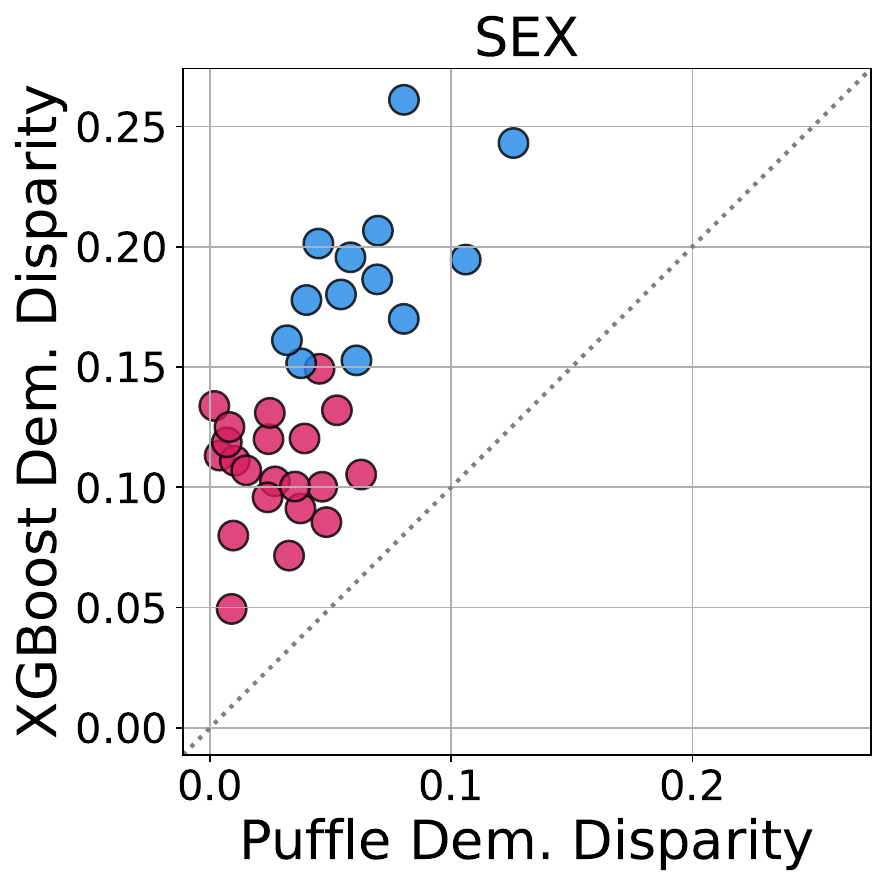}\label{fig:before_after_silo_attribute_xgb_puffle}}
    \subfloat[\textbf{Attribute-silo:} XGB vs. Reweighing]{\includegraphics[width=0.16\textwidth]{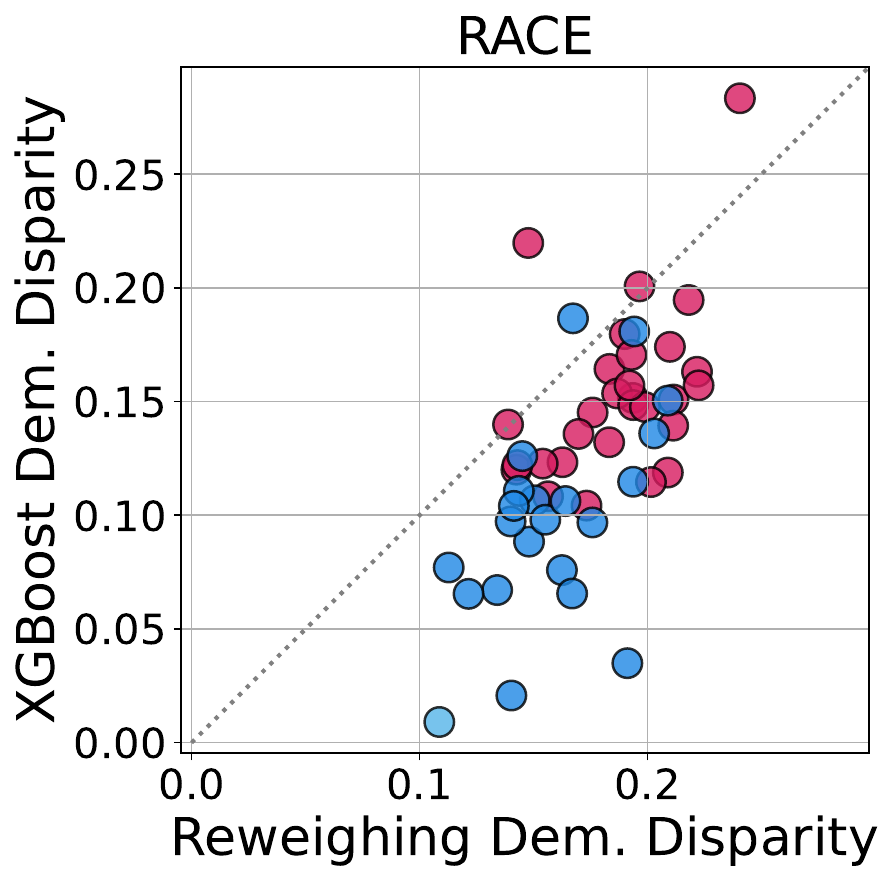}\includegraphics[width=0.16\textwidth]{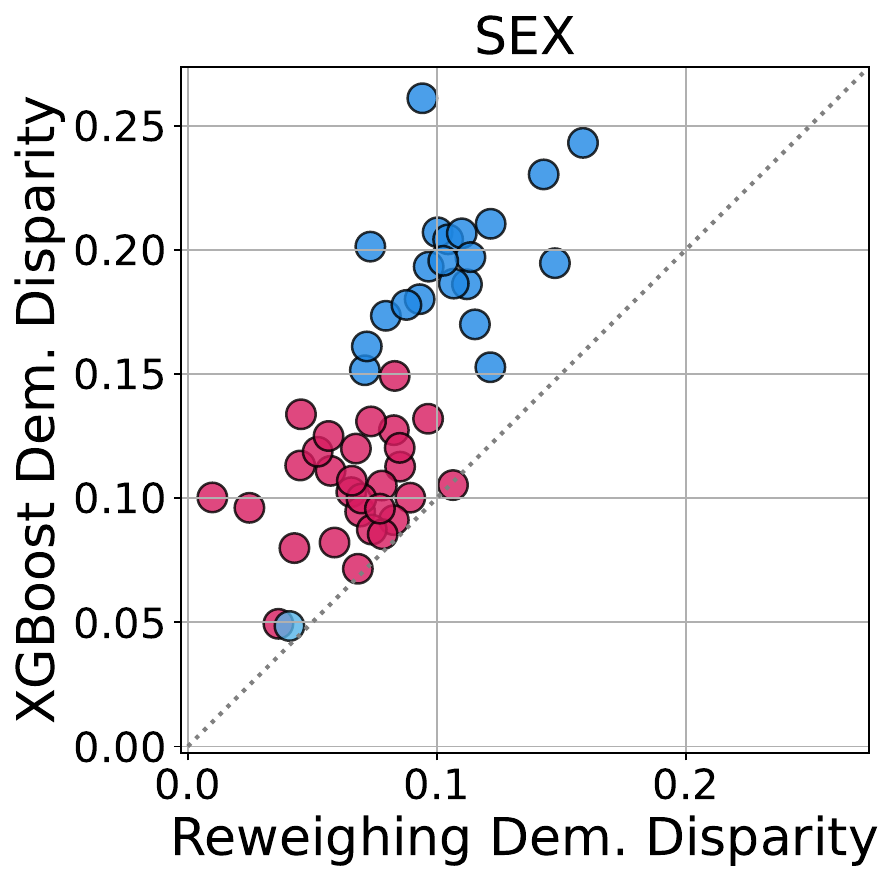}\label{fig:before_after_silo_attribute_xgb_reweighting}}\\
    \centering\includegraphics[width=0.40\textwidth]{images/legend/legend_blue_red.pdf}
  \caption{\income: Individual, per attribute DD measured on the local XGBoost models versus the global FedAvg model, the global PUFFLE model, and the global Reweighing model. Values above the diagonal indicate that the DD of the local model is higher than that of the global model, and the global model reduces bias for this client and attribute. Values below the diagonal indicate that the DD of the local model is lower than that of the global model; the global model increases bias for this client and attribute. }
  \label{fig:before_after_attribute_silo_xgb}
\end{figure}

\subsection{Evaluation}

We assess our methodology based on two principles: quantifying fairness at the individual client level and analysing shifts in bias distribution before and after the application of fair FL methods. Our analysis distinguishes between two FL scenarios:
\begin{itemize}
    \item Cross-Silo: In this scenario, typically, clients possess sufficient data to perform a local train/test split. Therefore, we train individual client‑specific models and compare their performance to that of the global FL models. This allows us to determine whether participation in FL mitigates bias for specific demographic groups or individuals, or conversely, if the existing bias is propagated or even exacerbated. 
    \item Cross-Device: this case involves clients with limited local data, which is insufficient for training robust local models. Here, FL represents the only feasible way to obtain a usable model. In these scenarios, a common practice is to partition clients into a train and a test group. However, a challenge arises: local fairness metrics can only be computed on true labels or external model predictions. A potential solution is to compare the fairness metrics on test clients using the FL model against those obtained from a set of external models trained on similar datasets, possibly provided by the FL orchestrator. However, we leave this for future discussion and evaluate fairness on 30 test clients, which all hold enough data to train a local model. 
\end{itemize}

\begin{figure}
\subfloat[\textbf{Attribute-silo:} XGB vs. FedAvg.]{\includegraphics[width=0.5\textwidth]{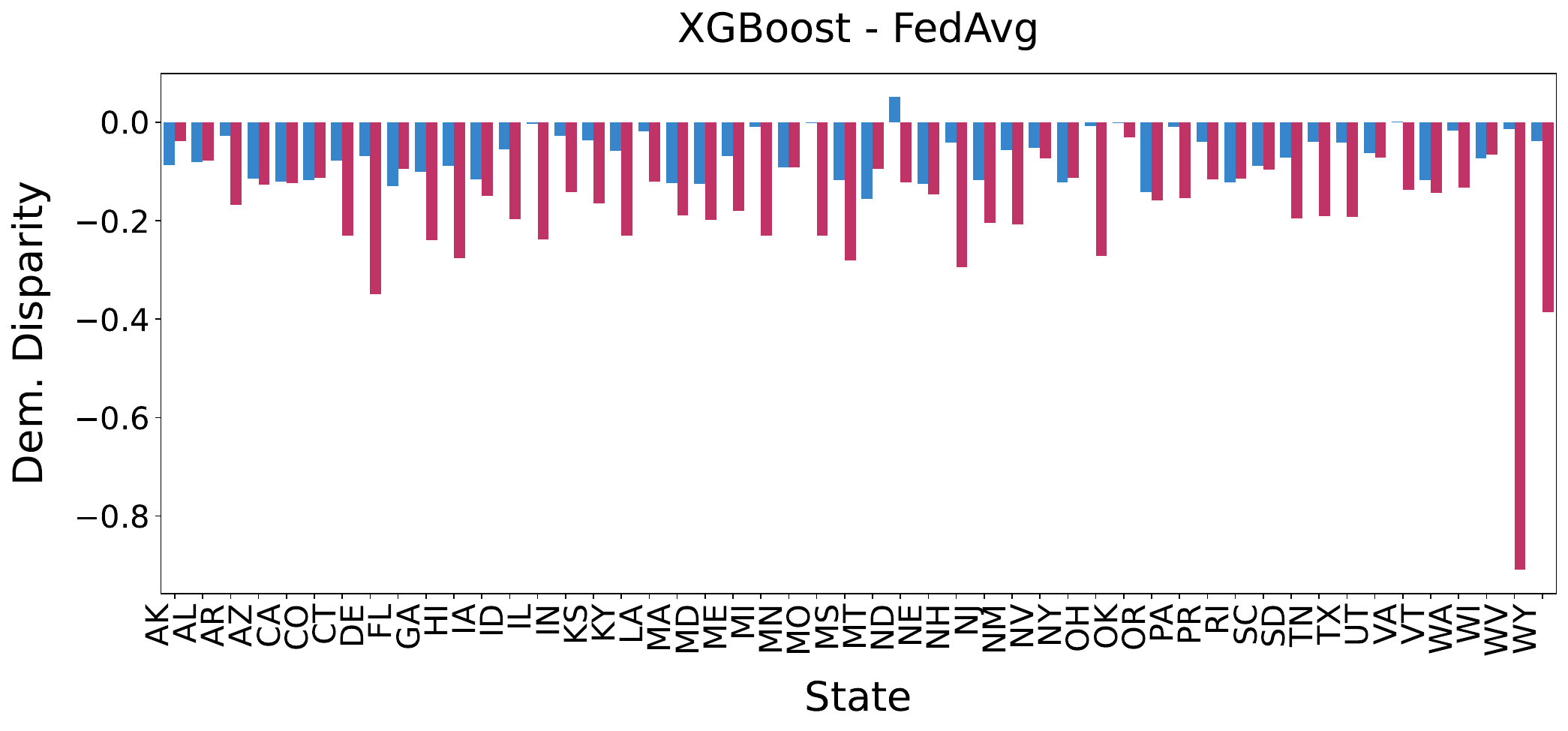}\label{fig:bar_plot_cross_silo_xgb_attribute}}
\subfloat[\textbf{Attribute-silo:} XGB vs. PUFFLE.]{\includegraphics[width=0.5\textwidth]{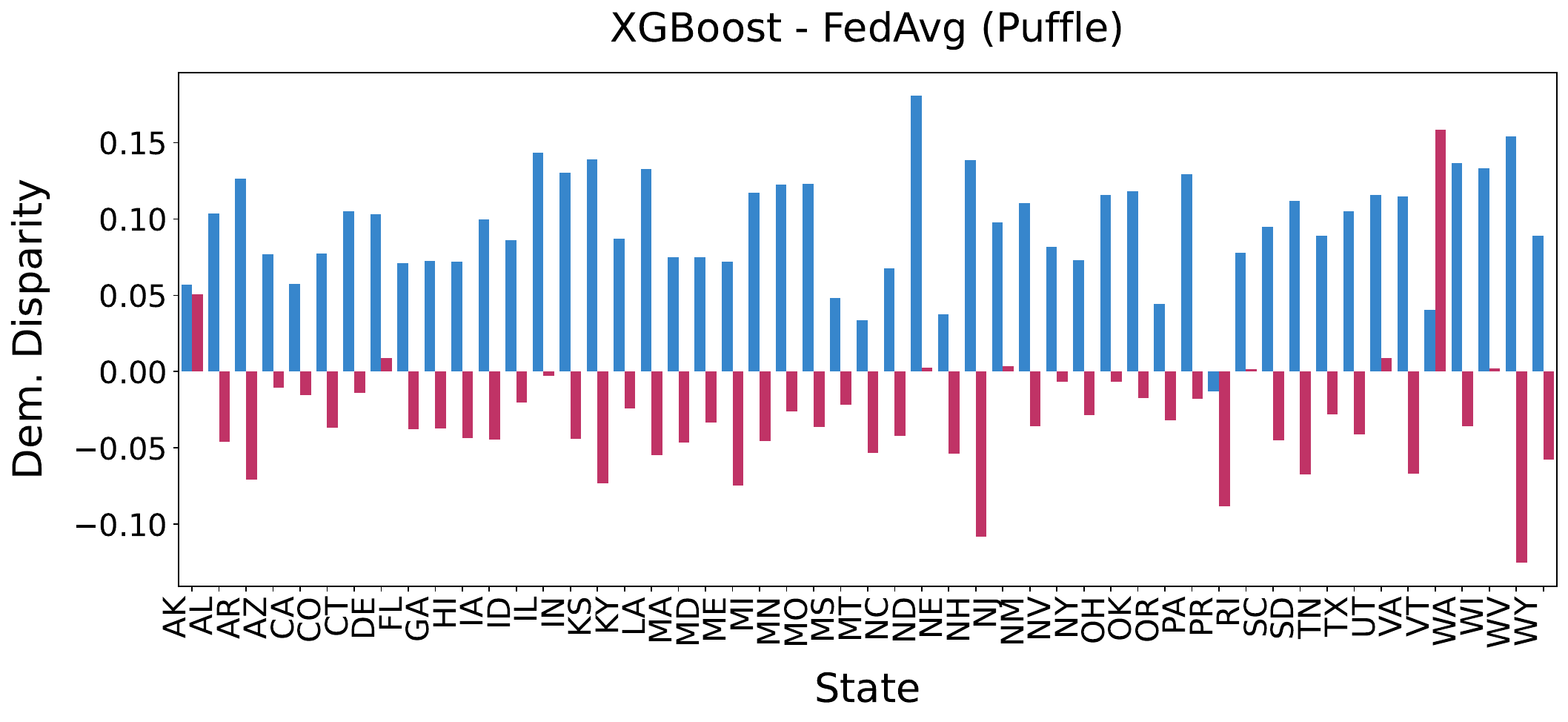}\label{fig:bar_plot_cross_silo_xgb_attribute_puffle}}\\
\subfloat[\textbf{Attribute-silo:} XGB vs. Reweighing]{\includegraphics[width=0.5\textwidth]{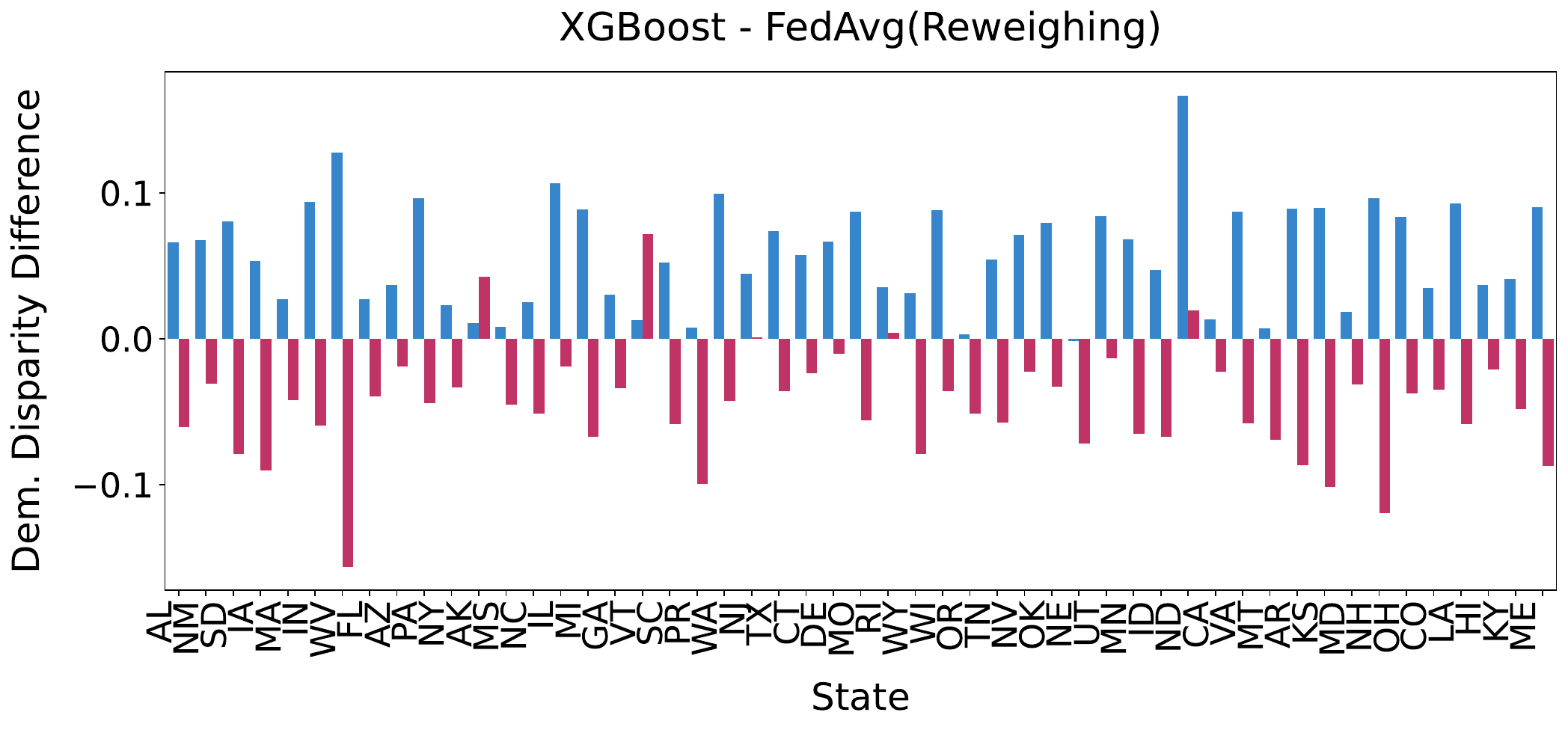}\label{fig:bar_plot_cross_silo_xgb_attribute_reweighing}}\\
\centering\includegraphics[width=0.4\textwidth]{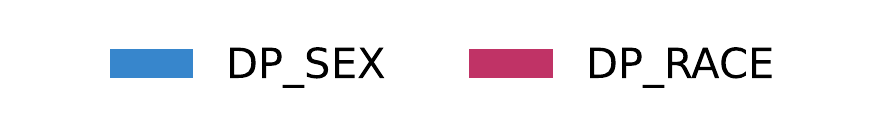}
 \caption{\income: Individual, per attribute, bias differences in DD measured on the local XGBoost model minus the global FedAvg model, the global PUFFLE model, and the global Reweighing model. Values above 0 indicate that the DD of the local model is higher than that of the global model, and the global model reduces bias for this client and attribute. Values below 0 indicate that the DD of the local model is lower than that of the global model; the global model increases bias for this client and attribute. }
 \label{fig:bar_attribute_silo}
\end{figure}

\textbf{\att benchmarks:} The results for the \textbf{attribute-silo} dataset illustrated in Figure~\ref{fig:before_after_attribute_silo_xgb} reveal a trade-off. In these plots, the diagonal line represents a parity line: points below the diagonal line indicate an increase in DD, i.e., an increase in unfairness after FL training compared to the local models. On the contrary, dots above the diagonal suggest greater unfairness with the local XGBoost model. As shown in Figure~\ref{fig:before_after_silo_attribute_xgb_fedavg}, training the baseline with the standard FedAvg drives the majority of clients below this line for both \race and \sex, increasing the DD unfairness and confirming that FL methods without fairness constraints propagate bias across the federation, as previous research has shown~\citep{biaspropagation}. 

When PUFFLE is applied, it effectively enforces fairness constraints on the \sex attribute, resulting in a disparity reduction for \sex across all clients. This is visualized in Figure~\ref{fig:before_after_silo_attribute_xgb_puffle} on the right, where all clients are above the diagonal line. However, this comes at the cost of increased \race disparity compared to the local XGBoost model as depicted in Figure~\ref{fig:before_after_silo_attribute_xgb_puffle} on the left. Thus, clients with maximal bias towards the \race attribute profit less, in terms of bias reduction on their most biased attribute, from participating in the FL training when using PUFFLE.

For Reweighing, a similar trend as for PUFFLE is visible in Figure~\ref{fig:before_after_silo_attribute_xgb_reweighting}. The method effectively enforces fairness constraints on the \sex attribute, resulting in a disparity reduction for \sex across all clients. This is visualized in Figure~\ref{fig:before_after_silo_attribute_xgb_puffle}. Again, on the right, all clients are above the diagonal line, indicating that the DD values for \sex decrease. However, with regard to \race disparity, Reweighing increases this value compared to the local XGBoost models. 
For both Puffle and Reweighing, however, the overall \race disparity is lower than for FedAvg, where the majority of DD values are above $0.25$.

This trend is also visible from the \textbf{client-level} evaluation reported in Figure~\ref{fig:bar_attribute_silo}, where we compare the local XGBoost models with the FL models. While for FedAvg the unfairness increases upon the local models, PUFFLE and Reweighing decrease the unfairness with regard to the \sex attribute. Only for some, such as ``AK'', ``VT'', PUFFLE and Reweighing also decrease the unfairness for \race, highlighting the limitations of single-attribute mitigation in heterogeneous settings. 

This trend remains consistent when using Logistic Regression as the local model (Figure~\ref{fig:before_after_attribute_silo_lr},~\ref{fig:bar_atribute_silo_lr} 
 in Appendix~\ref{app:experiments_attribute}). Additional results with the \textbf{attribute-device} dataset are detailed in Appendix~\ref{app:experiments_attribute}.

\begin{figure}   

    \subfloat[\textbf{Value-silo:} XGB vs. FedAvg]         
    {\includegraphics[width=0.33\textwidth]{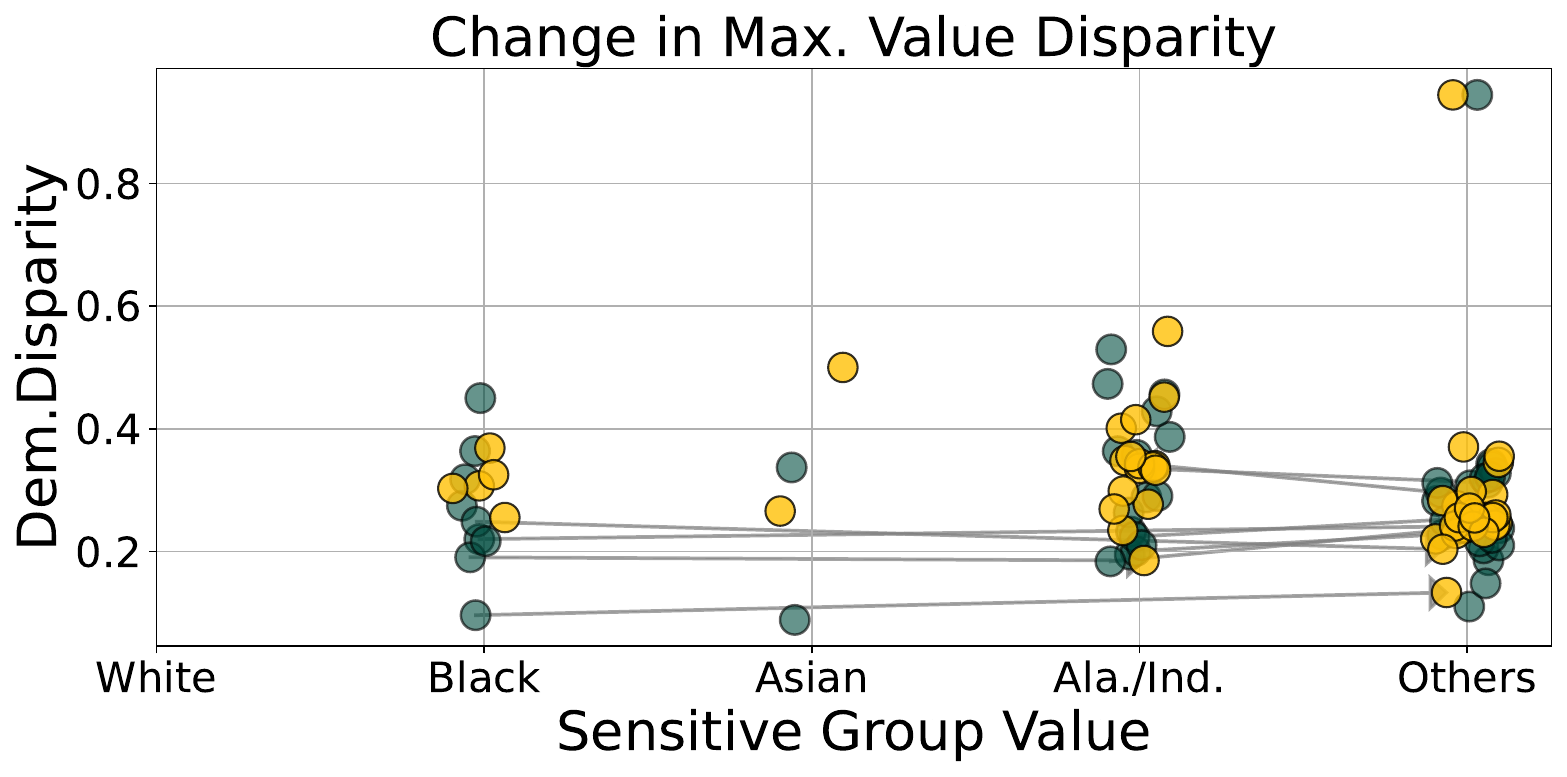}}
    \subfloat[\textbf{Value-silo:} XGB vs. PUFFLE]{\includegraphics[width=0.33\textwidth]{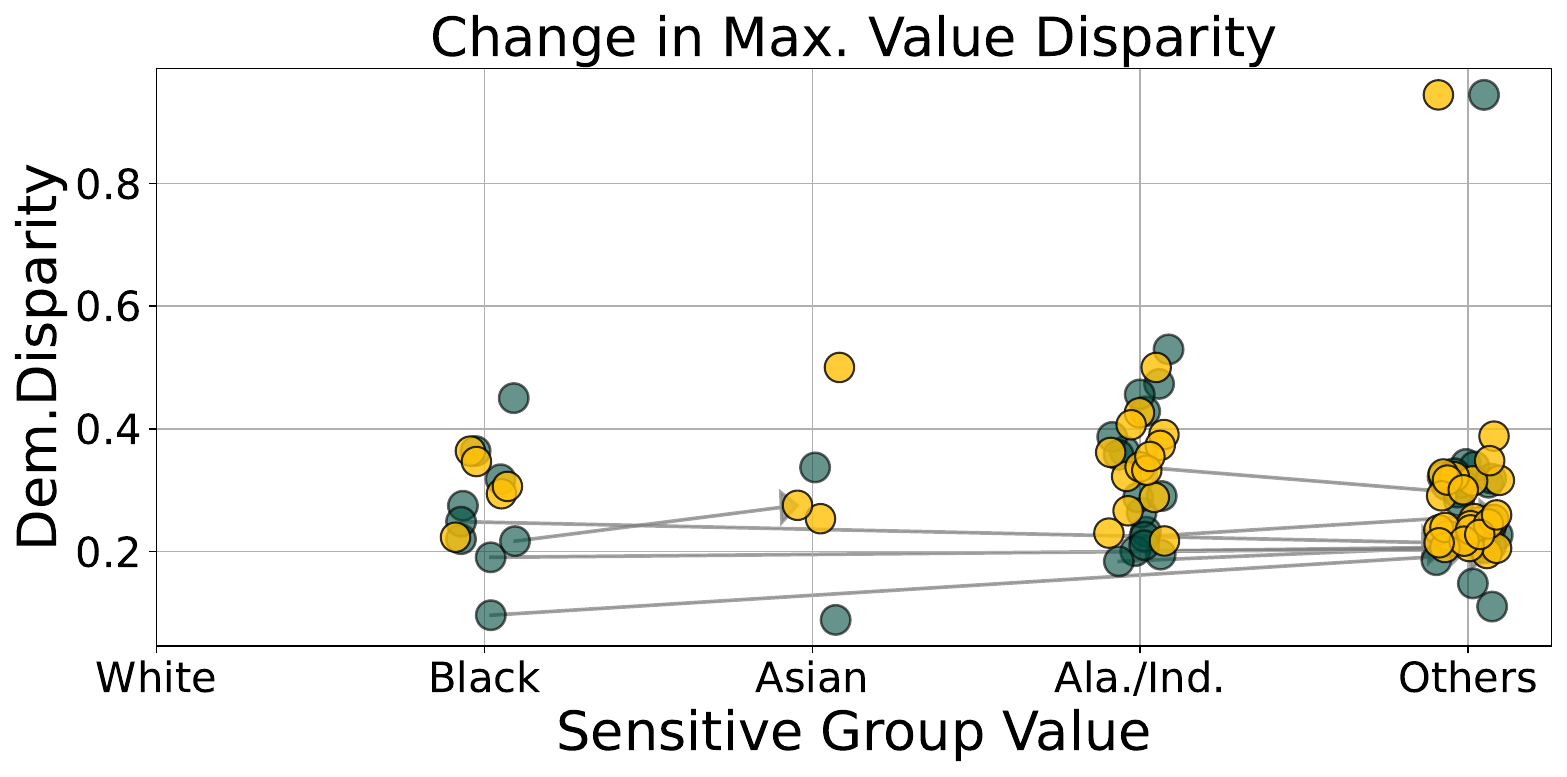}}
    \subfloat[\textbf{Value-silo:} XGB vs. Reweighing]{\includegraphics[width=0.33\textwidth]{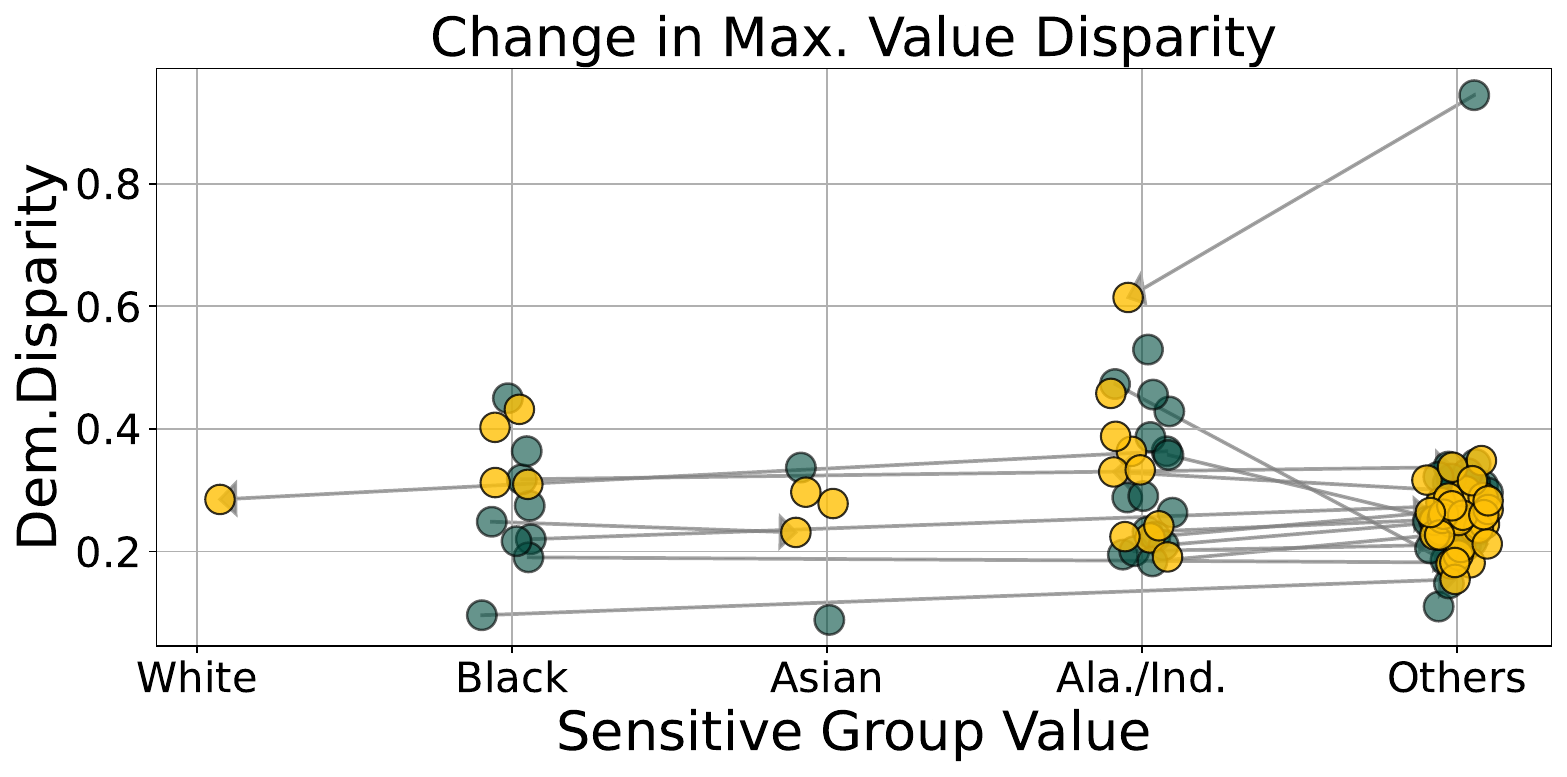}}\\
    \centering\includegraphics[width=0.35\textwidth]{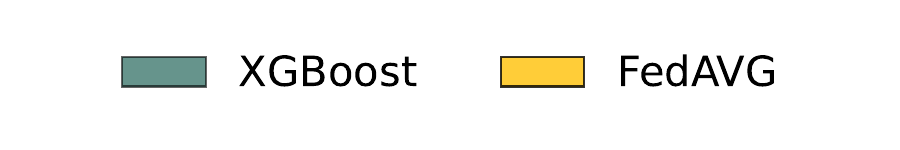}
    \caption{\income: Individual, per value DD measured on the local XGBoost models versus the global FedAvg model, the global PUFFLE model, and the global Reweighing model. The lines indicate how the value changes from local to FL, showing a movement towards underrepresented values when measuring on the global FL models.}
  \label{fig:before_after_value_silo_xgb}
\end{figure}

\textbf{\val benchmarks.}
For the \textbf{value-silo} dataset, Figure~\ref{fig:before_after_value_silo_xgb} shows both how DD changes when moving from local training to FL and how the distribution of the attribute value associated with the maximum DD shifts. A clear pattern emerges: underrepresented groups, specifically, \texttt{Alaska Native/American Indian} and \texttt{Others}, are disproportionately harmed by participation in FL. In particular, clusters corresponding to the \texttt{Others} category increase density after FL training, indicating that this group is more frequently associated with the highest DD values. This effect becomes more pronounced under PUFFLE and even more with Reweighing, where an increasing number of clients report \texttt{Others} as the attribute value with the maximum DD. Despite these shifts, the overall \race disparity shows only a marginal improvement trend across all three settings. A more detailed individual-level analysis is provided in Figure~\ref{fig:bar_value_silo} in Appendix~\ref{app:experiments_value}. 
We observe a consistent trend when using Logistic Regression as the local model, as shown in Figure~\ref{fig:before_after_silo_value_lr} and Figure~\ref{fig:bar_value_silo_lr} 
in Appendix~\ref{app:experiments_value}. Additional results with the \textbf{value-device} dataset are detailed in Appendix~\ref{app:experiments_value}.

Overall, we observe consistent results across both cross-silo and cross-device settings for each type of bias-heterogeneous client configuration. However, the specific form of bias heterogeneity, namely, whether it arises from different attributes or from different attribute values, leads to different outcomes. This observation highlights the importance of evaluating fair FL approaches across a diverse range of bias scenarios to assess their robustness and ensure equitable performance.

\subsection{Key Takeaways}
Our benchmarking datasets reveal several limitations in current Fair FL paradigms:

\begin{itemize}[label=\textcolor{attcolor}{$\blacktriangleright$}]
\item \textbf{Fairness effects are client-specific:} While global fairness metrics may show an improvement, client-level analysis may reveal substantial heterogeneity: FL participation can mitigate, preserve, or exacerbate bias depending on each client’s original data distribution before potential manipulation.
\item \textbf{Single-attribute mitigation has limitations:} Current Fair FL methods successfully reduce disparity for the targeted sensitive attribute but can increase unfairness for other attributes, particularly in bias-heterogeneous federations where clients have conflicting biases (Figure~\ref{fig:before_after_attribute_silo_xgb} and~\ref{fig:bar_attribute_silo}).
\item \textbf{\val exposes limitations of current Fair FL solutions:} Existing approaches typically rely on minimizing the maximum DD for a given sensitive attribute. These methods ignore which specific attribute value is responsible for the highest disparity. In \val heterogeneous settings, this leads to patterns where underrepresented groups are negatively affected by participation in FL training despite the low unfairness of the global aggregated model (Figure~\ref{fig:before_after_value_silo_xgb}).
\item \textbf{Biases in heterogeneous settings:} A shift is required in both the design and evaluation of Fair FL methods. Developing solutions that can handle multiple \att and \val heterogeneity remains an open challenge, highlighting the need for future work on more advanced Fair FL solutions.
\end{itemize}

\section{Conclusion}

Current fairness evaluation in FL is usually based on the assumption of a uniform bias distribution across the clients. This often creates an ``illusion of fairness'' at the global level while ignoring the complex, heterogeneous biases that exist at the individual client's level. Previous work \citep{ourpaper, biaspropagation} and our analysis performed on the benchmarking datasets created with the \fairdataset library revealed that current state-of-the-art solutions are insufficient in realistic scenarios involving \val and \att.

To address this limitation, we introduced \fairdataset, a library designed to create datasets for FL experimentation with heterogeneous client bias. By providing a reproducible and extensible framework, \fairdataset is a crucial first step towards enabling more robust and systematic evaluation of fair FL methodologies. To support this goal, we also published twelve benchmarking datasets spanning tabular and image data, cross-silo and cross-device scenarios, as well as \att and \val settings.

Thanks to its extensibility, \fairdataset is designed to support the continual expansion and diversification of the fairness evaluation landscape in FL. This facilitates two critical directions for future research: first, the inclusion of even more complex real-world scenarios and second, the investigation of fairness in cross-device settings where clients possess only limited data.

\clearpage
\section*{Acknowledgment}
XH and MC were supported by the ``TOPML: Trading Off Non-Functional Properties of Machine Learning'' project funded by Carl Zeiss Foundation, grant number P2021-02-014. AM and LC were supported by the Italian Project Fondo Italiano per la Scienza FIS00001966 ``MIMOSA'' and by the European Union's Horizon Europe research and innovation program ``TANGO'' under G.A. 101120763. Views and opinions expressed are, however, those of the author(s) only and do not necessarily reflect those of the European Union or the European Health and Digital Executive Agency (HaDEA). Neither the European Union nor the granting authority can be held responsible for them.

\section*{Generative AI usage statement}

LLMs were used to aid non-native speakers with grammar and word corrections. During the implementation of the code needed for this submission, we used LLMs for autocompletion of code and to fix bugs.

\section*{Ethical Considerations Statement}
\fairdataset allows for manipulating datasets such that they reflect more complex bias patterns to facilitate fair FL method evaluation and to improve bias mitigation development. However, we stress that manipulated data is, of course, no longer representative of the state-level distributions and demographics in the U.S. and Puerto Rico (for datasets based on ACS) or the Netherlands (for datasets based on Dutch). We strongly advise against employing data manipulated with \fairdataset for anything but the testing and development of federated bias mitigation strategies.

\bibliographystyle{ACM-Reference-Format}
\bibliography{bibliography}
\clearpage

\appendix

\section{\fairdataset Dashboard}\label{app:dashboard}

\fairdataset has been developed with high accessibility in mind. This is why we included a dashboard developed using Streamlit~\footnote{Streamlit \scriptsize{\url{https://github.com/streamlit/streamlit}}} that allows practitioners to easily access all the features offered by \fairdataset without the need to write Python code~\footnote{The dashboard is available here: \scriptsize{\url{https://feda4fair-submission.streamlit.app/}}}. 

\begin{figure}[h]
\centering
\includegraphics[width=0.9\linewidth]{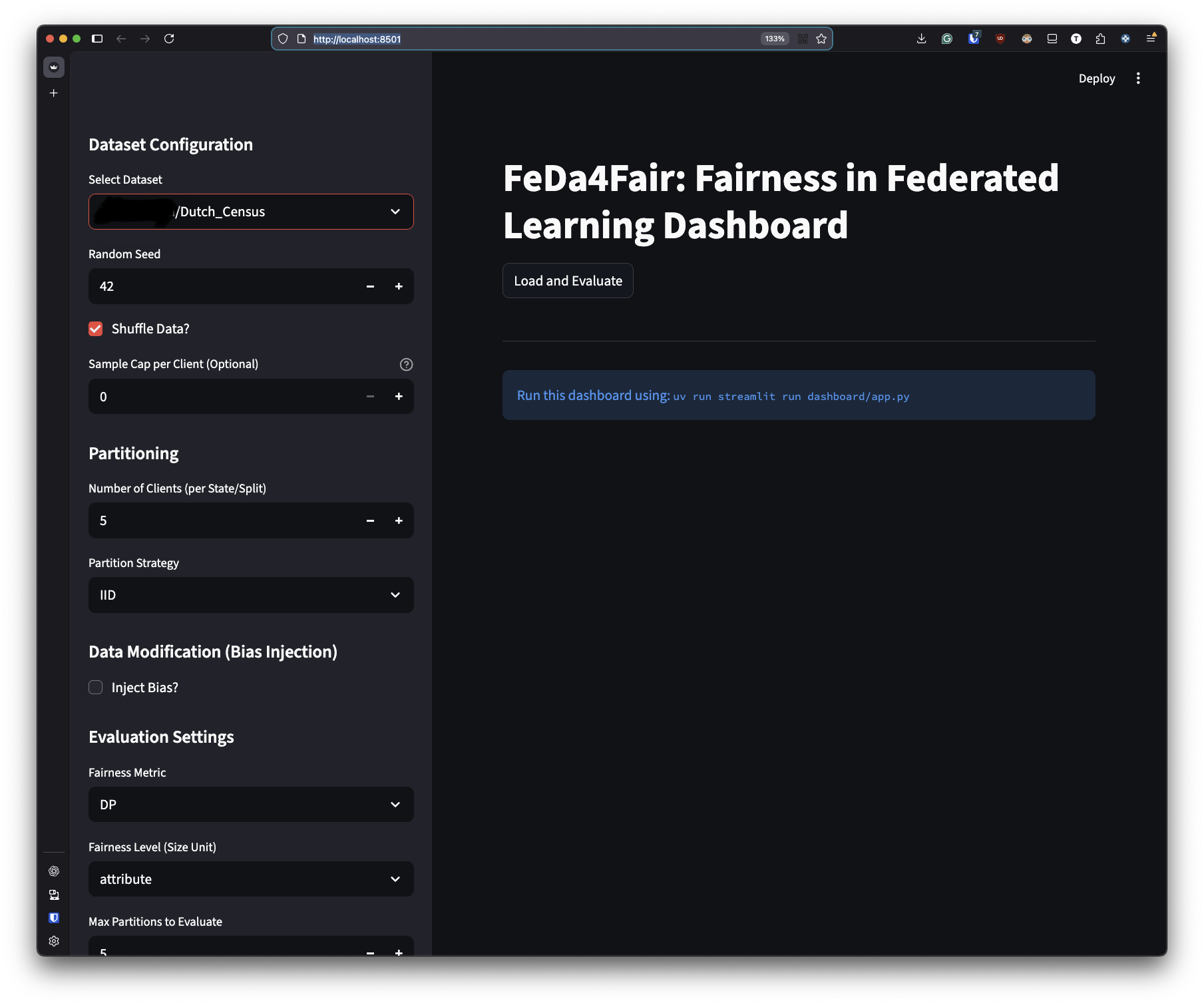}
\caption{Our dashboard allows practitioners to select a dataset that can be split into different clients and to define modifications to inject biases.}
\label{fig:dashboard_1}
\end{figure}

\begin{figure}[t]
\centering
\includegraphics[width=0.8\linewidth]{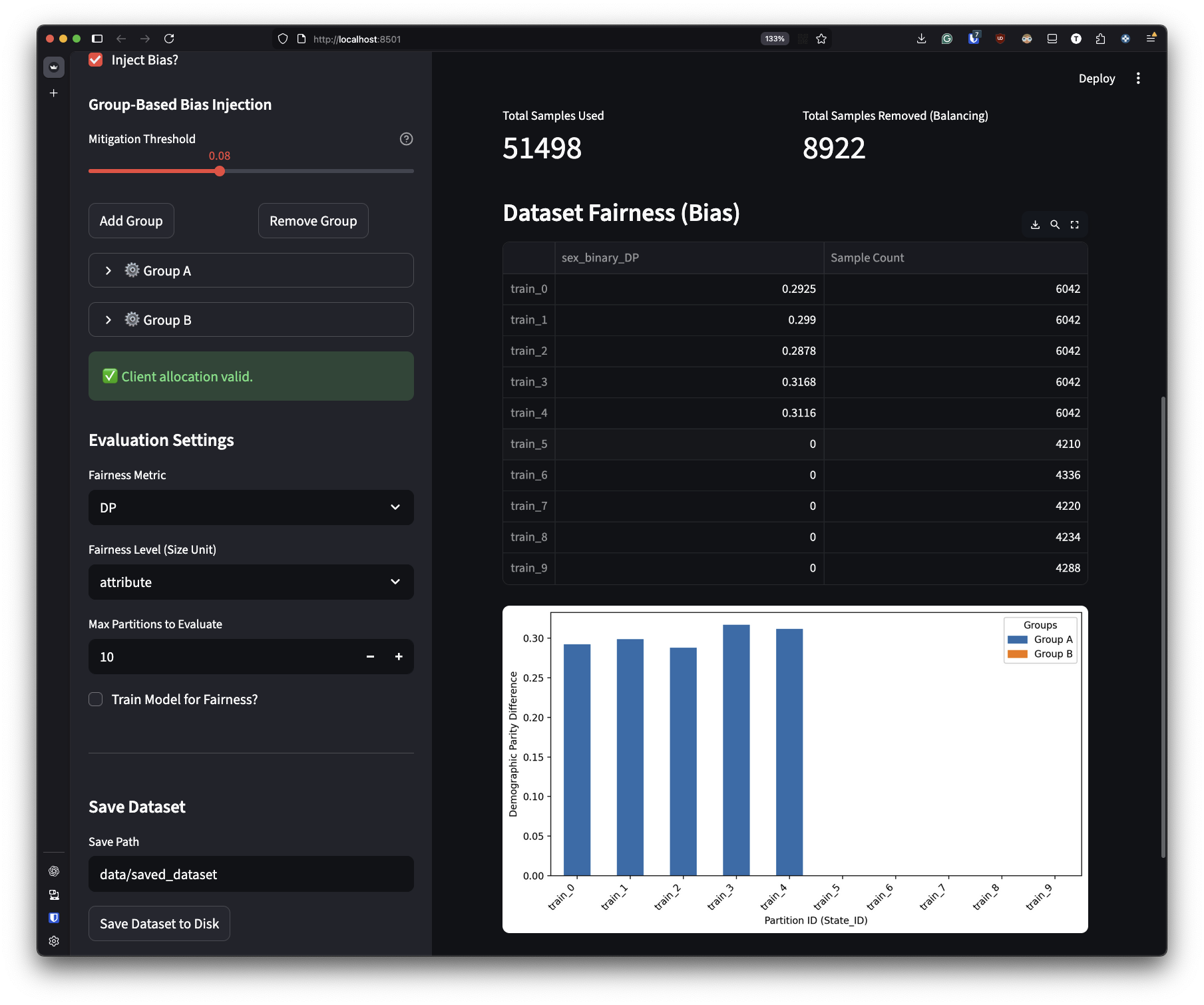}
\caption{Our dashboard includes the ability to plot the distribution of the fairness metrics of different dataset splits.}
\label{fig:dashboard_2}
\end{figure}

As you can see in Figure~\ref{fig:dashboard_1}, the dashboard allows practitioners to select a dataset (either from one of the locally supported datasets or one from the Hugging Face Hub) and split it, simulating the assignment of each partition to a separate FL client. 
It is also possible to define configurations to inject bias for certain groups of clients or for all participants. Moreover, our dashboard includes the ability to plot the unfairness of the different clients, measuring it with both Demographic Disparity and Equalized Odds Difference. As shown in Figure~\ref{fig:dashboard_2}, this is useful for an easy visualization of the configurations defined for the bias injection.

\section{Reproducibility of the Benchmark Suite}\label{app:fixed_datasets}

Alongside the \fairdataset library, we publish twelve datasets covering different bias scenarios to explore the behavior of fair FL methods. As discussed in Section~\ref{sec:fixed_datasets}, we exploited three different datasets in the benchmark suite. These datasets can be divided into four modalities: 
\begin{itemize}
    \item[(I)] \textbf{Attribute-silo} datasets: \att dataset for the cross-silo setting; 
     \item[(II)]  \textbf{Value-silo} datasets: \val dataset for the cross-silo setting; 
      \item[(III)] \textbf{Attribute-device} datasets: \att dataset for the cross-device setting; 
       \item[(IV)] \textbf{Value-device} datasets: \val dataset for the cross-device setting. 
\end{itemize}

\begin{figure*}[h!]
    \centering
    \begin{minipage}[b]{0.48\textwidth}
        \centering
        \subfloat[Cross-Silo]{
            \includegraphics[width=0.45\linewidth]{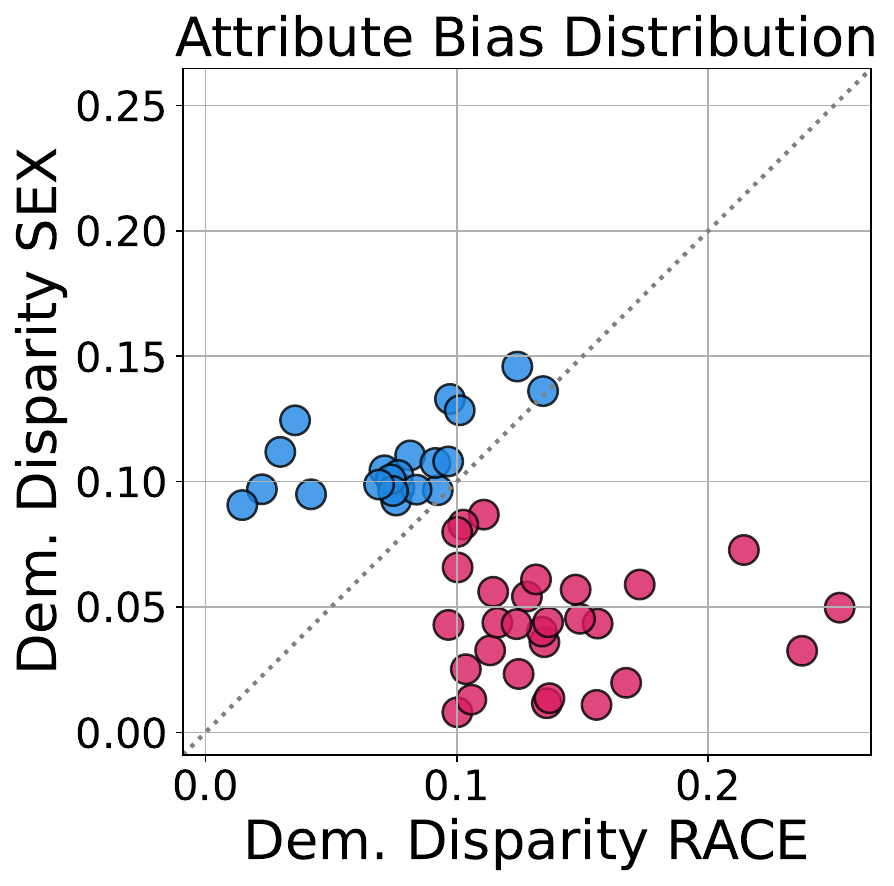}
            \label{fig:appendix_fd_cs_att_income}
        }
        \hfill
        \subfloat[Cross-Device]{
            \includegraphics[width=0.45\linewidth]{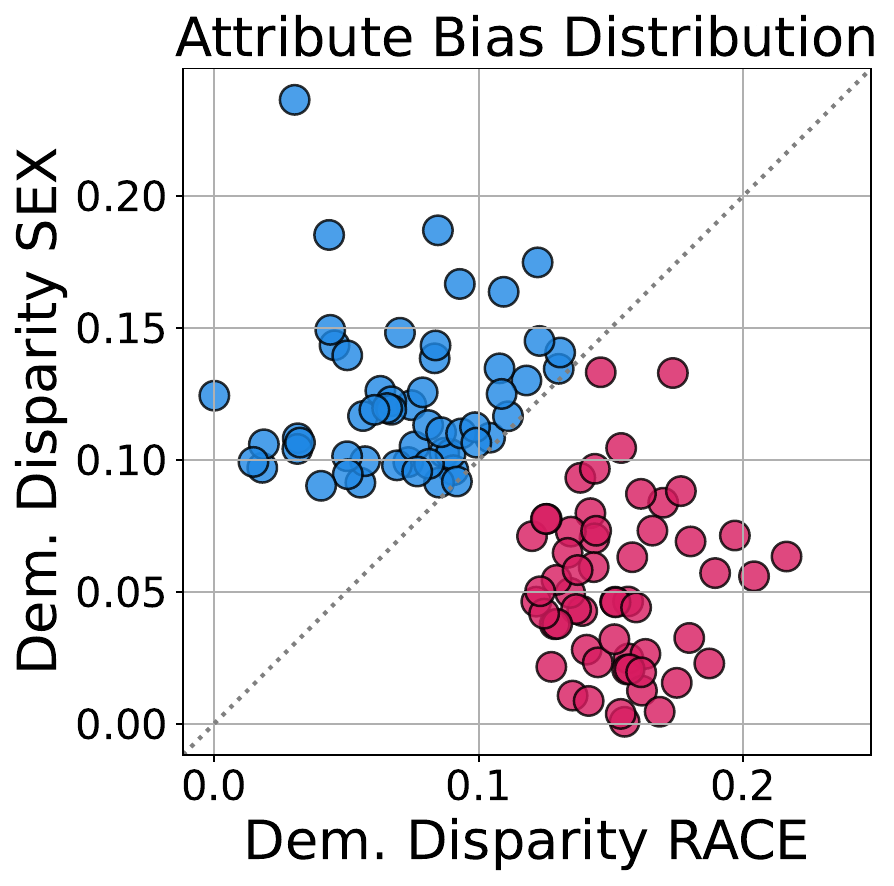}
            \label{fig:appendix_fd_cd_att_income}
        }
        \par\vspace{2mm}
        \includegraphics[width=0.8\linewidth]{images/legend/legend_blue_red.pdf}
        
        \caption{\att measured with DD on the local Logistic Regression models for attribute benchmark datasets using \income.}
        \label{fig:appendix_lr_attribute_income}
    \end{minipage}
    \hfill 
    \begin{minipage}[b]{0.48\textwidth}
        \centering
        \subfloat[Cross-Silo]{
            \includegraphics[width=0.45\linewidth]{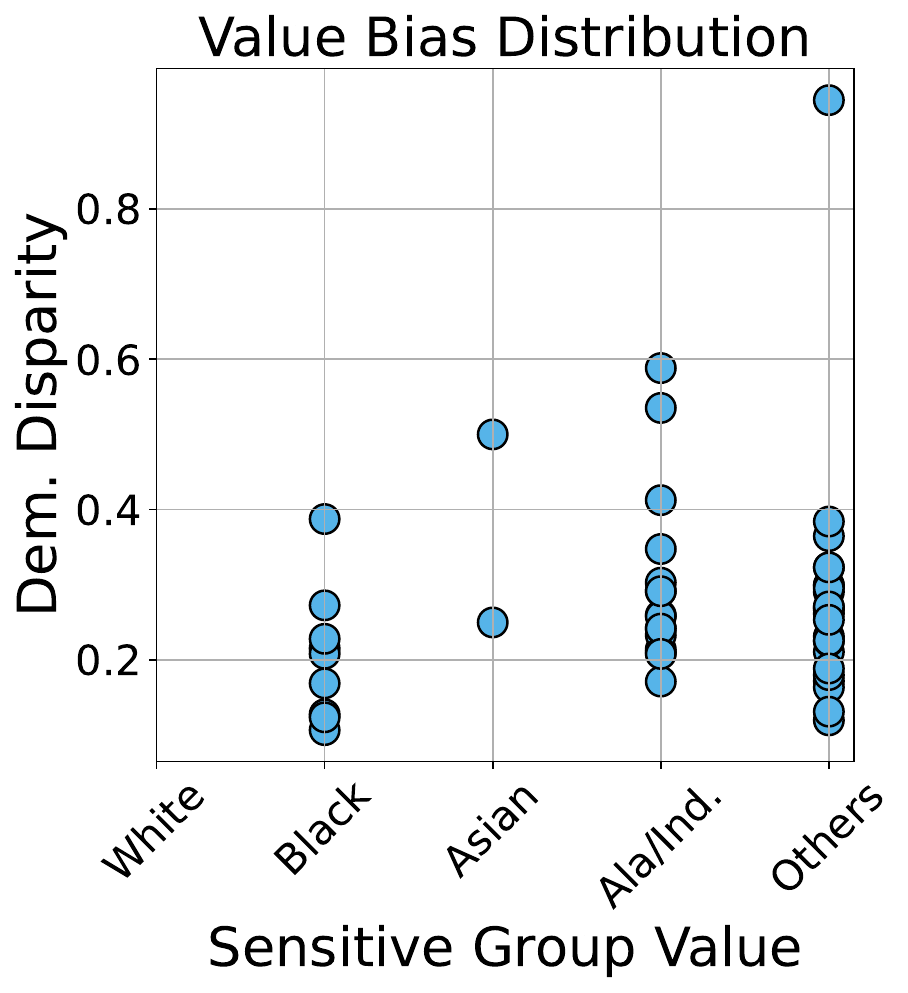}
            \label{fig:appendix_fd_cs_val_income}
        }
        \hfill
        \subfloat[Cross-Device]{
            \includegraphics[width=0.45\linewidth]{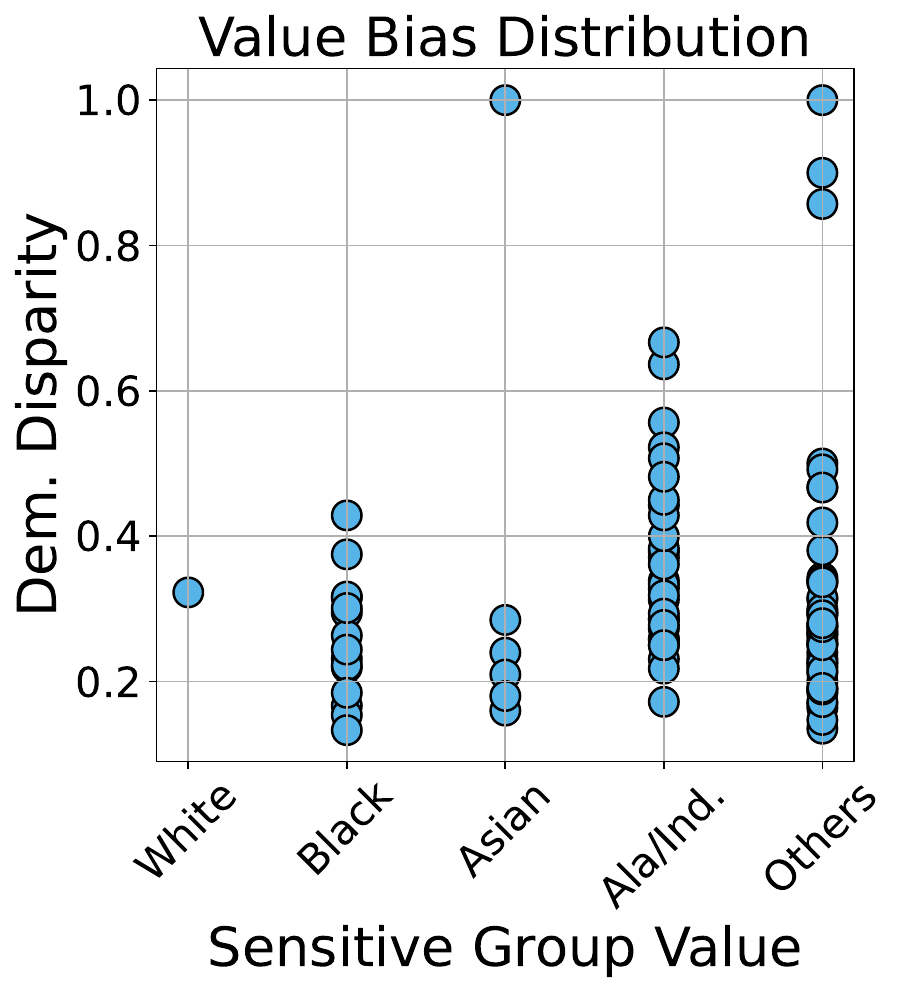}
            \label{fig:appendix_fd_cd_val_income}
        }
        \par\vspace{2mm}
        
        \caption{\val measured with DD on the local Logistic Regression models for value benchmark datasets using \income.}
        \label{fig:appendix_value_example_income}
    \end{minipage}
\end{figure*}

For the \income dataset, we provide the concrete modifications applied for bias exacerbation in Table ~\ref{tab:fixds_att_bias_modific} for datasets (I) and (III) and in Table~\ref{tab:fixds_value_bias_modific} for datasets (II) and (IV). Furthermore, for each of these four datasets, we list the counts and states sorted by where they take on the maximal bias in Table~\ref{tab:fixds_cs_att_bias_dis},~\ref{tab:fixds_cs_value_bias_dis},~\ref{tab:fixds_cd_att_bias_dis}, and~\ref{tab:fixds_cd_value_bias_dis} respectively. Additionally, in Figures~\ref{fig:appendix_lr_attribute_income} and~\ref{fig:appendix_value_example_income}, we report the Demographic Disparity distribution for each of the four datasets on Logistic Regression models.

For the \dutch dataset, we split the clients into two groups (see Section~\ref{sec:dutchdatasets}), and we defined the bias injection settings for each of these groups. In this case, instead of setting a $P_{drop}$ and $P_{flip}$ for each client, we set the parameters $\mu$ and $\sigma$ of the Gaussian Distribution for each group we define. Then, from this distribution, we draw the $P_{drop}$ and $P_{flip}$ for each client $k$ of the group. This allows practitioners to easily set up the dataset creation, especially when the dataset is split among hundreds of clients. All configurations utilize for $P_{drop}$ the configuration $\mu=0.5, \sigma=0.05$ and for $P_{flip}$ we set $\mu=0.4,\sigma=0.02$. Specifically, in the \att experiment, we considered the sensitive attributes \sex and \mar and fixed $\mu$ and $\sigma$ for both of them. In the \val experiment, instead, we considered the \mar sensitive attribute with its values \texttt{Never Married}, \texttt{Married}, \texttt{Divorced}, and \texttt{Widowed}, fixing $\mu$ and $\sigma$ for these groups.

For the \celeba dataset, we used the same approach we adopted with \dutch. We report in Table~\ref{tab:celeba_settings} the bias injection configurations used for the \att and \val datasets. Specifically, we set the parameters of $P_{drop}$ to $\mu=0.3, \sigma=0.05$ and for $P_{flip}$ we set $\sigma=0.02$ while $\mu$ varies (see Table ~\ref{tab:celeba_settings}).

\section{Additional experiments} \label{app:additional_ex}
In line with Section~\ref{sec:experiments}, we provide additional experiments in settings we did not include in the main paper. 
In Section~\ref{app:additional_ex_income} we present the results with the \income dataset. Specifically, we present results obtained with all cross-device datasets, as well as with Logistic Regression as an additional local model.
In Section~\ref{app:experiments_dutch} we present the additional results obtained with the \dutch dataset.
Lastly, in Section~\ref{app:celeba} we present the results obtained with the \celeba dataset.

\subsection{Experiments with \income}\label{app:additional_ex_income}

\subsubsection{Additional \att benchmark experiments} \label{app:experiments_attribute}
\begin{figure}
  \subfloat[\textbf{Attribute-device:} XGB vs. FedAvg]{\includegraphics[width=0.16\textwidth]{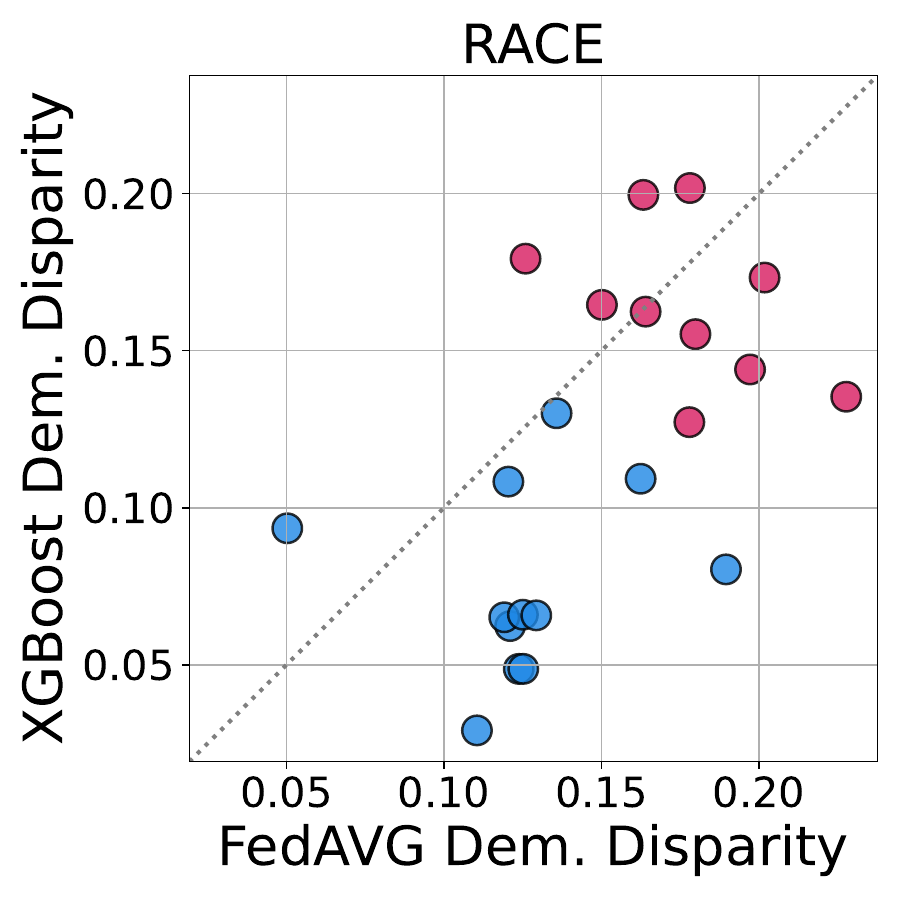}\includegraphics[width=0.16\textwidth]{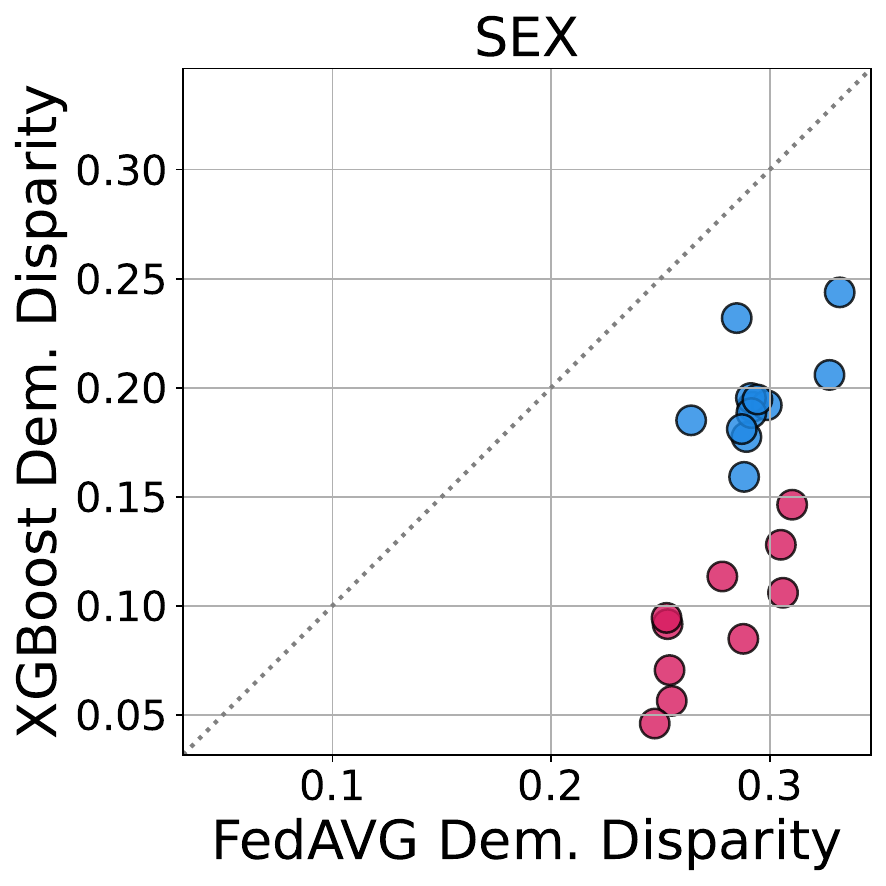}\label{fig:before_after_attribute_device_xgb}}
  \subfloat[\textbf{Attribute-device:} XGB vs. PUFFLE]{\includegraphics[width=0.16\textwidth]{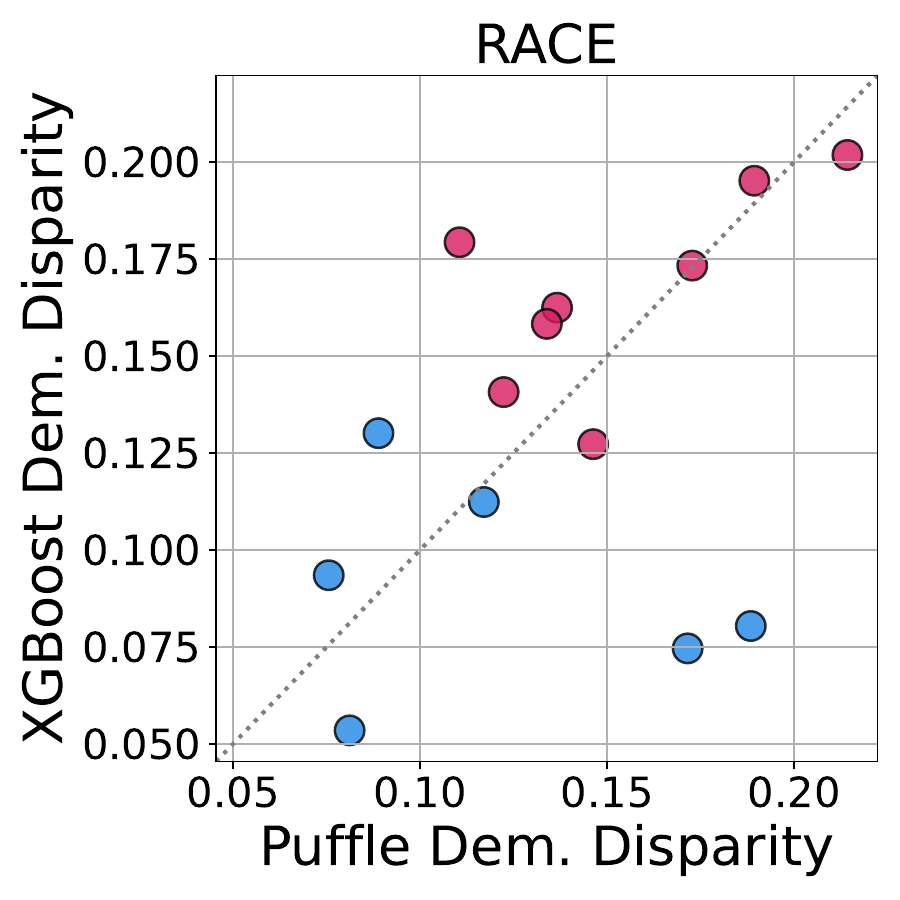}\includegraphics[width=0.16\textwidth]{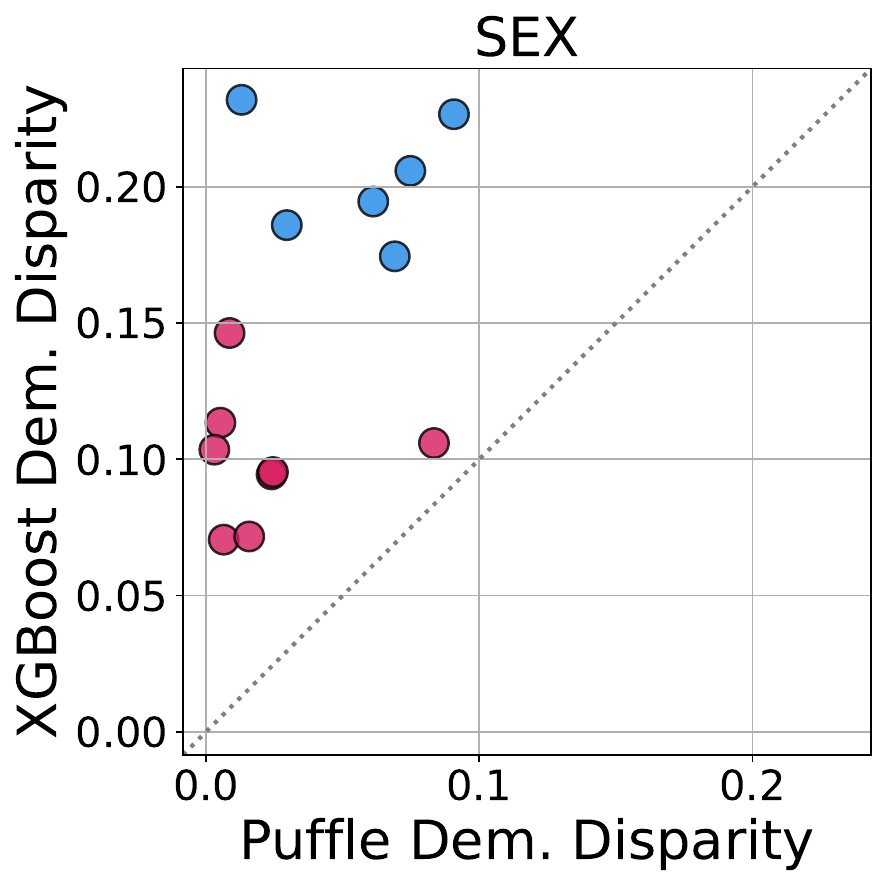}\label{fig:before_after_attribute_device_xgb_puffle}}
  \subfloat[\textbf{Attribute-device:} XGB vs. Reweighing]{\includegraphics[width=0.16\textwidth]{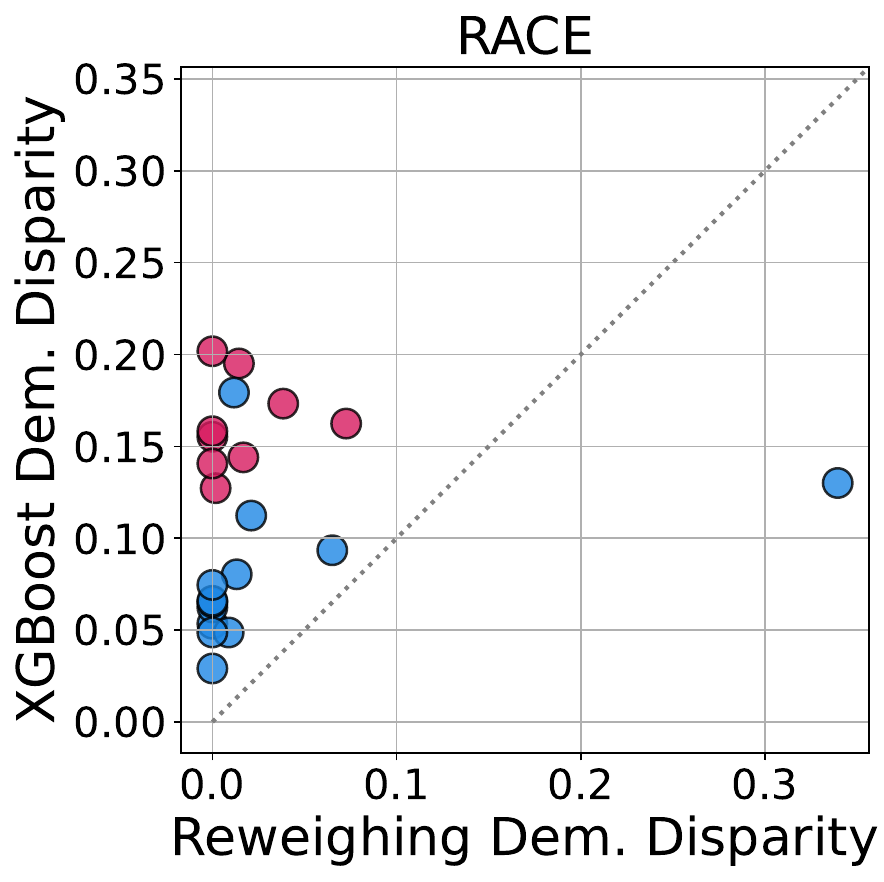}\includegraphics[width=0.16\textwidth]{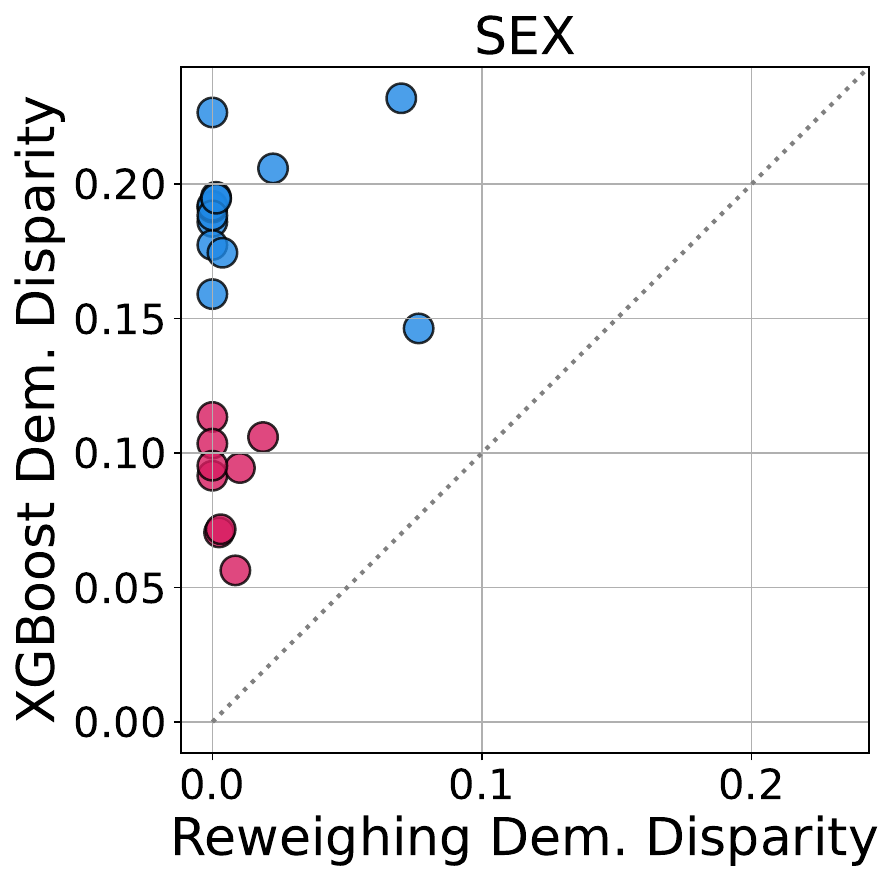}\label{fig:before_after_attribute_device_xgb_reweighing}}\\
    \centering\includegraphics[width=0.40\textwidth]{images/legend/legend_blue_red.pdf}
  \caption{\att toward \race and \sex measured with Demographic Disparity on the local XGBoost models versus the FedAvg model, the PUFFLE model, and the Reweighing model for the attribute-device \income dataset.}
  \label{fig:before_after_attribute_device}
\end{figure}

\begin{figure}
  \subfloat[\textbf{Attribute-device:} LR vs. FedAvg]{\includegraphics[width=0.16\textwidth]{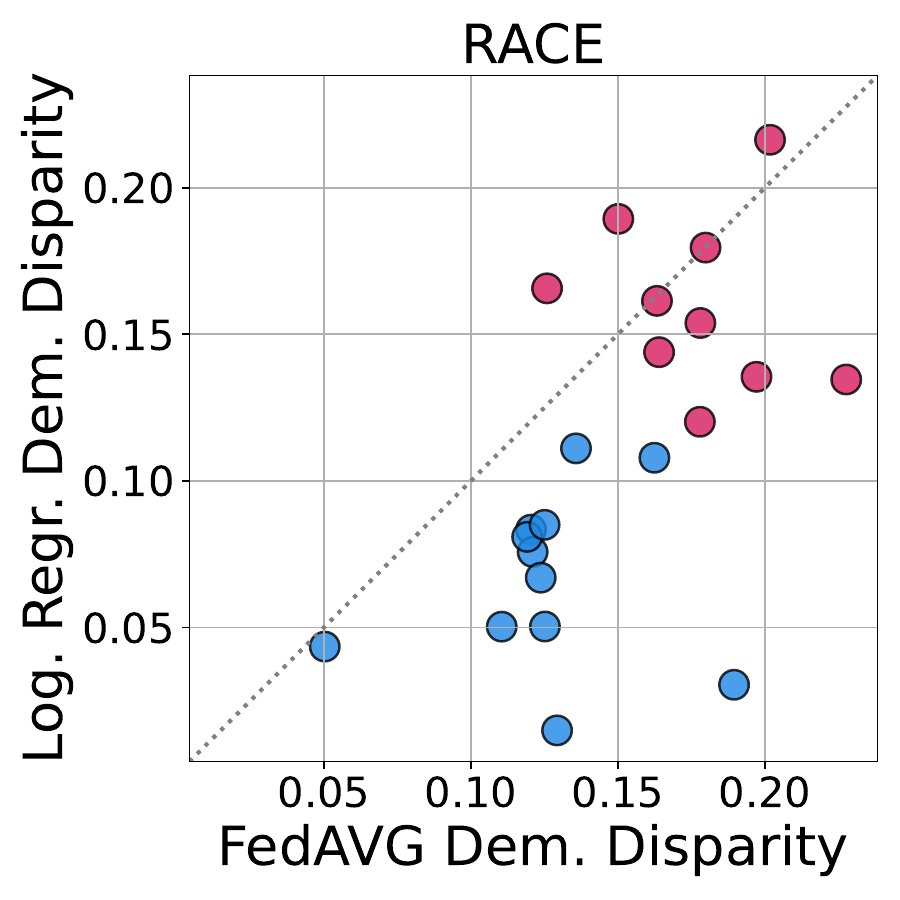}\includegraphics[width=0.16\textwidth]{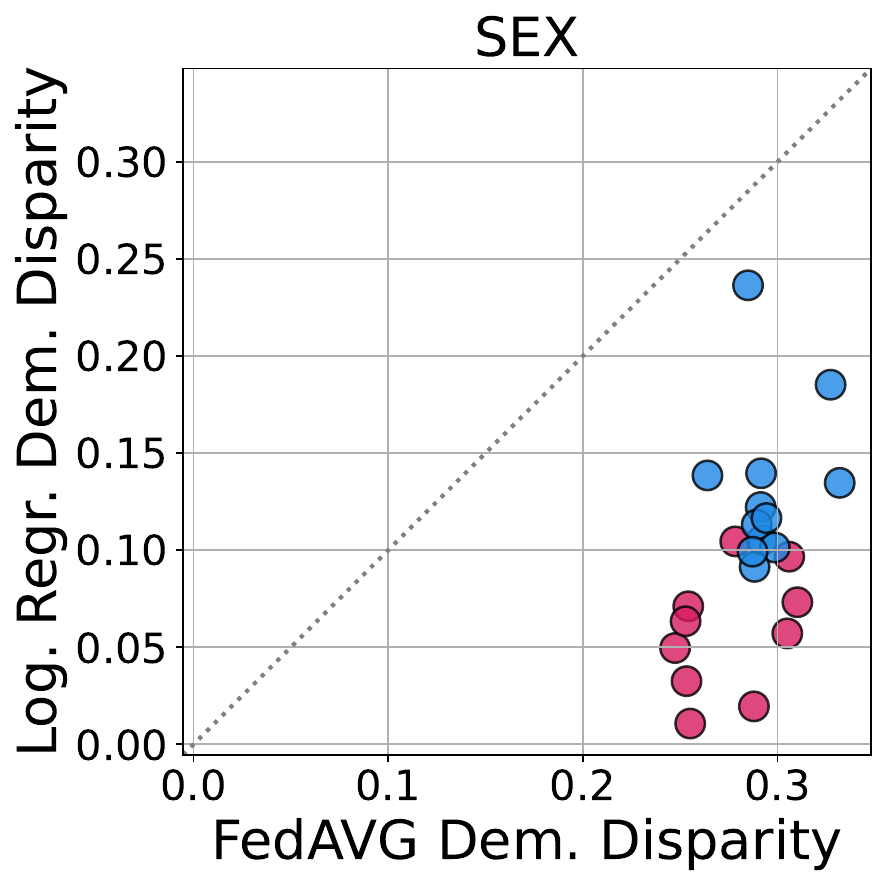}\label{fig:before_after_attribute_device_logistic}}
  \subfloat[\textbf{Attribute-device:} LR vs. PUFFLE ]{\includegraphics[width=0.16\textwidth]{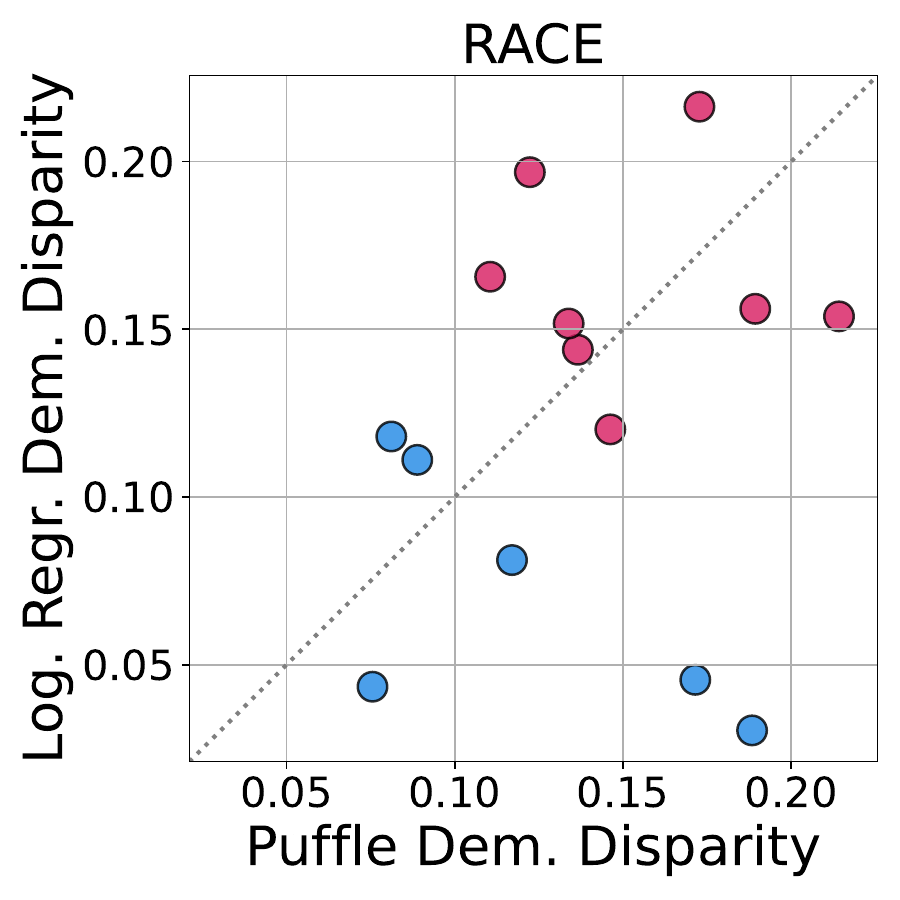}\includegraphics[width=0.16\textwidth]{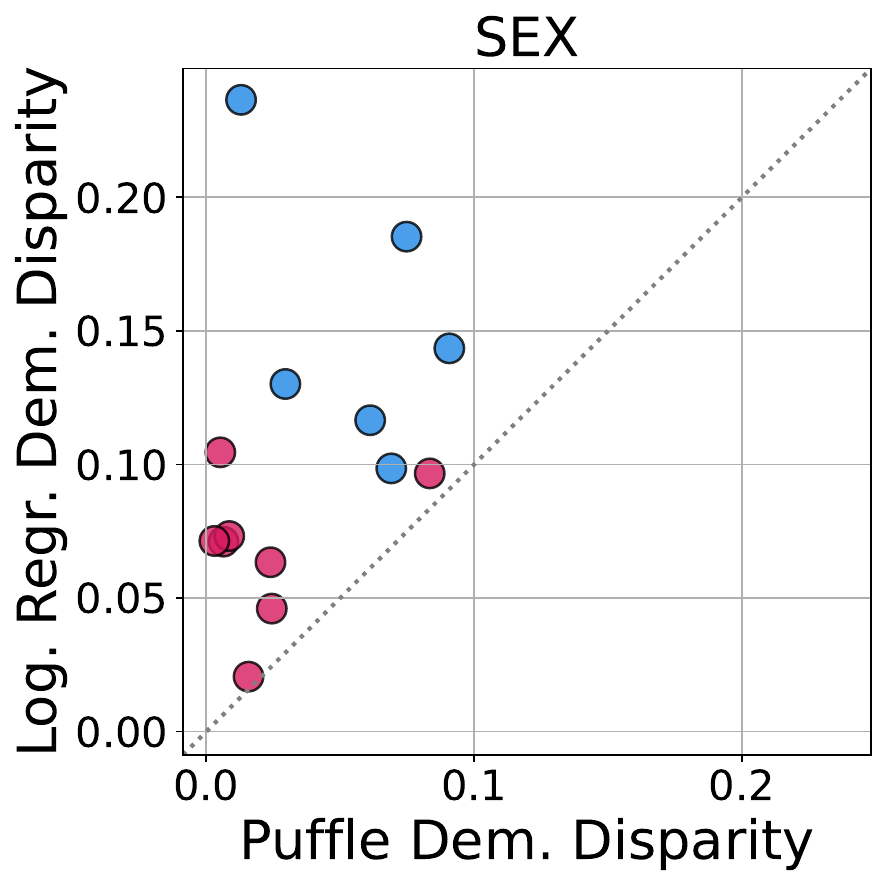}\label{fig:before_after_attribute_device_lr_puffle}}
  \subfloat[\textbf{Attribute-device:} LR vs. Reweighing]{\includegraphics[width=0.16\textwidth]{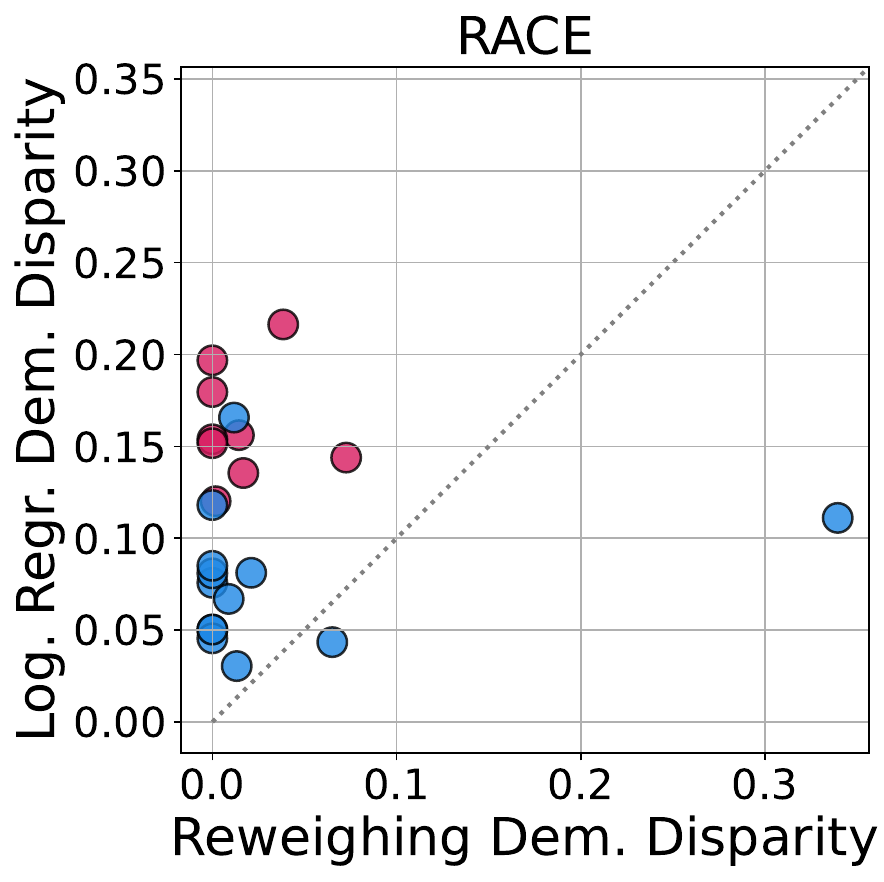}\includegraphics[width=0.16\textwidth]{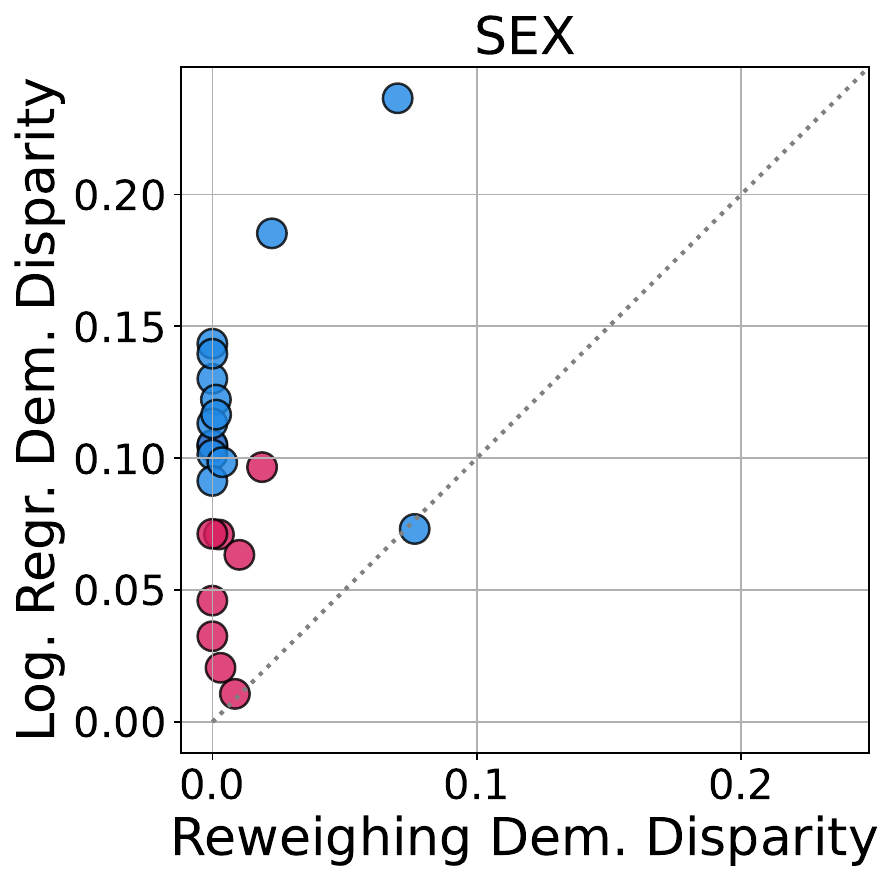}\label{fig:before_after_attribute_device_lr_reweighing}}\\
    \centering\includegraphics[width=0.40\textwidth]{images/legend/legend_blue_red.pdf}
  \caption{\att toward \race and \sex measured with Demographic Disparity on the local Logistic Regression (LR) models versus the FedAvg model, the PUFFLE model, and the Reweighing model for the attribute-device \income dataset.}
  \label{fig:before_after_attribute_device_lr}
\end{figure}

Group-level results for the attribute-device dataset are shown in Figures~\ref{fig:before_after_attribute_device} and~\ref{fig:before_after_attribute_device_lr}. Each point represents a dataset, plotting its Demographic Disparity (DD) value with local versus the three FL models (FedAVG, PUFFLE, Reweighing). 
Points below the diagonal indicate increased DD of the global models, while those above the diagonal show higher DD for the local model. For PUFFLE and Reweighing, fairness was constrained for the \sex attribute. 
Figures~\ref{fig:before_after_attribute_device_xgb} and~\ref{fig:before_after_attribute_device_logistic} demonstrate that FedAvg training often propagates bias, with DD increasing for both \race and \sex across most datasets. 
In contrast, Figures~\ref{fig:before_after_attribute_device_xgb_puffle} and~\ref{fig:before_after_attribute_device_lr_puffle} reveal that PUFFLE’s fairness constraints on \sex lead to reductions in \sex disparity across all clients. While some clients experience an increase in \race disparity with PUFFLE, it’s unclear whether there exists a specific client group that benefits from improvements across both attributes. PUFFLE’s performance on the \race attribute is comparable to FedAvg training without fairness constraints.
For Reweighing, Figures~\ref{fig:before_after_attribute_device_xgb_reweighing} and~\ref{fig:before_after_attribute_device_lr_reweighing} reveal that Reweighing successfully leads to reductions in \sex disparity across all clients. While for a few clients, \race disparity increases, the majority experience reduced \race disparity. 

\subsubsection{Additional \val benchmark experiments} \label{app:experiments_value}

\begin{figure}   

\subfloat[\textbf{Value-device:} XGB vs. FedAvg]{\includegraphics[width=0.44\textwidth]{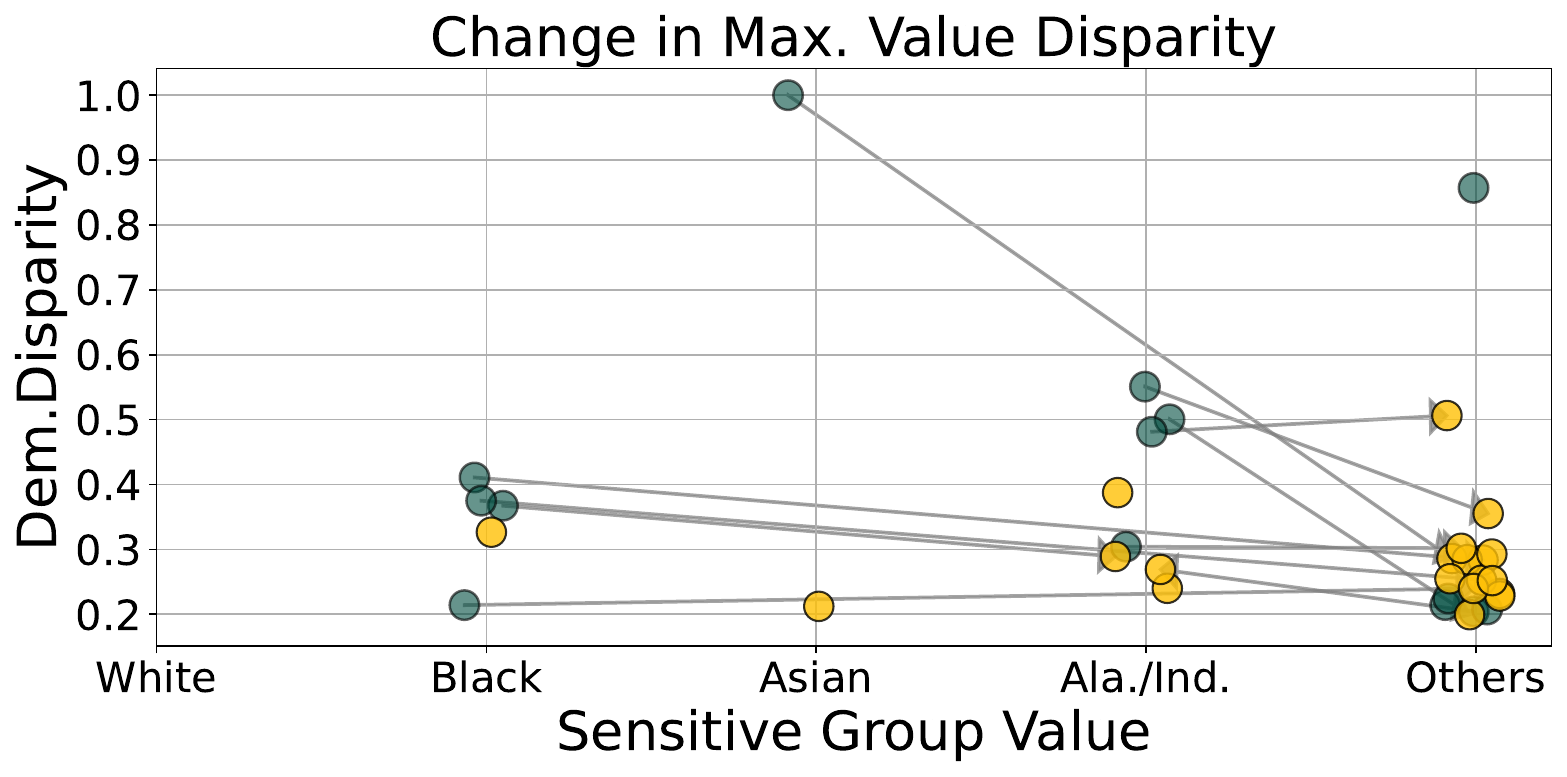}\label{fig:before_after_cross_device_value_xgb}}
  \subfloat[\textbf{Value-device:} XGB vs. PUFFLE]{\includegraphics[width=0.44\textwidth]{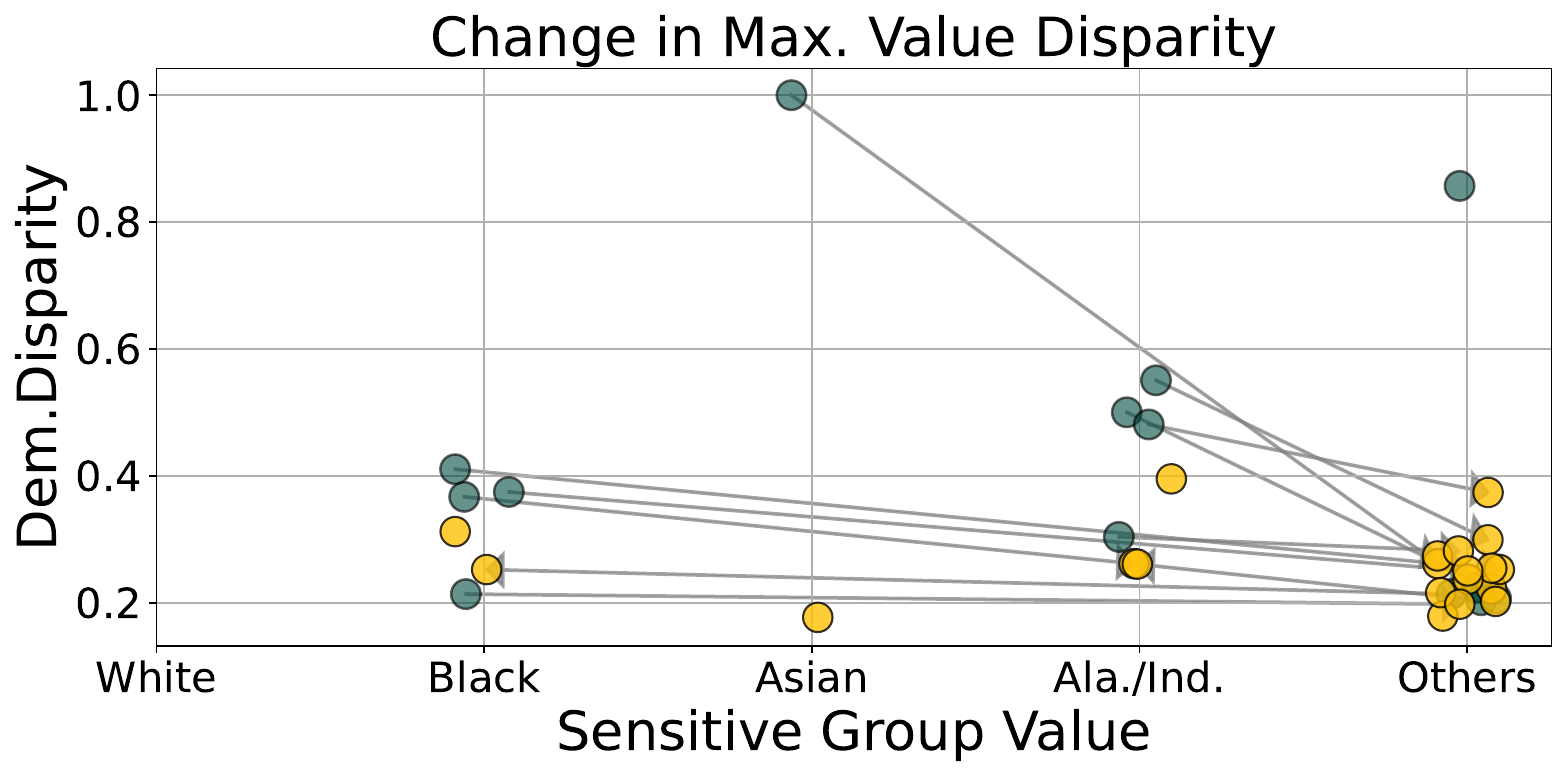}}\\
  \subfloat[\textbf{Value-device:} XGB vs. Reweighing]{\includegraphics[width=0.44\textwidth]{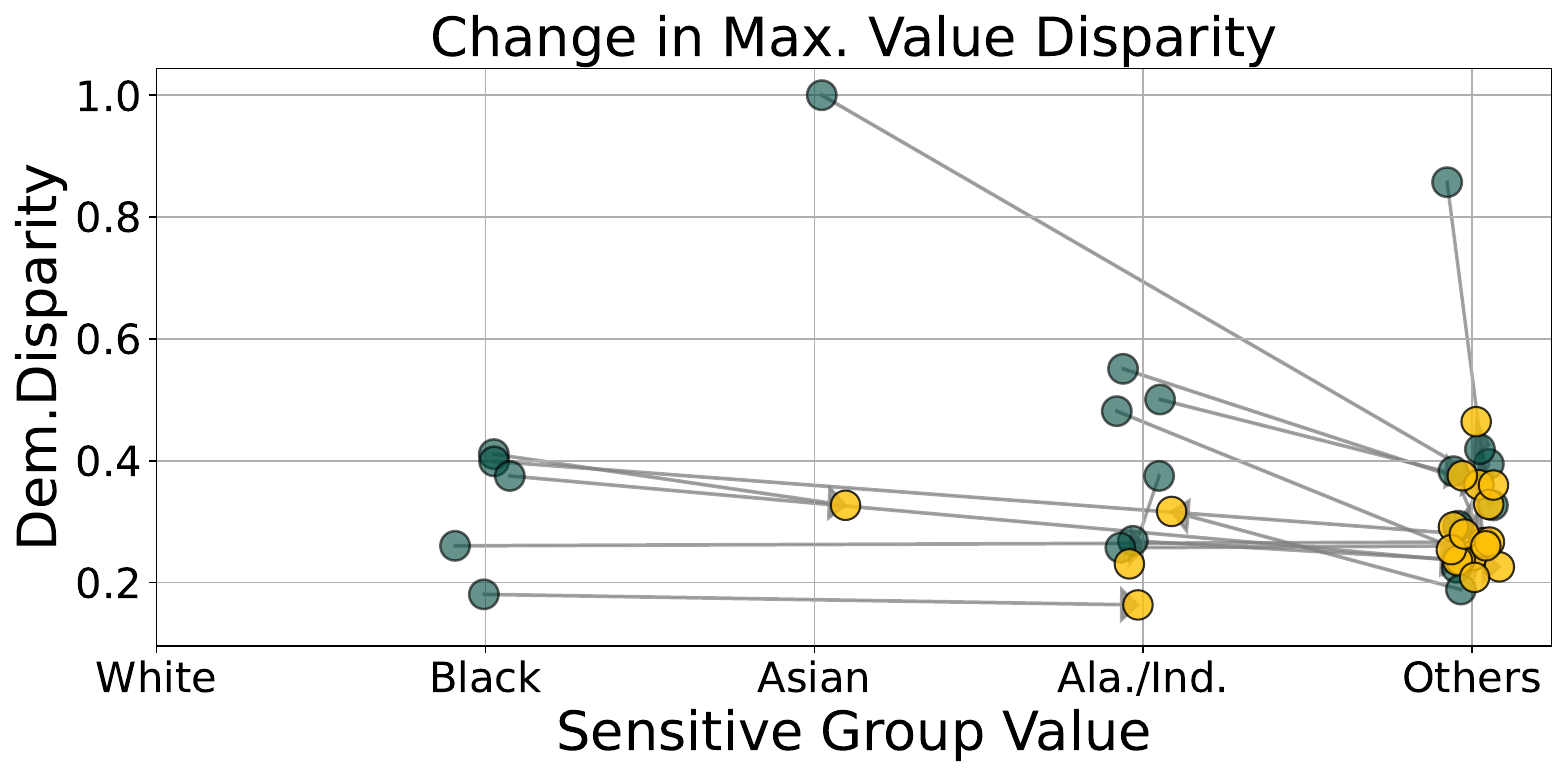}}\\
  \centering\includegraphics[width=0.35\textwidth]{images/legend/value_change_legend_xgboost.pdf}
  \caption{\val toward \race as well as value changes measured with Demographic Disparity on the local XGBoost models versus the FedAvg model, the PUFFLE model, and the Reweighing model for the value-device \income datasets.}
  \label{fig:before_after_device_value_xgb}
\end{figure}

\begin{figure}   

\subfloat[\textbf{Value-device:} LR vs. FedAvg]{\includegraphics[width=0.44\textwidth]{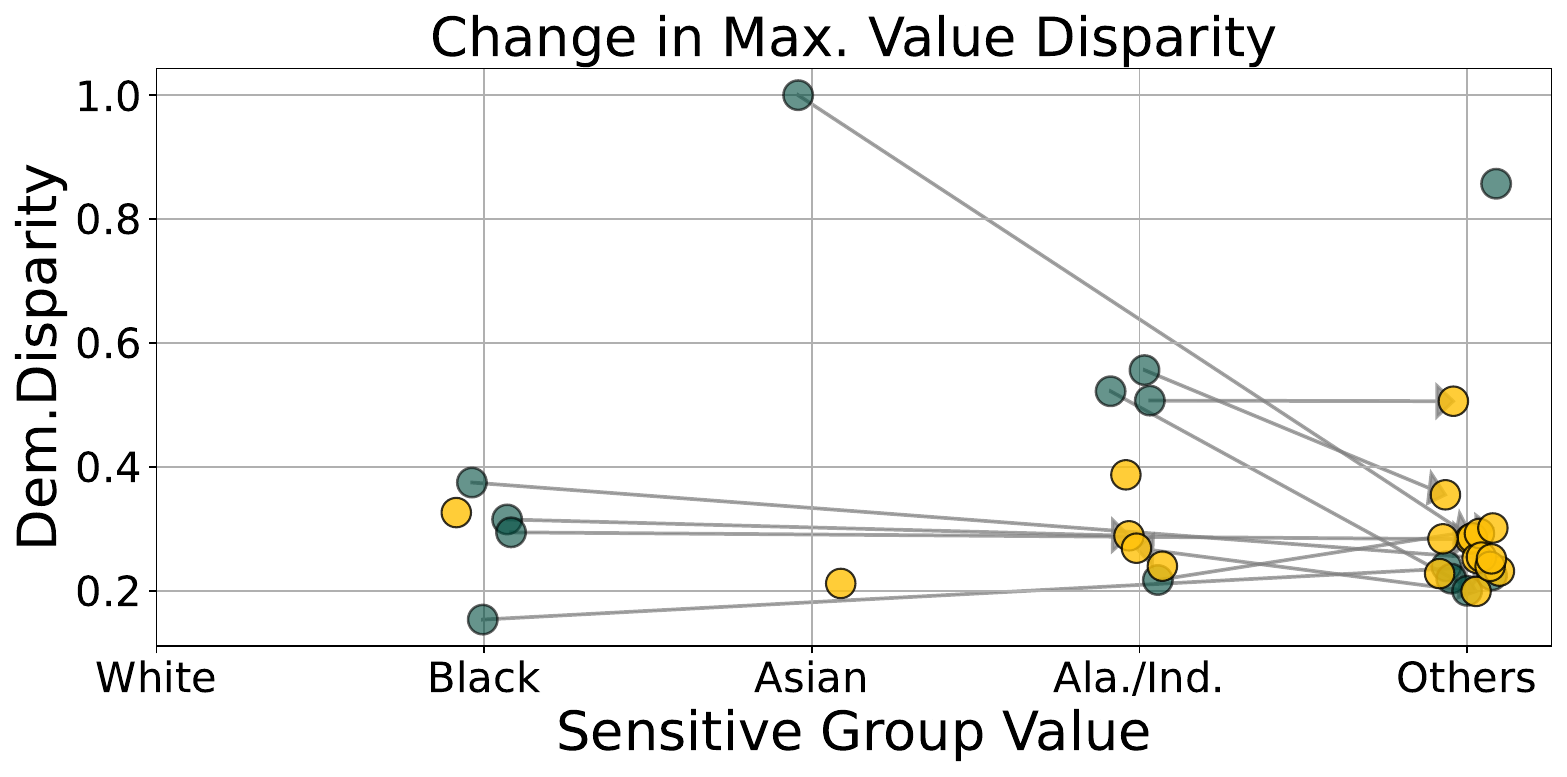}\label{fig:before_after_cross_device_value_lr}}
  \subfloat[\textbf{Value-device:} LR vs. PUFFLE]{\includegraphics[width=0.44\textwidth]{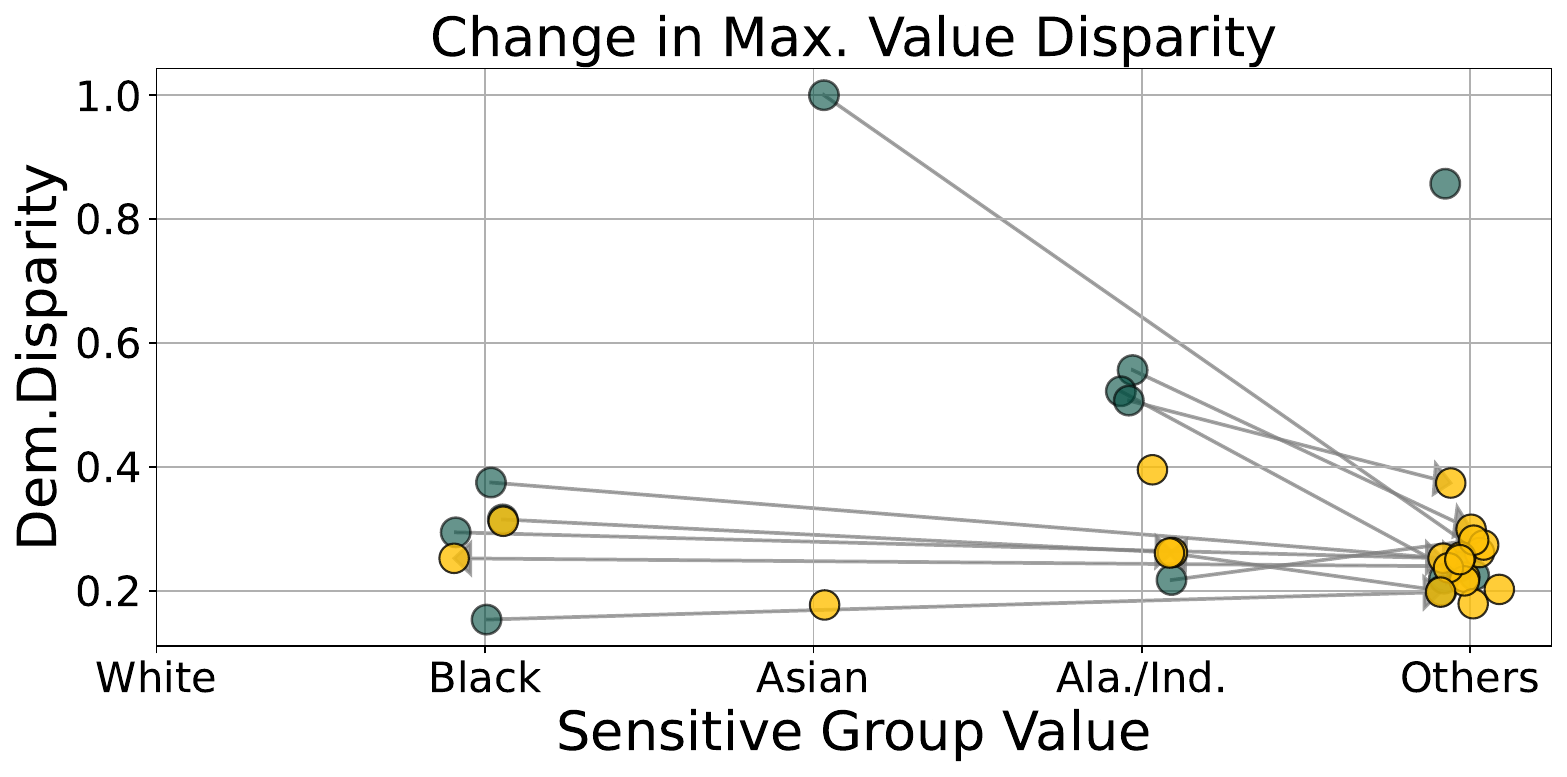}\label{fig:before_after_cross_device_value_lr_puffle}}\\
  \subfloat[\textbf{Value-device:} LR vs. Reweighing]{\includegraphics[width=0.44\textwidth]{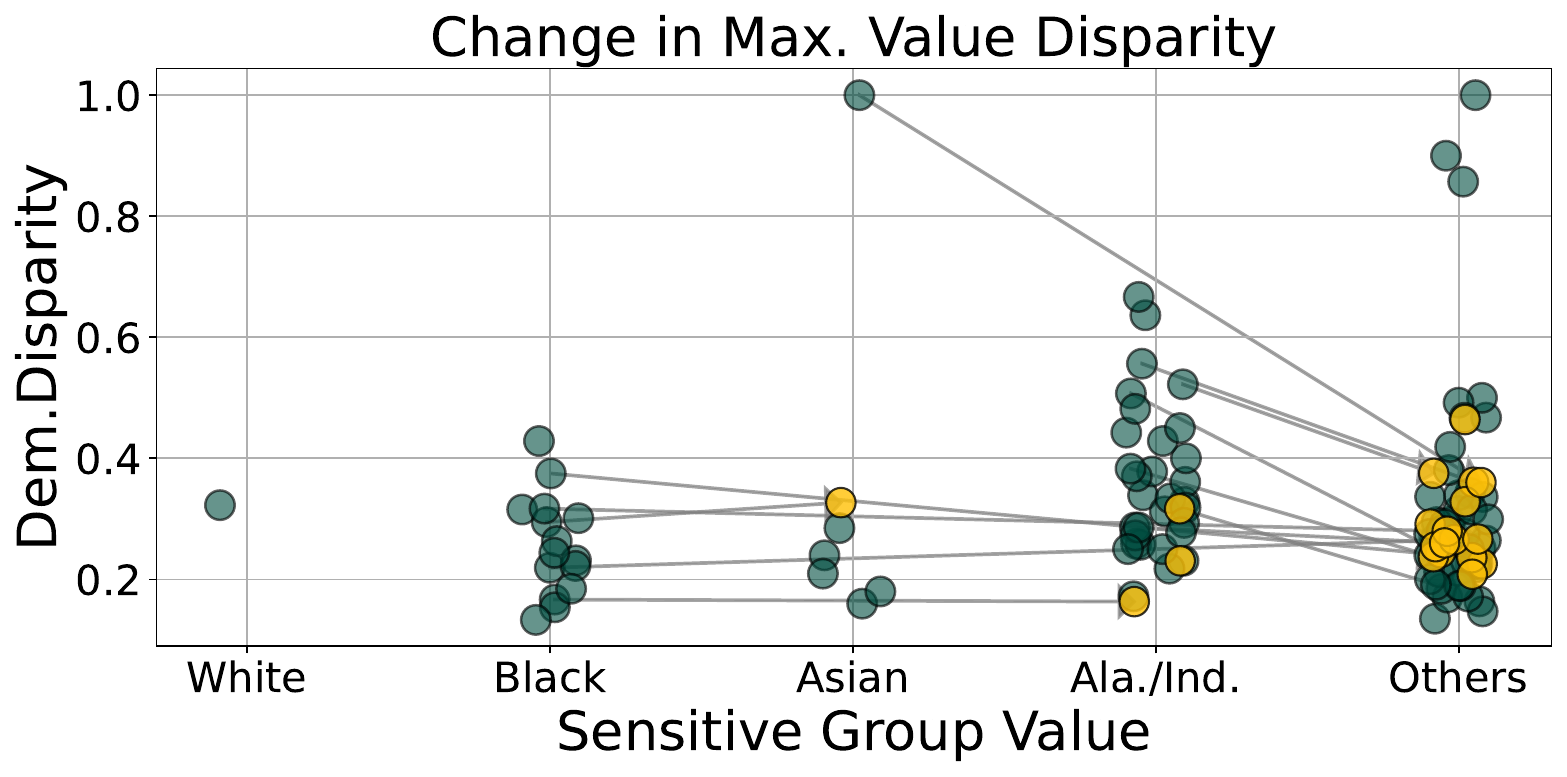}\label{fig:before_after_cross_device_value_lr_rw}}\\
  \centering\includegraphics[width=0.40\textwidth]{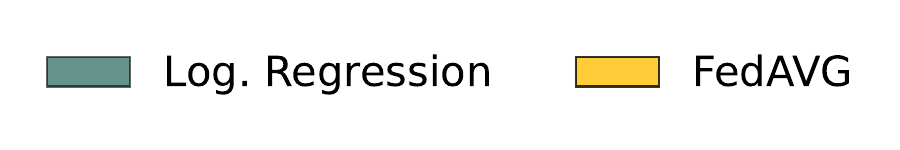}
  \caption{\val toward \race as well as value changes measured with Demographic Disparity on the local Logistic Regression (LR) models versus the FedAvg model, the PUFFLE model, and the Reweighing model for the value-device \income dataset.}
  \label{fig:before_after_value_device_lr}
\end{figure}

For the value-device dataset, Figures~\ref{fig:before_after_device_value_xgb} and~\ref{fig:before_after_value_device_lr} show results obtained when comparing FedAvg and PUFFLE models, respectively, with local XGBoost and local Logistic regression models. In particular, the plots highlight changes in Demographic Disparity (DD) and how the distribution of maximal DD values shifts. The trend indicates that underrepresented groups (here, values \texttt{Alaska Native/American Indian} and \texttt{Others}) experience a greater disadvantage from FL models. Their clusters grow, particularly for value \texttt{Others}, while other clusters, like \texttt{Black}, shrink. 
However, overall \race disparity appears to improve slightly in the visualization. 

\subsection{Experiments with \dutch}
\label{app:experiments_dutch}

As explained in Section~\ref{sec:experiments}, besides the results obtained with the benchmarking dataset \income, we also performed all experiments on \dutch. The settings considered for these experiments are the same; we run experiments on all four \dutch datasets. 

\subsubsection{Additional \att benchmark experiments} 
\begin{figure}
  \subfloat[\textbf{Attribute-silo:} XGB vs. FedAvg]{
  \includegraphics[width=0.16\textwidth]{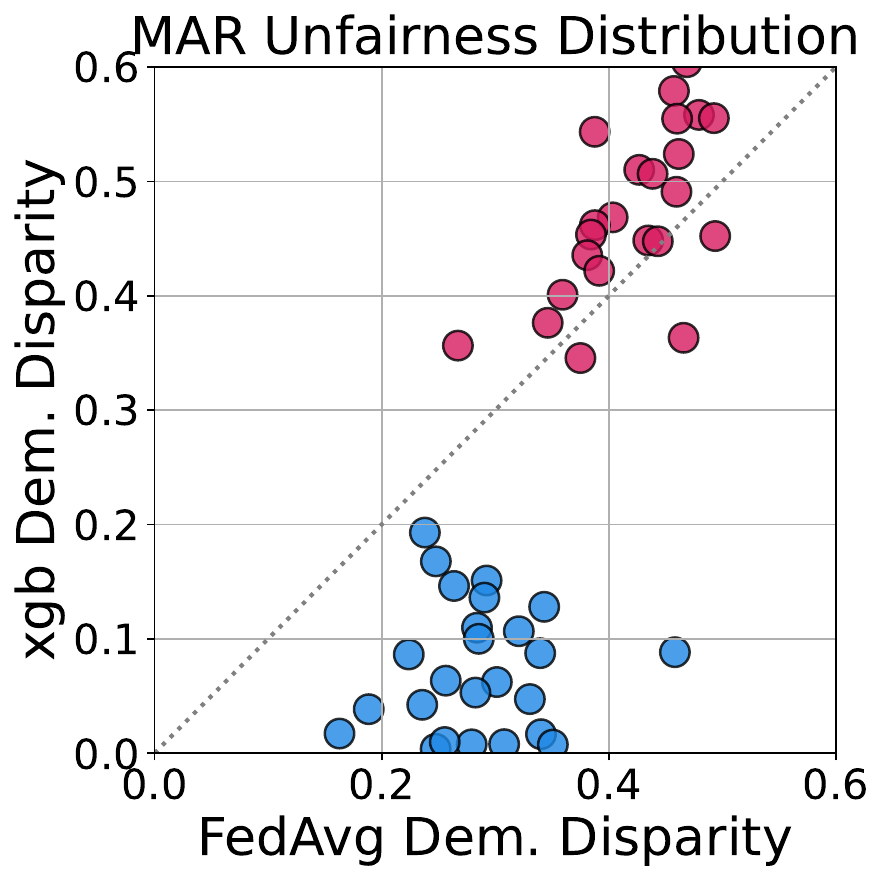}
  \includegraphics[width=0.16\textwidth]{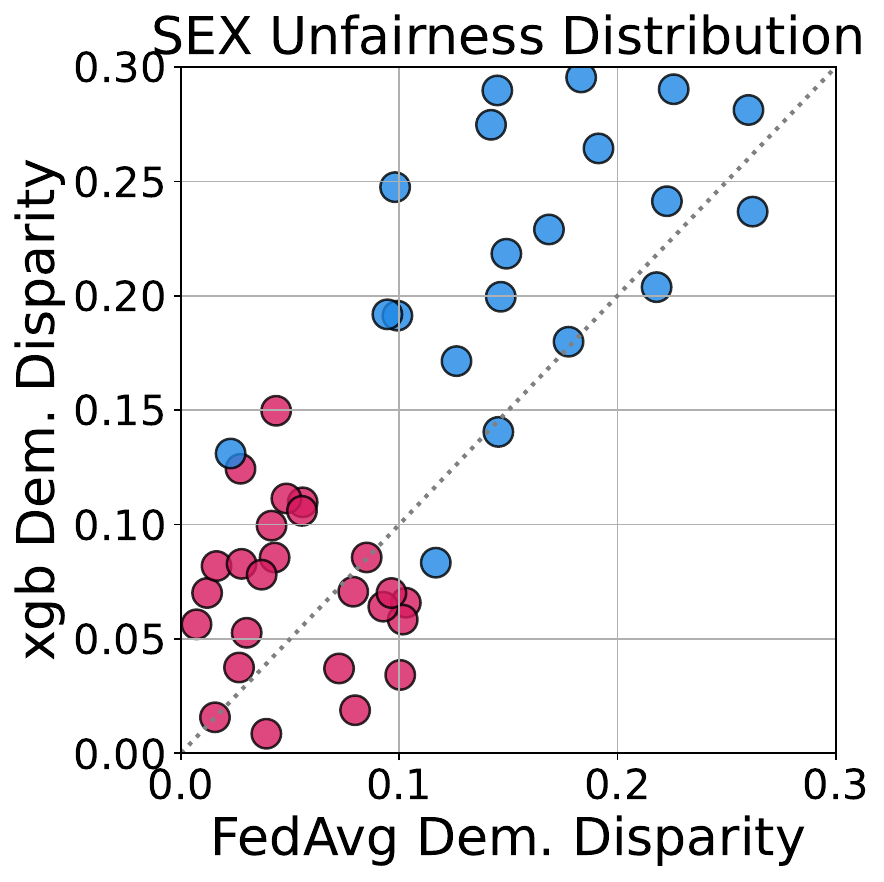}
  \label{fig:dutch_before_after_silo_attribute_xgb}}
  \subfloat[\textbf{Attribute-silo:} XGB vs. PUFFLE]{
  \includegraphics[width=0.16\textwidth]{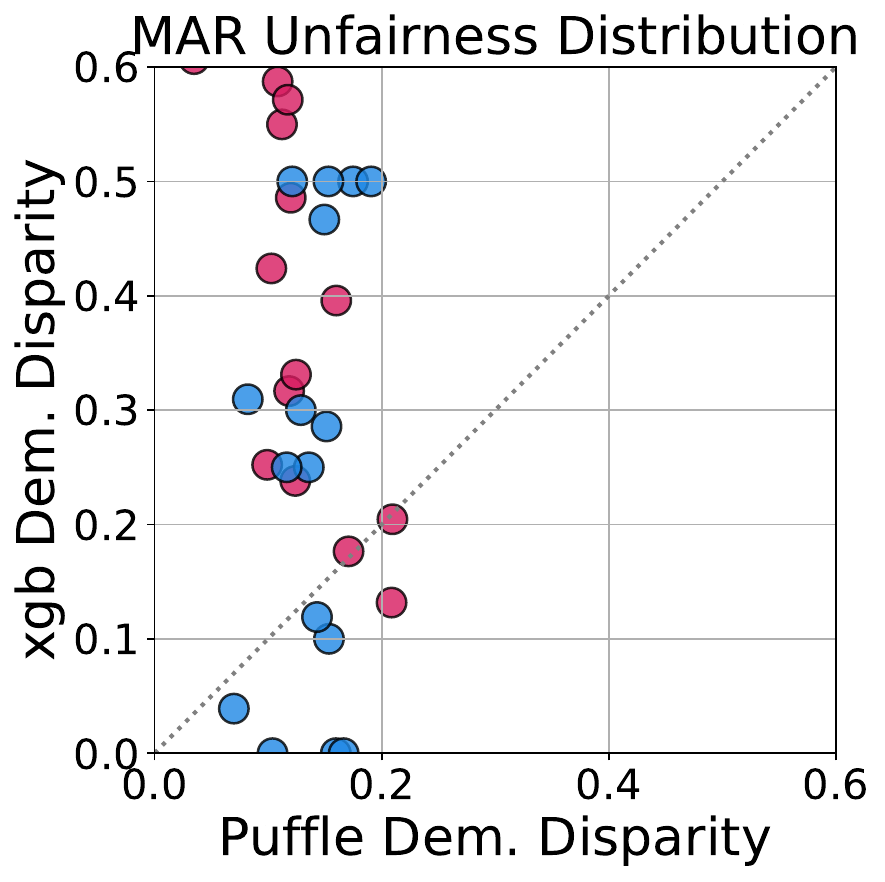}
  \includegraphics[width=0.16\textwidth]{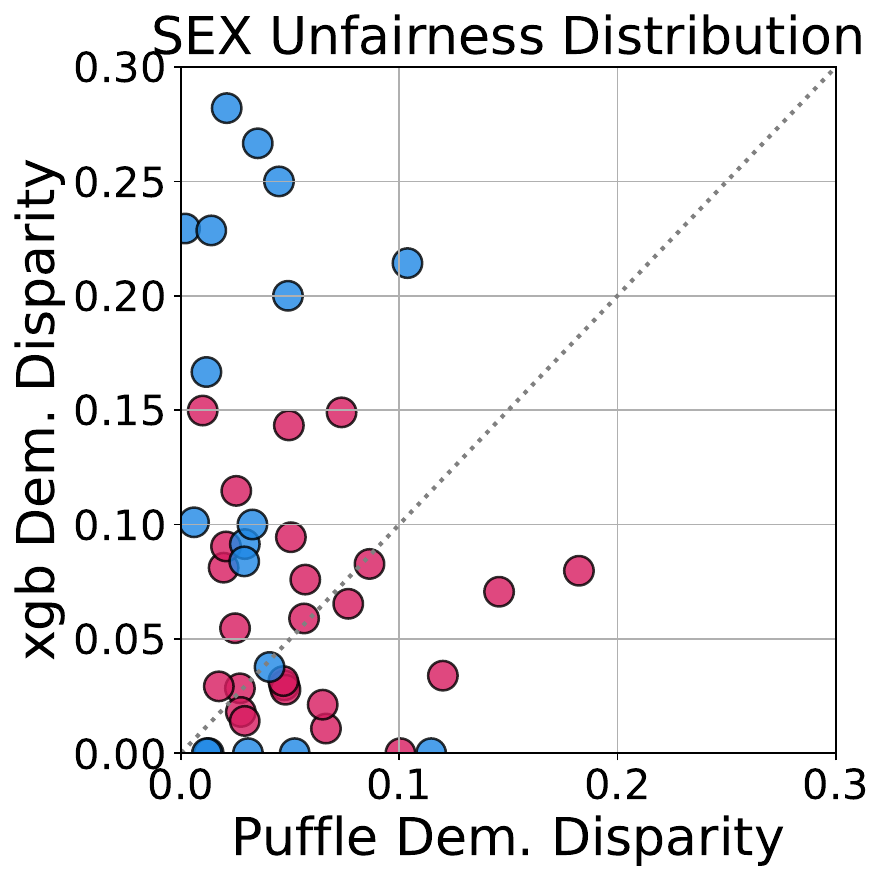}
  \label{fig:dutch_before_after_attribute_silo_xgb_puffle}}
  \subfloat[\textbf{Attribute-silo:} XGB vs. Reweighing]{
  \includegraphics[width=0.16\textwidth]{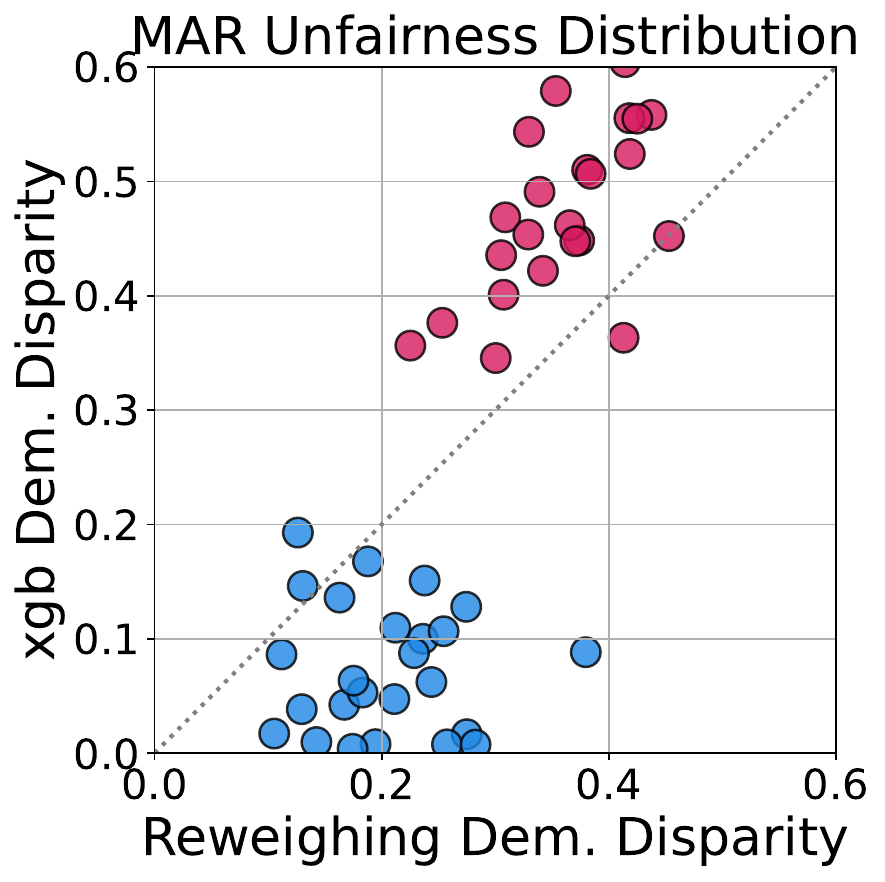}
  \includegraphics[width=0.16\textwidth]{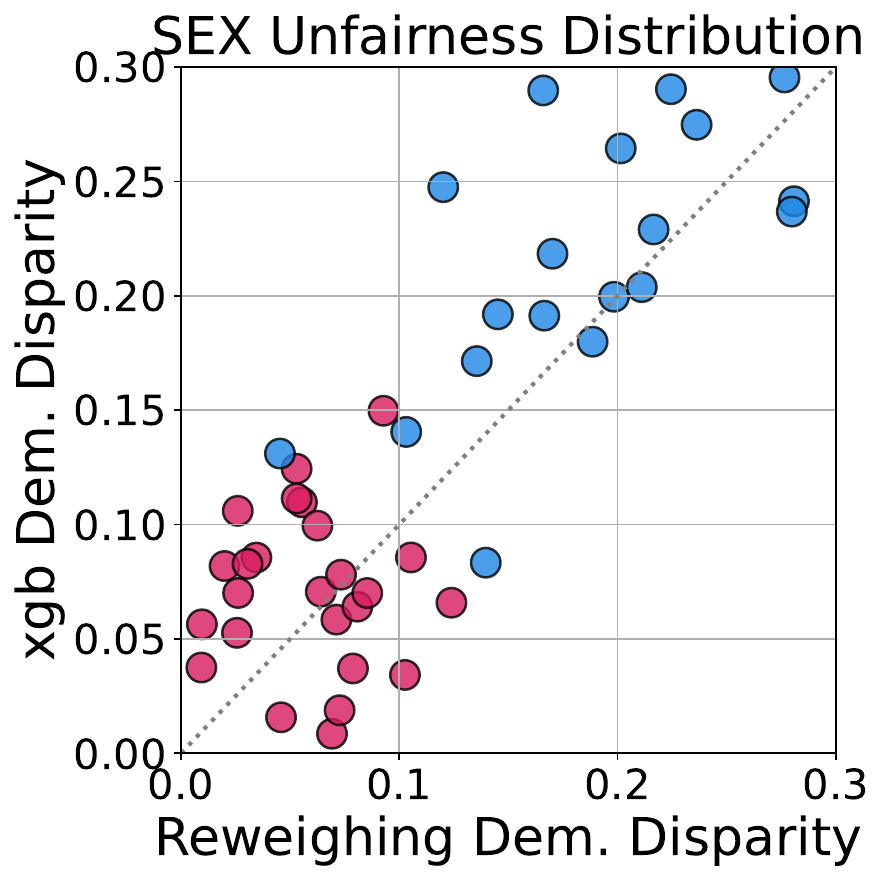}
  \label{fig:dutch_before_after_attribute_silo_xgb_reweighing}}\\
    \centering\includegraphics[width=0.40\textwidth]{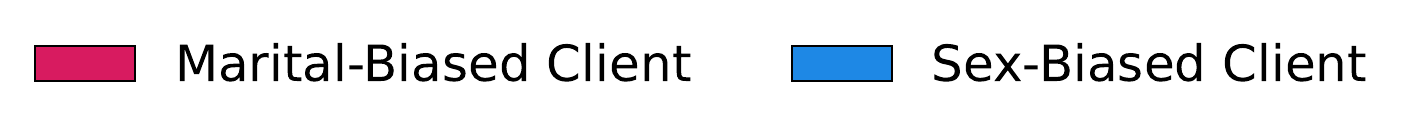}
  \caption{\att toward \mar and \sex measured with Demographic Disparity on the local XGBoost models versus the FedAvg model, the PUFFLE model, and the Reweighing model for the attribute-silo \dutch dataset.}
  \label{fig:dutch_cross_silo_attribute_XGB}
\end{figure}

\begin{figure}
  \subfloat[\textbf{Attribute-silo:} LR vs. FedAvg]{
  \includegraphics[width=0.15\textwidth]{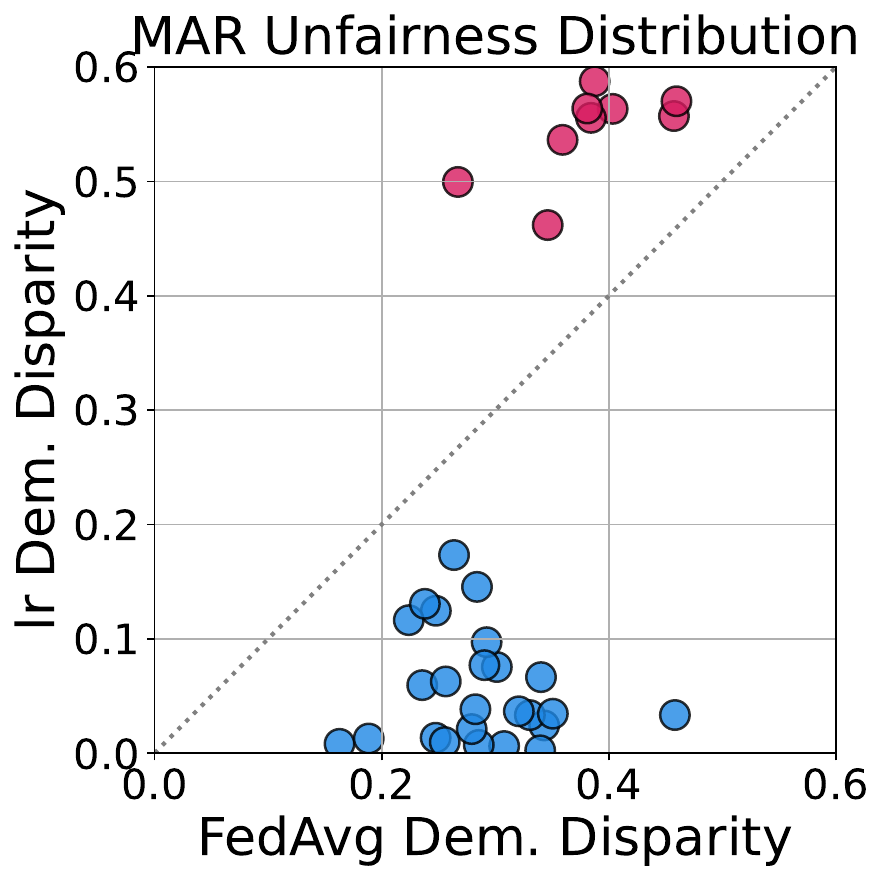}
  \includegraphics[width=0.15\textwidth]{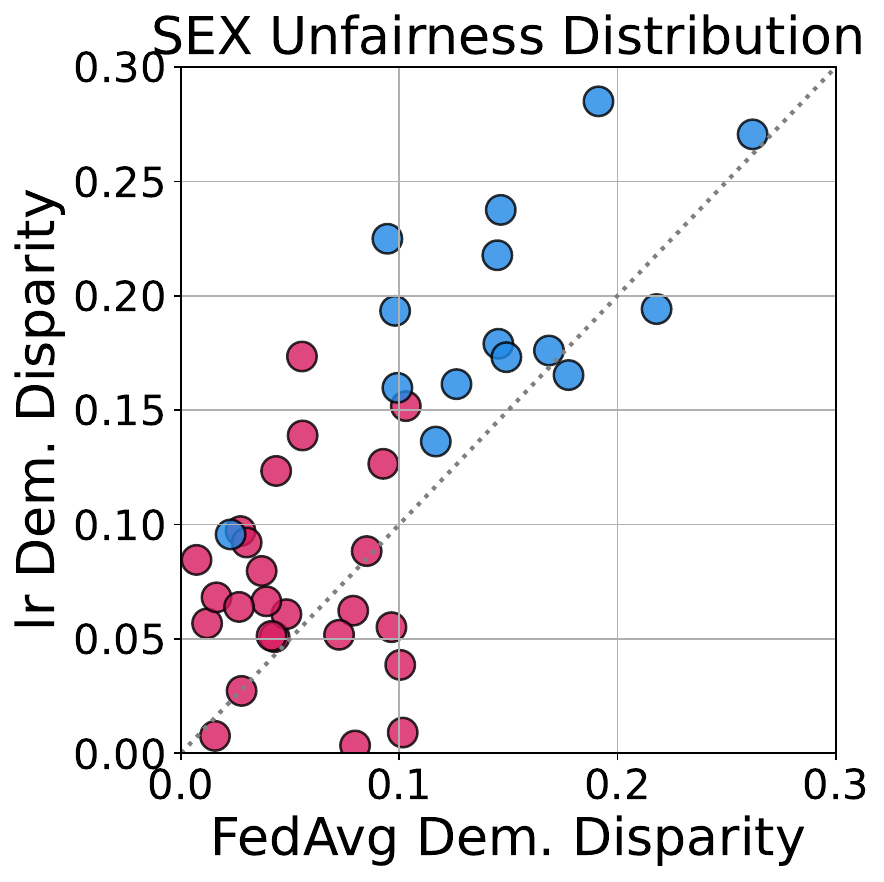}
  \label{fig:dutch_before_after_attribute_silo_lr}}
  \subfloat[\textbf{Attribute-silo:} LR vs. PUFFLE]{
  \includegraphics[width=0.15\textwidth]{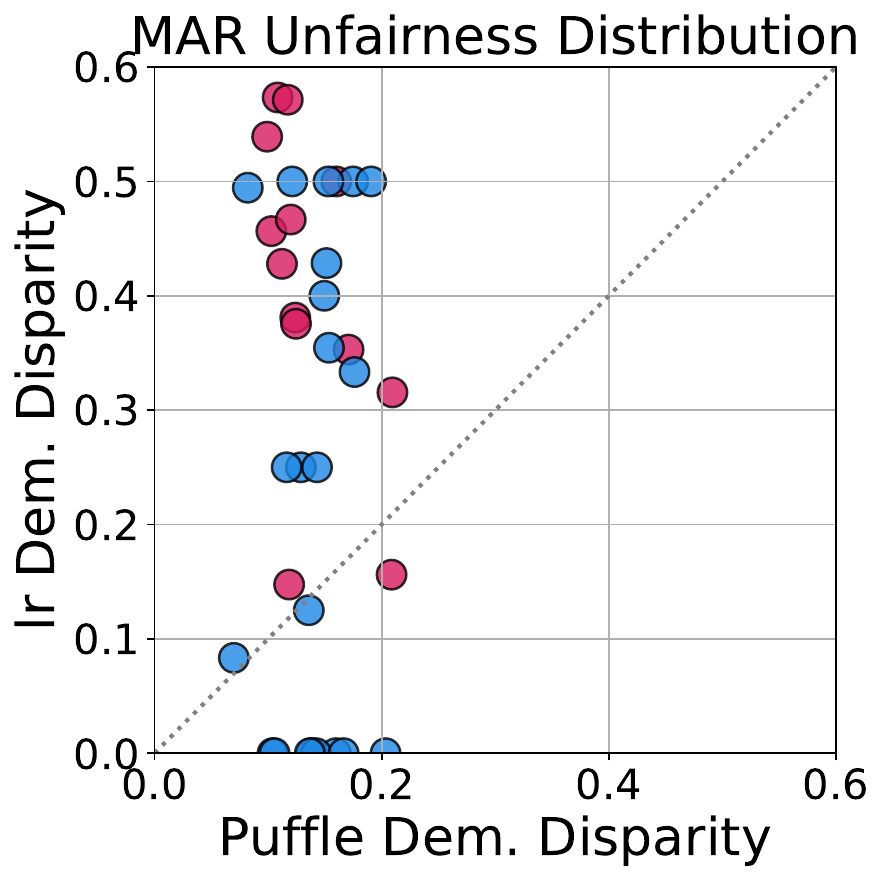}
  \includegraphics[width=0.15\textwidth]{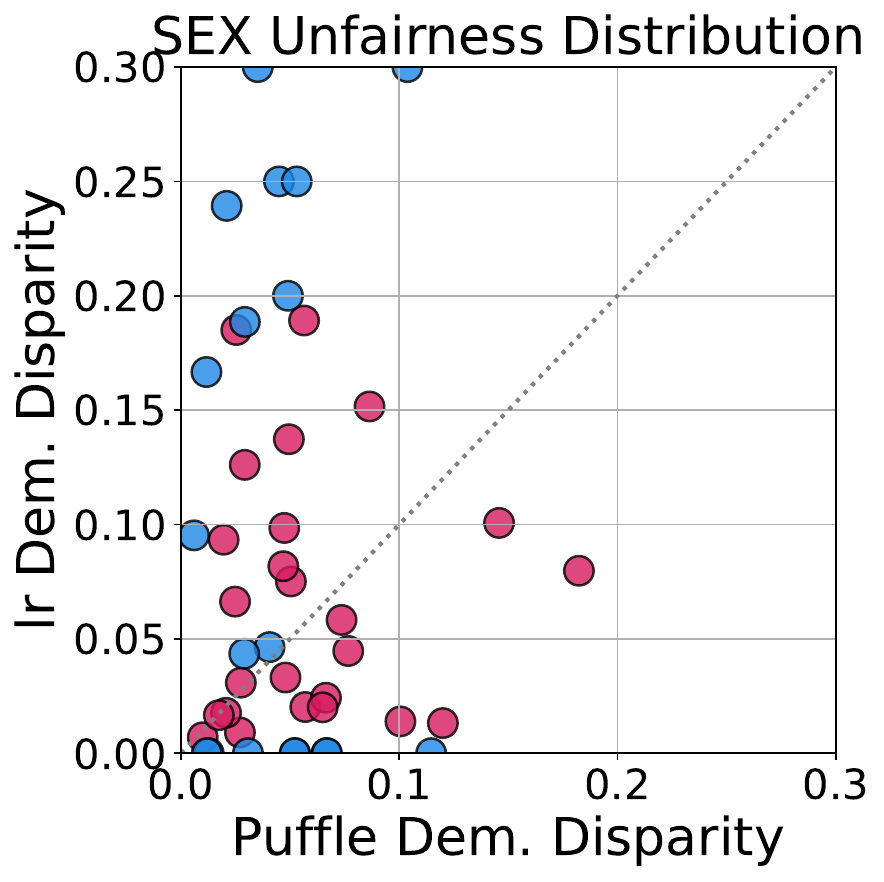}
  \label{fig:dutch_before_after_attribute_silo_lr_puffle}}
  \subfloat[\textbf{Attribute-silo:} LR vs. Reweighing]{
  \includegraphics[width=0.15\textwidth]{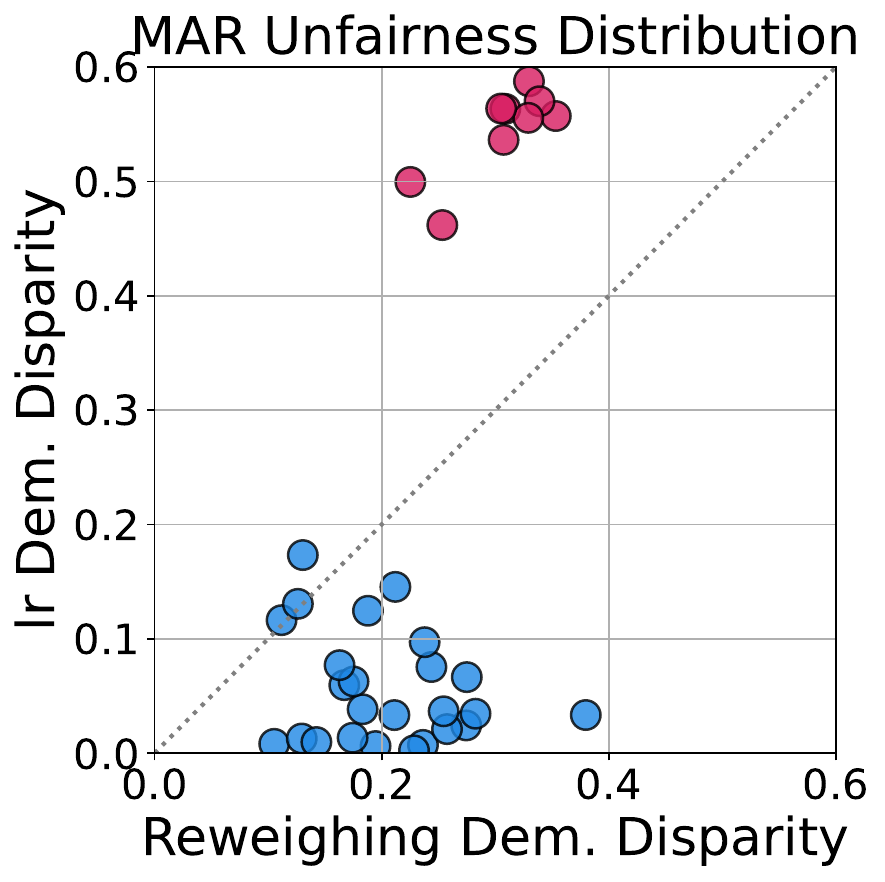}
  \includegraphics[width=0.15\textwidth]{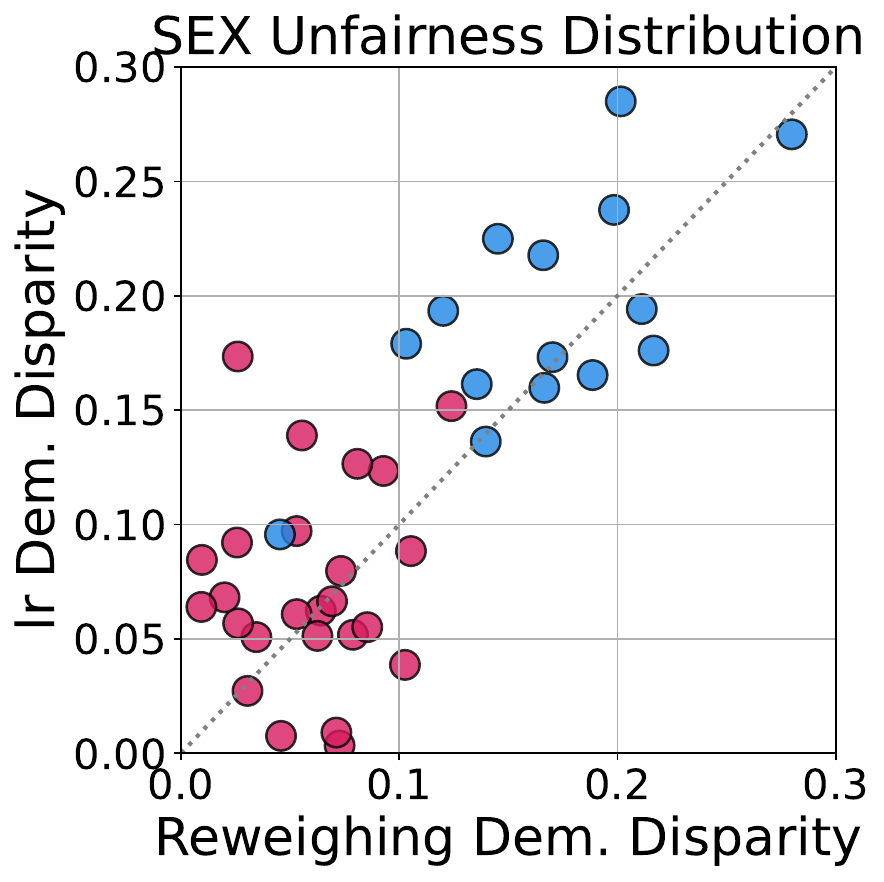}
  \label{fig:dutch_before_after_attribute_silo_lr_reweighing}}\\
    \centering\includegraphics[width=0.40\textwidth]{images/dutch/cross_device_attribute/baseline/legend_blue_red.pdf}
  \caption{\att toward \mar and \sex measured with Demographic Disparity on the local Logistic Regression (LR) models versus the FedAvg model, the PUFFLE model, and the Reweighing model for the attribute-silo \dutch dataset.}
  \label{fig:dutch_cross_silo_attribute_LR}
\end{figure}

\begin{figure}
  \subfloat[\textbf{Attribute-device:} XGB vs. FedAvg]{
  \includegraphics[width=0.15\textwidth]{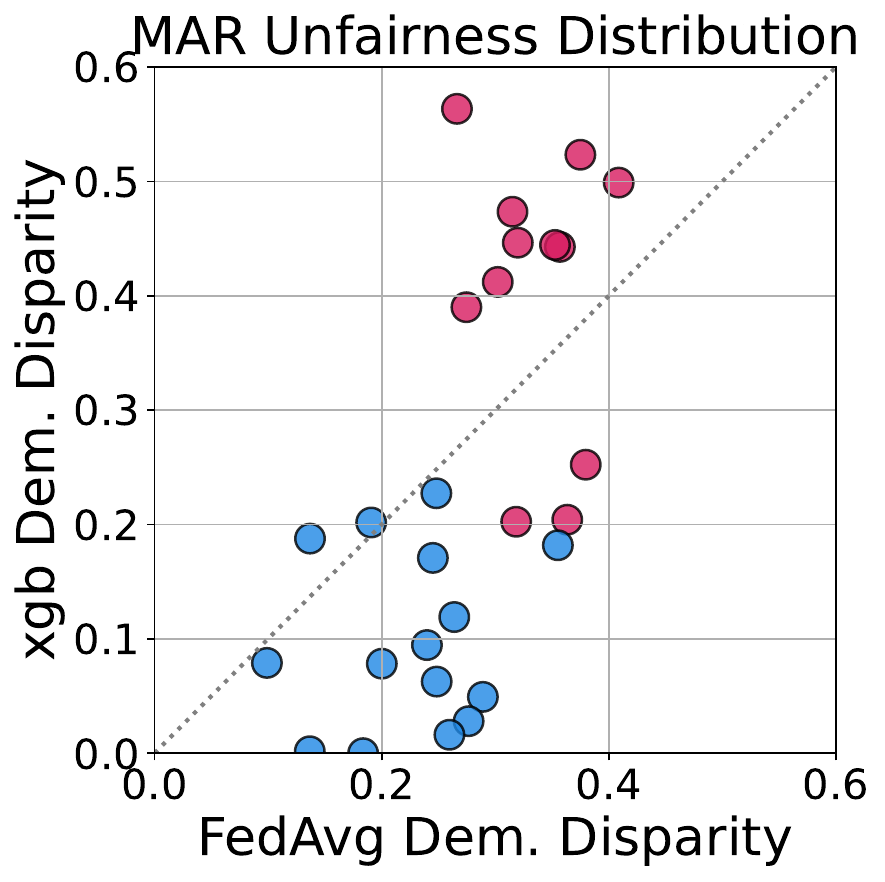}
  \includegraphics[width=0.15\textwidth]{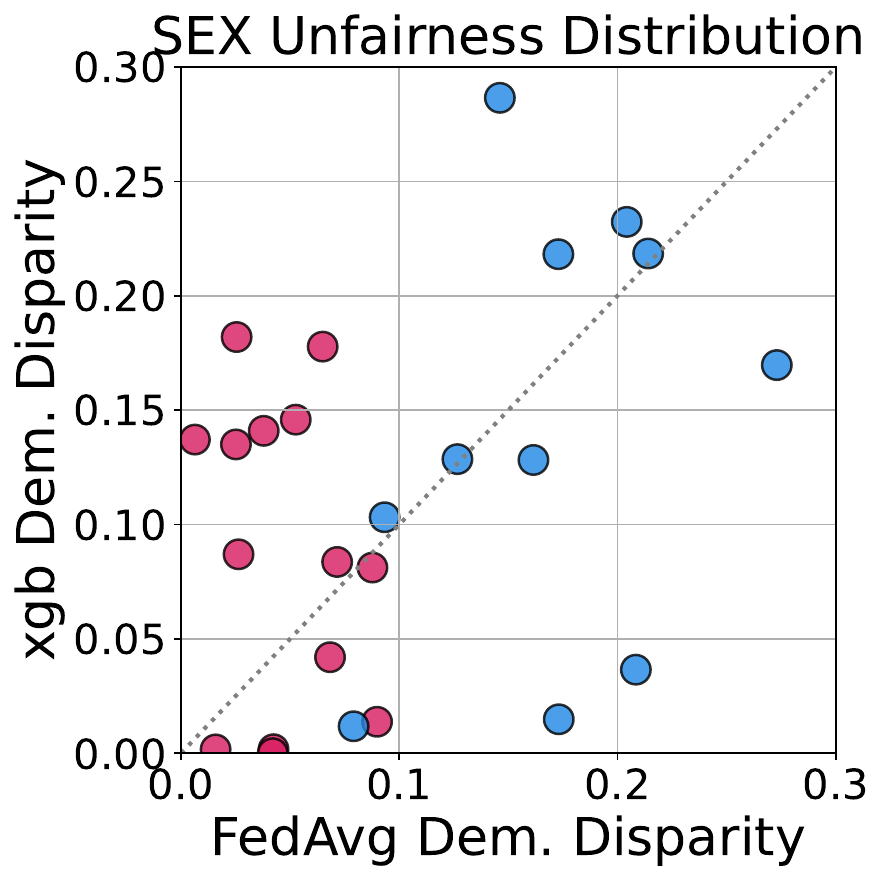}
  \label{fig:dutch_before_after_attribute_device_xgb}}
  \subfloat[\textbf{Attribute-device:} XGB vs. PUFFLE]{
  \includegraphics[width=0.15\textwidth]{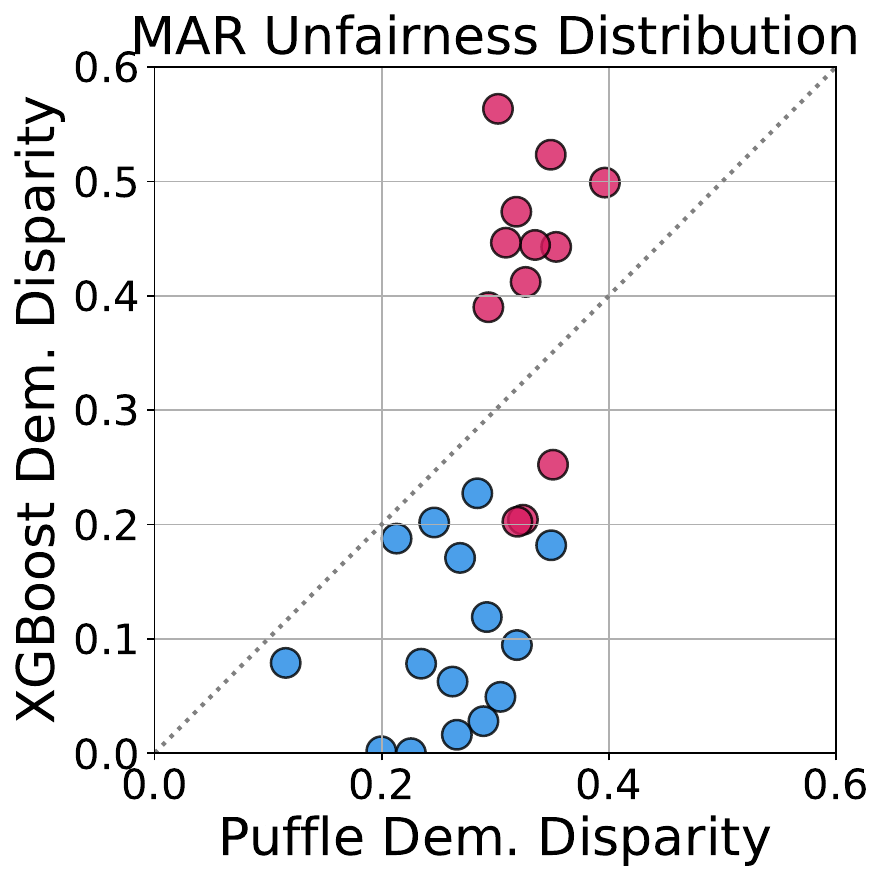}
  \includegraphics[width=0.15\textwidth]{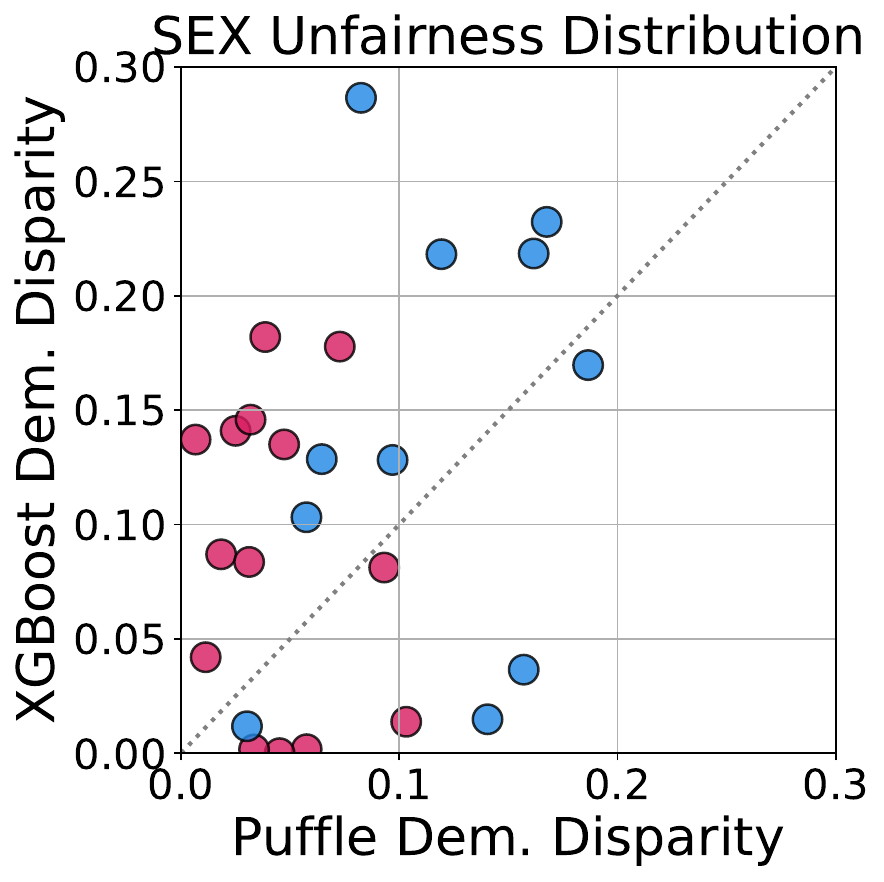}
  \label{fig:dutch_before_after_attribute_device_xgb_puffle}}
  \subfloat[\textbf{Attribute-device:} XGB vs. Reweighing]{
  \includegraphics[width=0.15\textwidth]{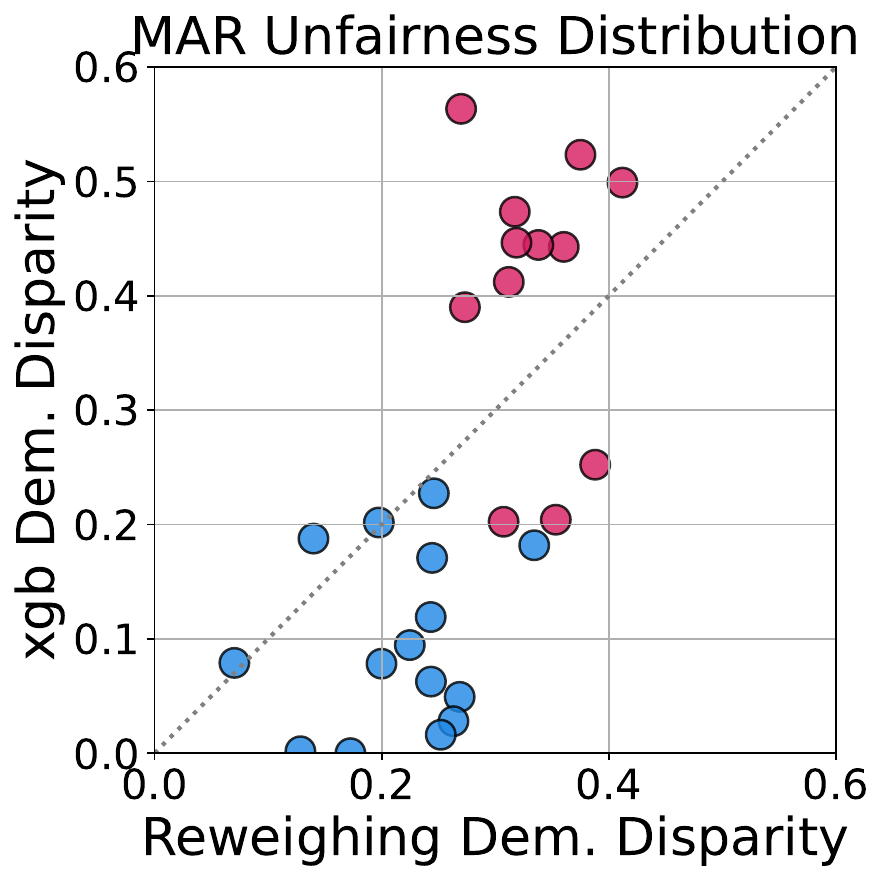}
  \includegraphics[width=0.15\textwidth]{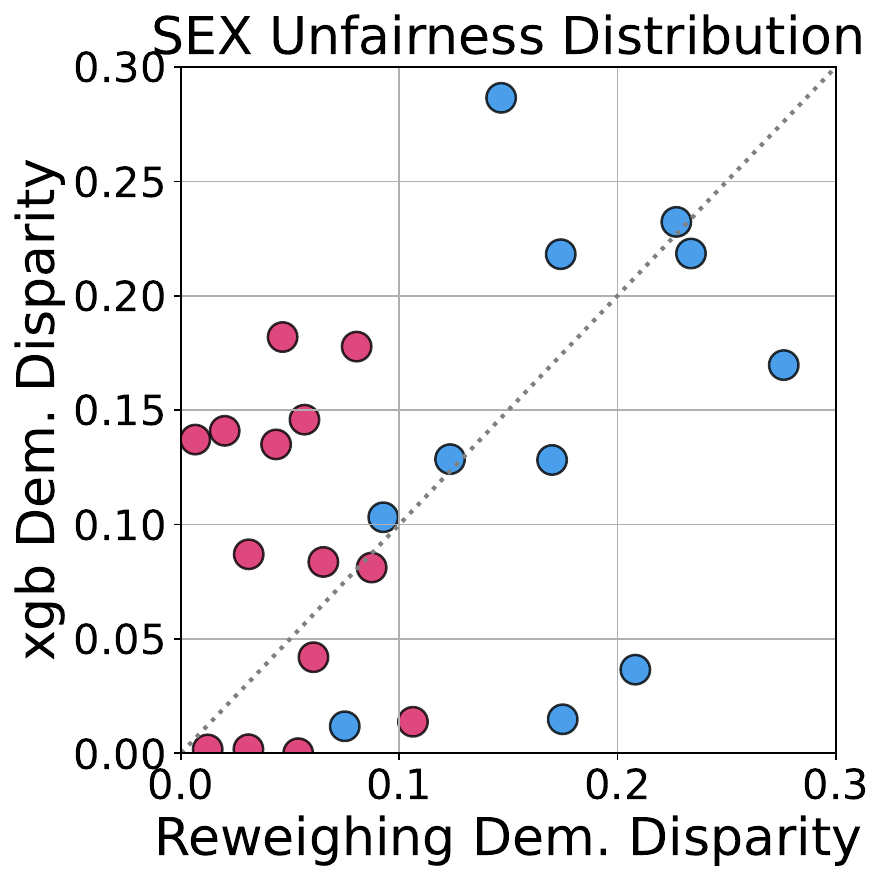}
  \label{fig:dutch_before_after_attribute_device_xgb_reweighing}}\\
    \centering\includegraphics[width=0.40\textwidth]{images/dutch/cross_device_attribute/baseline/legend_blue_red.pdf}
  \caption{\att toward \mar and \sex measured with Demographic Disparity on the local XGBoost models versus the FedAvg model, the PUFFLE model, and the Reweighing model for the attribute-device \dutch dataset.}
  \label{fig:dutch_cross_device_attribute_XGB}
\end{figure}

\begin{figure}
  \subfloat[\textbf{Attribute-device:} LR vs. FedAvg]{
  \includegraphics[width=0.15\textwidth]{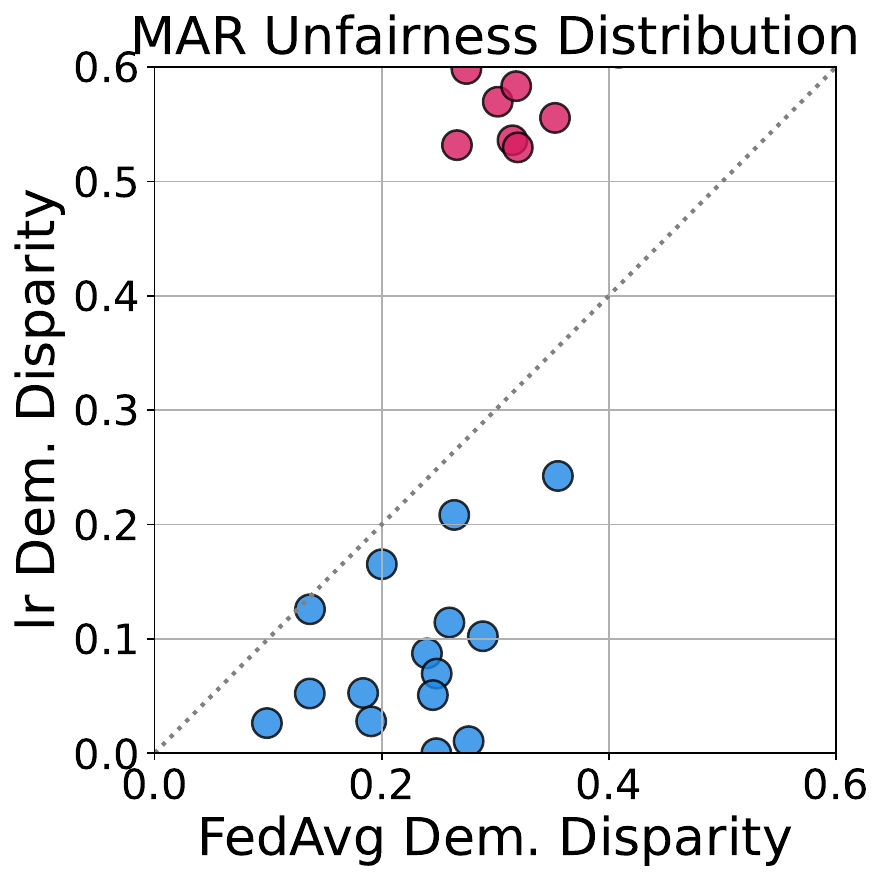}
  \includegraphics[width=0.15\textwidth]{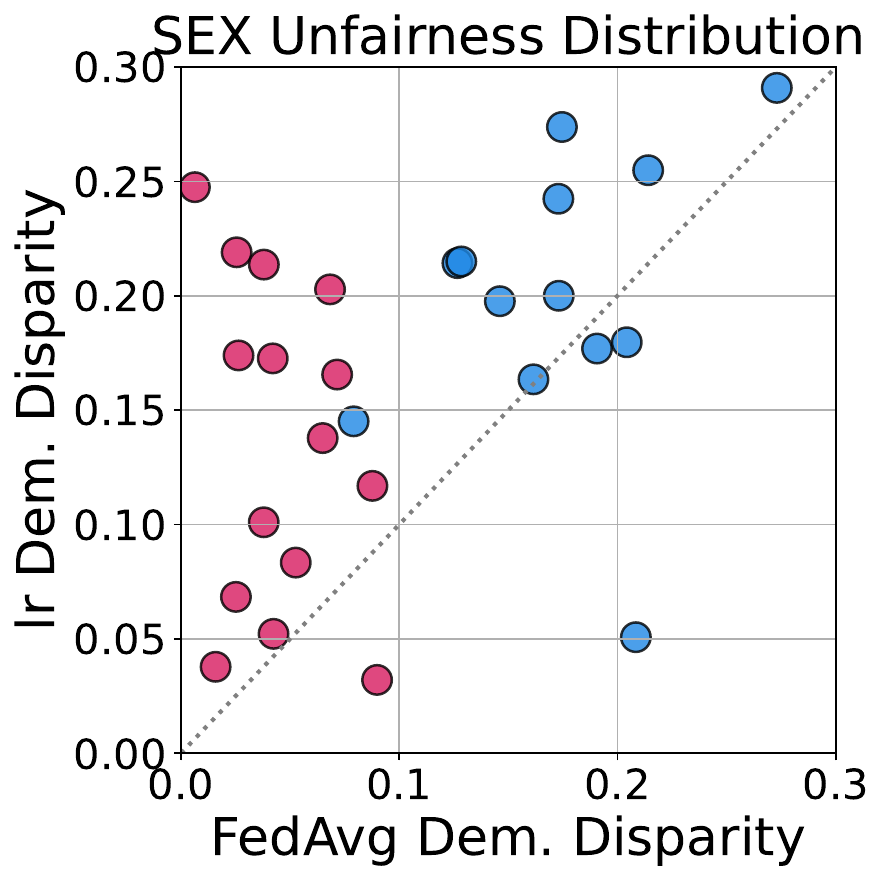}
  \label{fig:dutch_before_after_attribute_device_lr}}
  \subfloat[\textbf{Attribute-device:} LR vs. PUFFLE]{
  \includegraphics[width=0.15\textwidth]{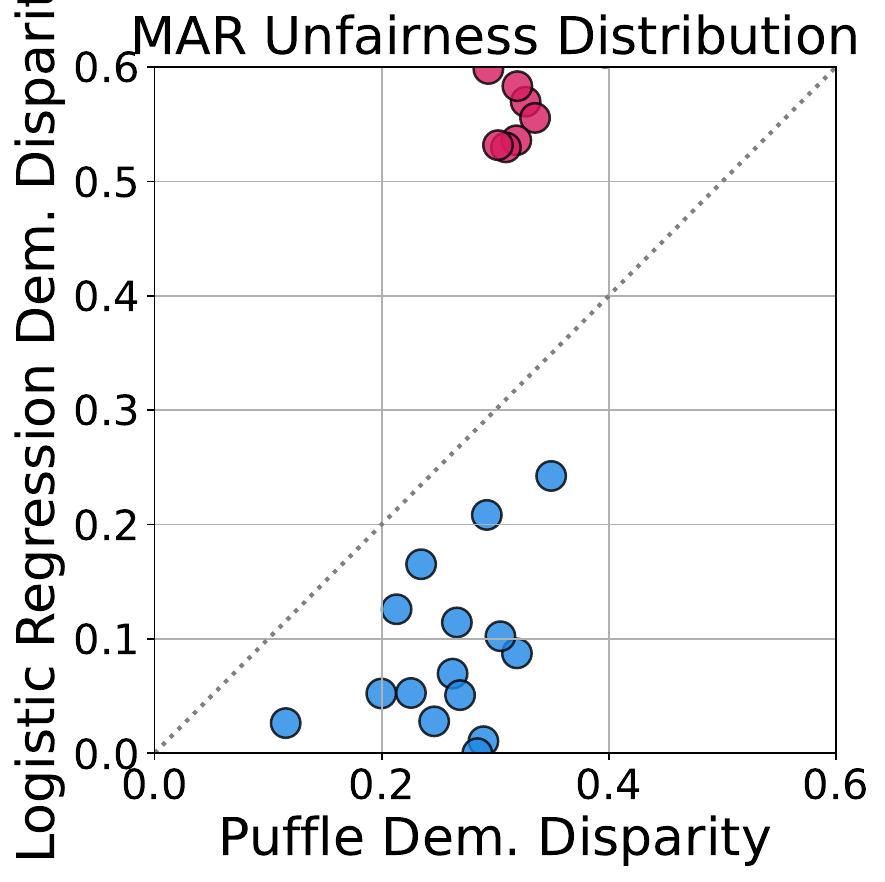}
  \includegraphics[width=0.15\textwidth]{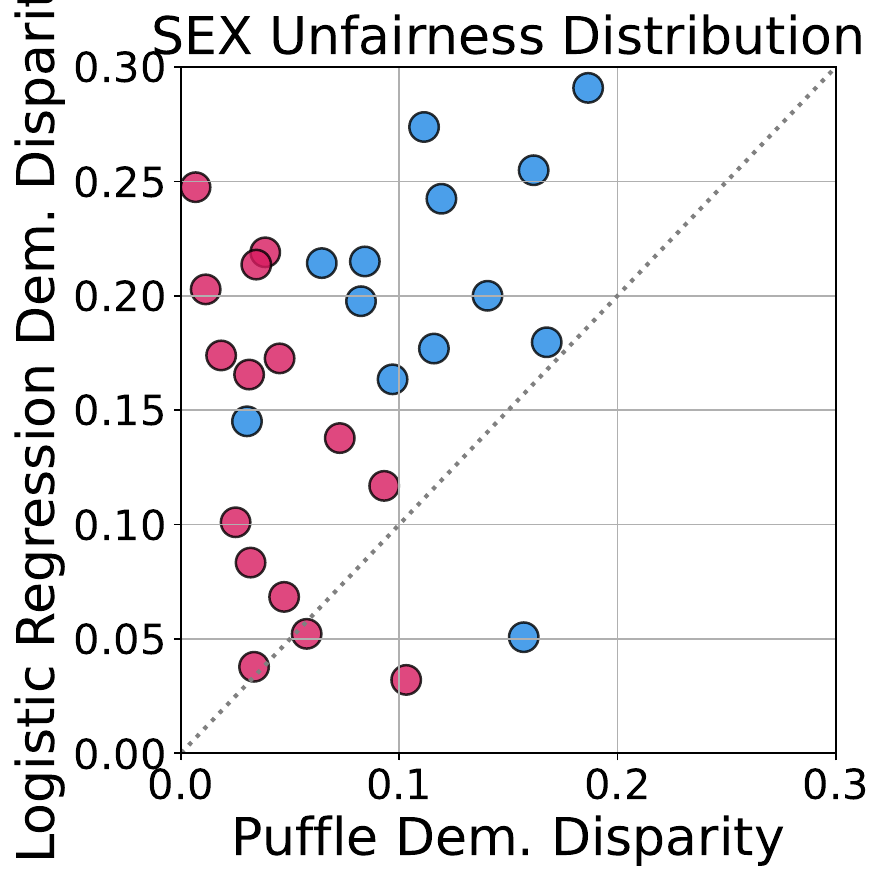}
  \label{fig:dutch_before_after_attribute_device_lr_puffle}}
  \subfloat[\textbf{Attribute-device:} LR vs. Reweighing]{
  \includegraphics[width=0.15\textwidth]{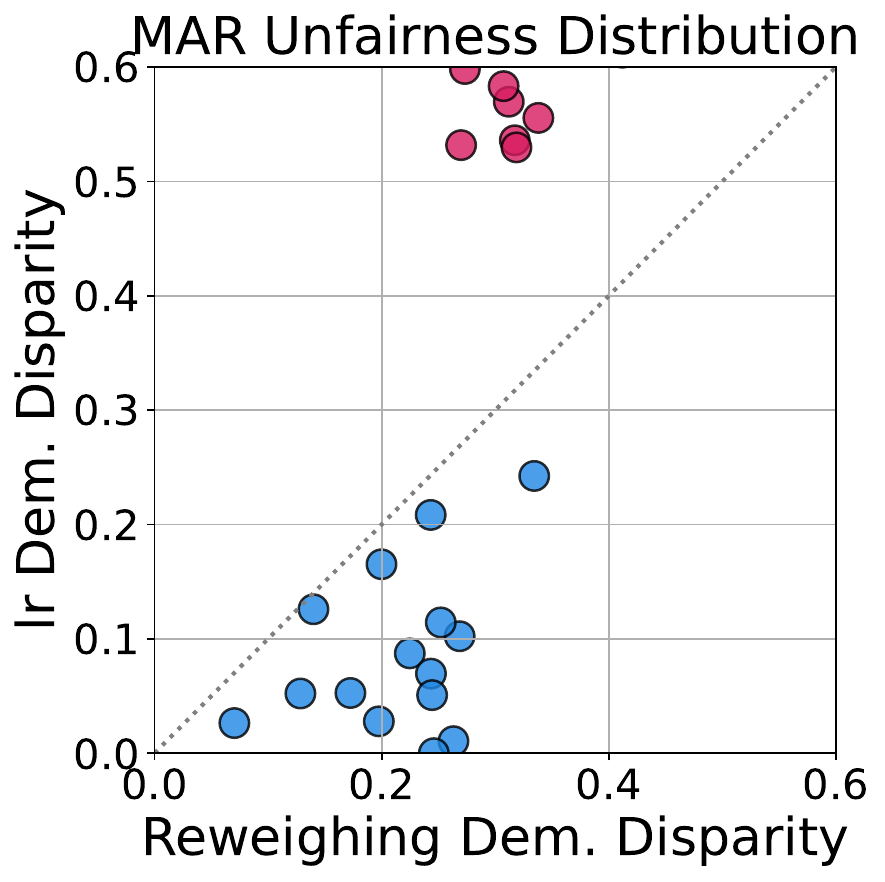}
  \includegraphics[width=0.15\textwidth]{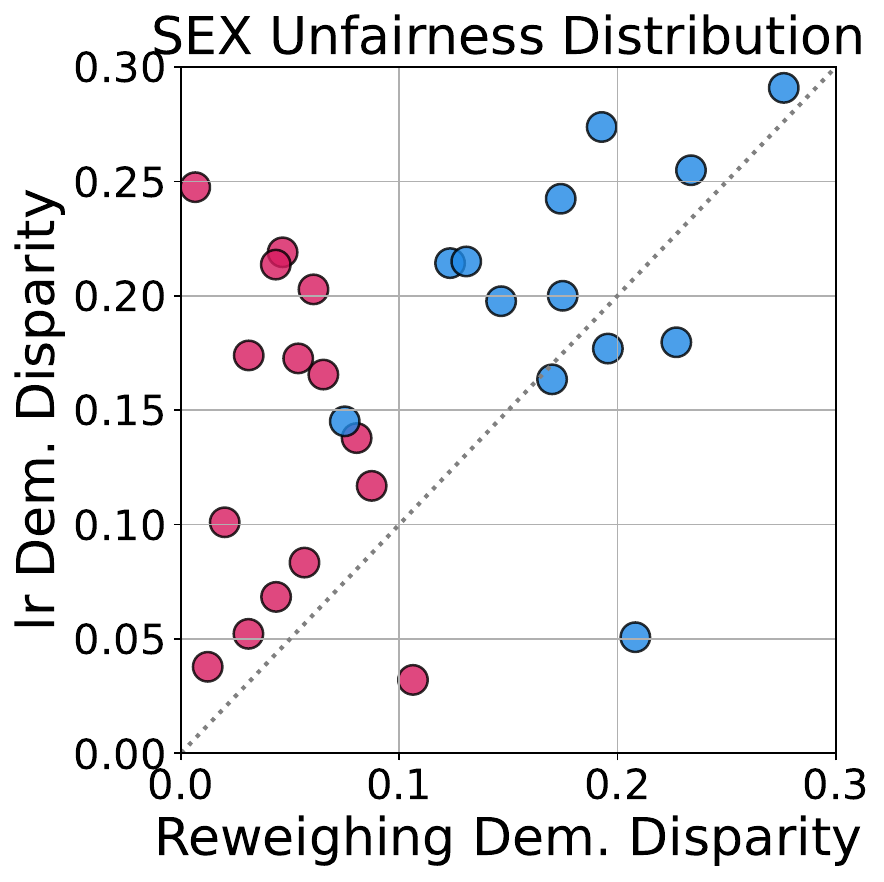}
  \label{fig:dutch_before_after_attribute_device_lr_reweighing}}\\
    \centering\includegraphics[width=0.40\textwidth]{images/dutch/cross_device_attribute/baseline/legend_blue_red.pdf}
  \caption{\att toward \mar and \sex measured with Demographic Disparity on the local Logistic Regression (LR) models versus the FedAvg model, the PUFFLE model, and the Reweighing model for the attribute-device \dutch dataset.}
  \label{fig:dutch_cross_device_attribute_LR}
\end{figure}

In Figures ~\ref{fig:dutch_cross_silo_attribute_XGB}, ~\ref{fig:dutch_cross_silo_attribute_LR}, ~\ref{fig:dutch_cross_device_attribute_XGB}, and ~\ref{fig:dutch_cross_device_attribute_LR}, we report the results for the attribute-silo and attribute-device datasets, comparing the Demographic Disparity (DD) of local models (XGBoost and Logistic Regression) against global models trained with FedAvg, PUFFLE, and Reweighing. Similar to our previous analysis, the diagonal line acts as a parity line: points below it indicate that the global model increased unfairness compared to the local model, while points above indicate a reduction of bias. Across both cross-silo and cross-device settings, training with standard FedAvg pushes the clients across the parity line, except the blue clients when evaluating the DD on the \mar attribute. 
When applying mitigation strategies, we notice different behaviors with PUFFLE and Reweighing. PUFFLE overall reduces the DD value for most clients compared to local models when measured for the \sex attribute. Considering the \mar attribute, a similar trend can be seen for the attribute-silo datasets in Figures ~\ref{fig:dutch_cross_silo_attribute_XGB}, ~\ref{fig:dutch_cross_silo_attribute_LR}. Conversely, for the cross-device datasets, Figures~\ref{fig:dutch_cross_device_attribute_XGB}, and ~\ref{fig:dutch_cross_device_attribute_LR} show a different behavior. Here, the two groups of clients differ; PUFFLE improves the DD value for \mar biased clients, whereas the \sex-biased clients are below the diagonal, indicating their bias towards the \mar attribute was exacerbated by the global model. 
For Reweighing, instead, bias reduction is less pronounced, as most clients lie close to the diagonal and overall similar findings to the FedAvg baseline. These results validate our previous findings: single-attribute mitigation strategies struggle in heterogeneous federations and can inadvertently increase unfairness for clients with conflicting fairness objectives.

\subsubsection{Additional \val benchmark experiments}
\begin{figure}   
\subfloat[\textbf{Value-silo:} XGB vs. FedAvg]{\includegraphics[width=0.44\textwidth]{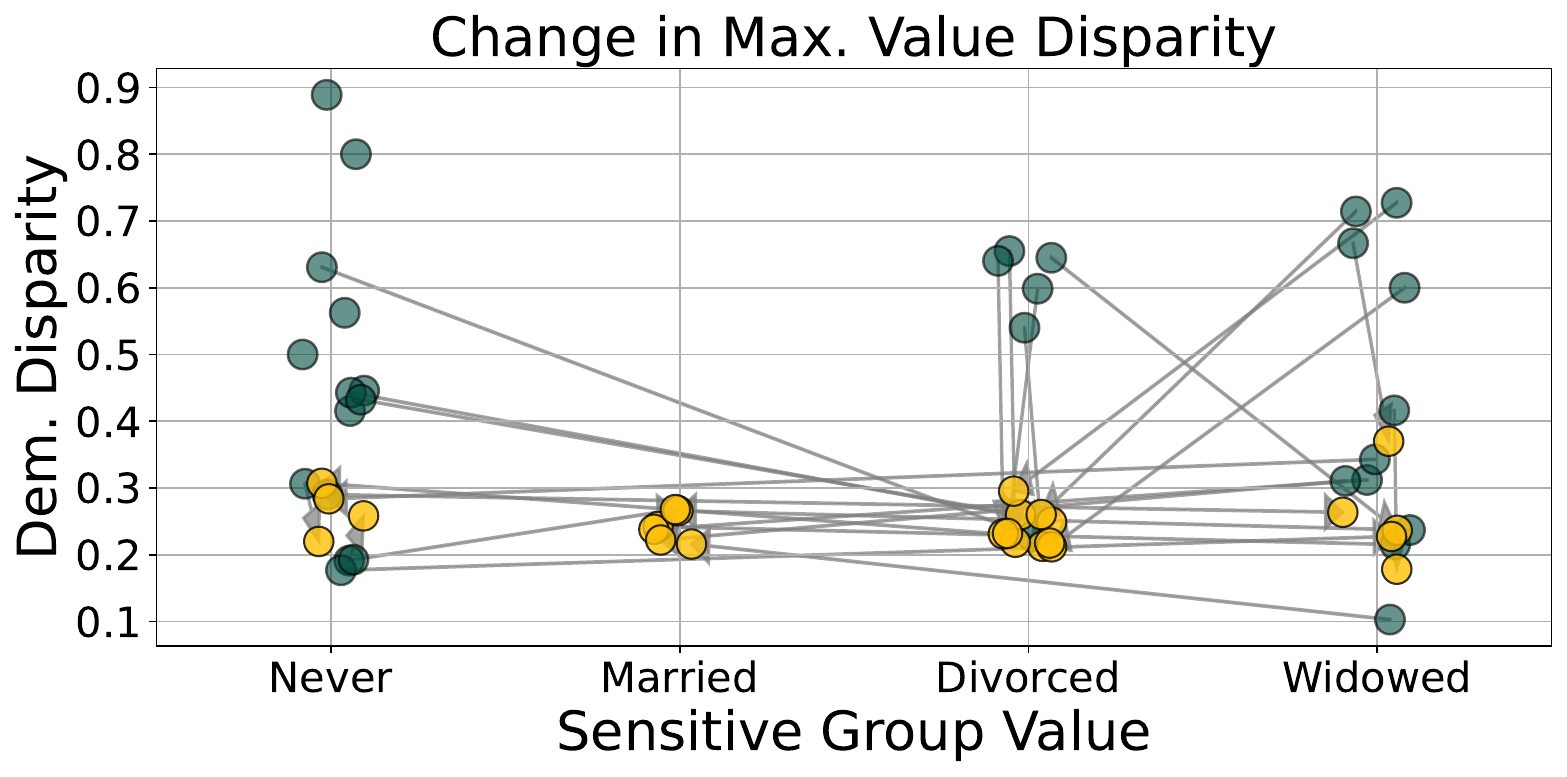}}
  \subfloat[\textbf{Value-silo:} XGB vs. PUFFLE]{\includegraphics[width=0.44\textwidth]{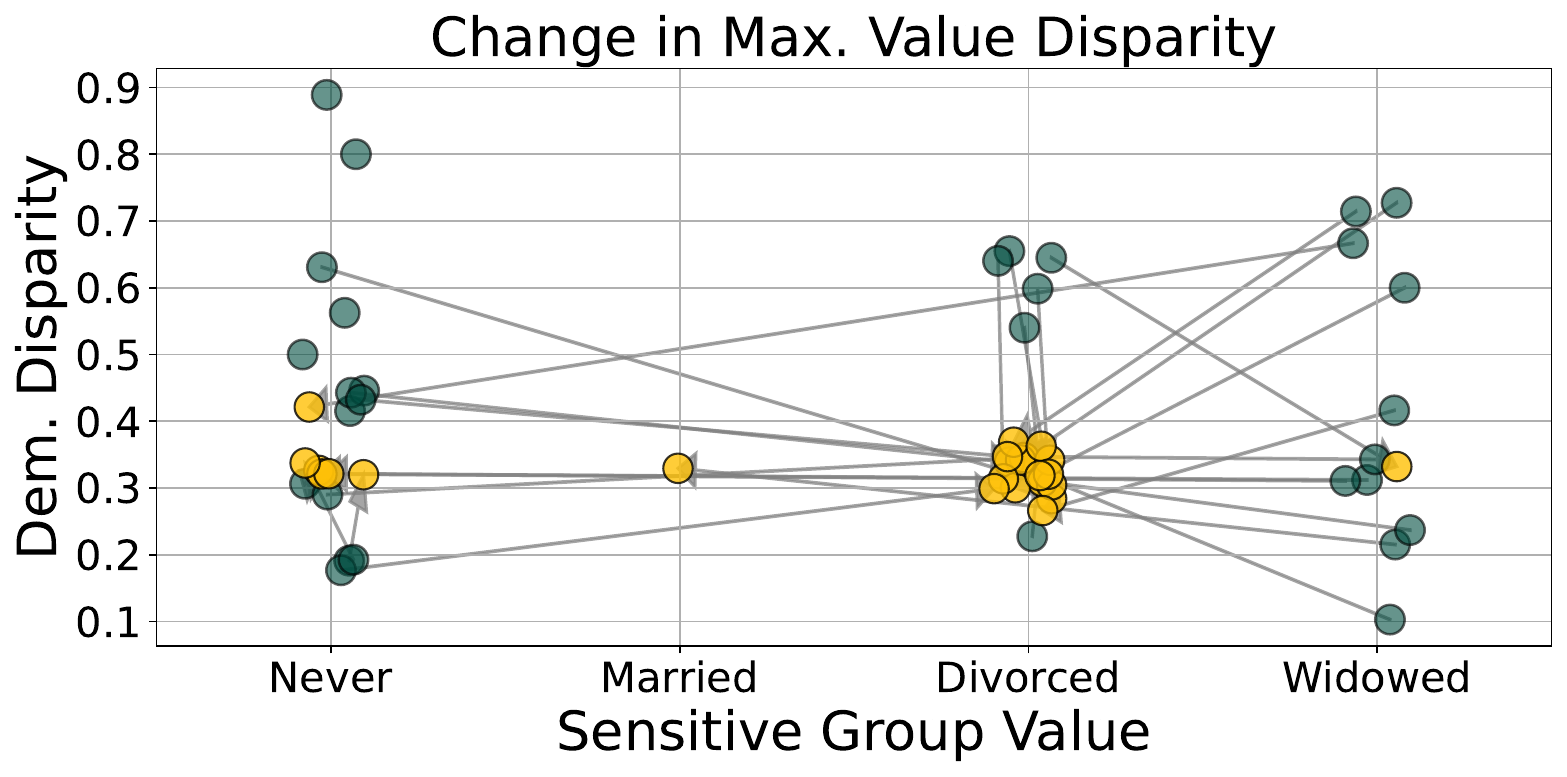}}\\
  \subfloat[\textbf{Value-silo:} XGB vs. Reweighing]
  {\includegraphics[width=0.44\textwidth]{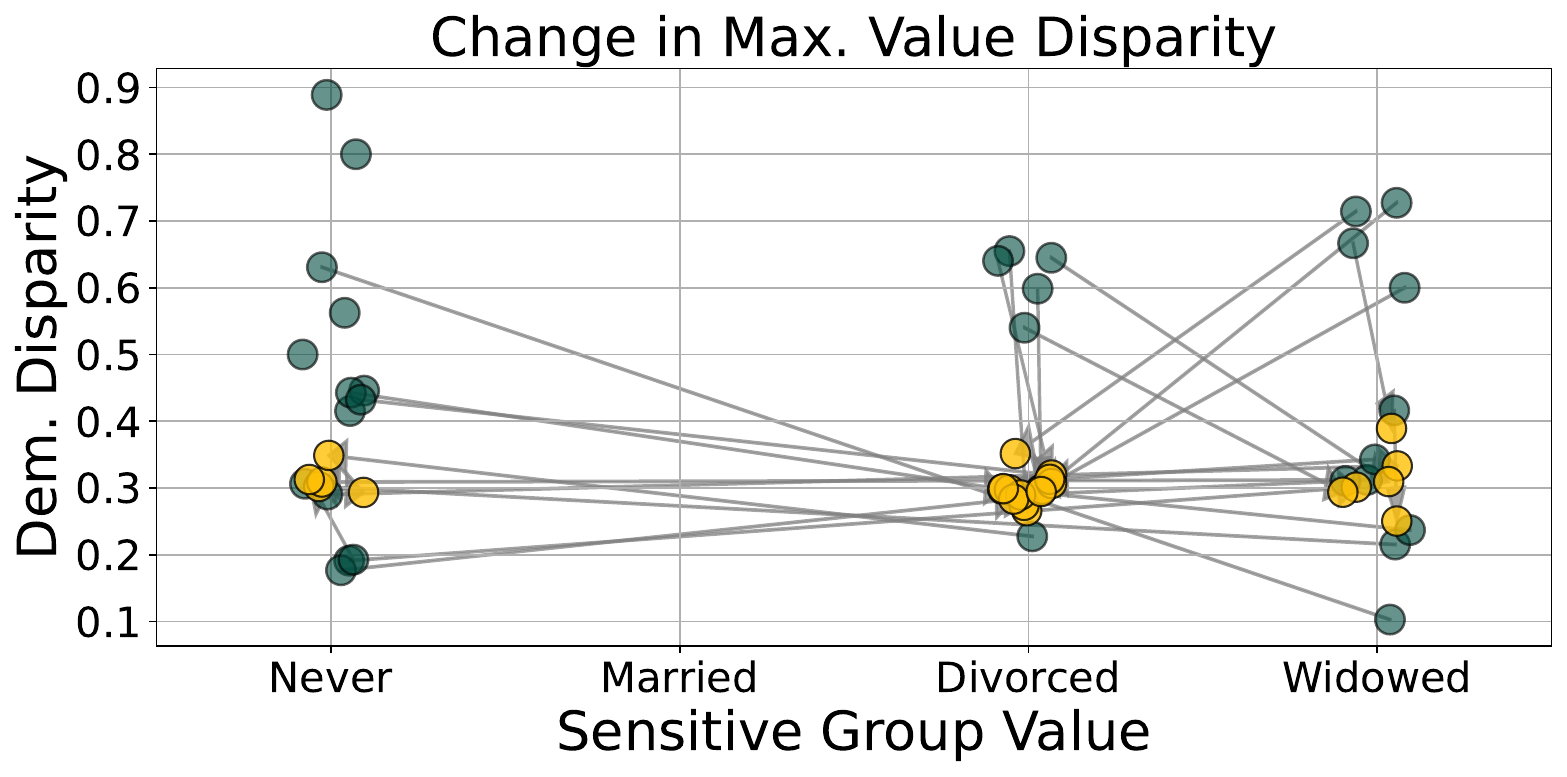}}\\
  \centering\includegraphics[width=0.40\textwidth]{images/legend/value_change_legend_xgboost.pdf}
  \caption{\val toward \mar as well as value changes measured with Demographic Disparity on the local XGBoost models versus the FedAvg model, the PUFFLE model, and the Reweighing model for the value-silo \dutch datasets.}
  \label{fig:before_after_silo_value_xgboost_dutch}
\end{figure}

\begin{figure}   
\subfloat[\textbf{Value-silo:} LR vs. FedAvg]{\includegraphics[width=0.44\textwidth]{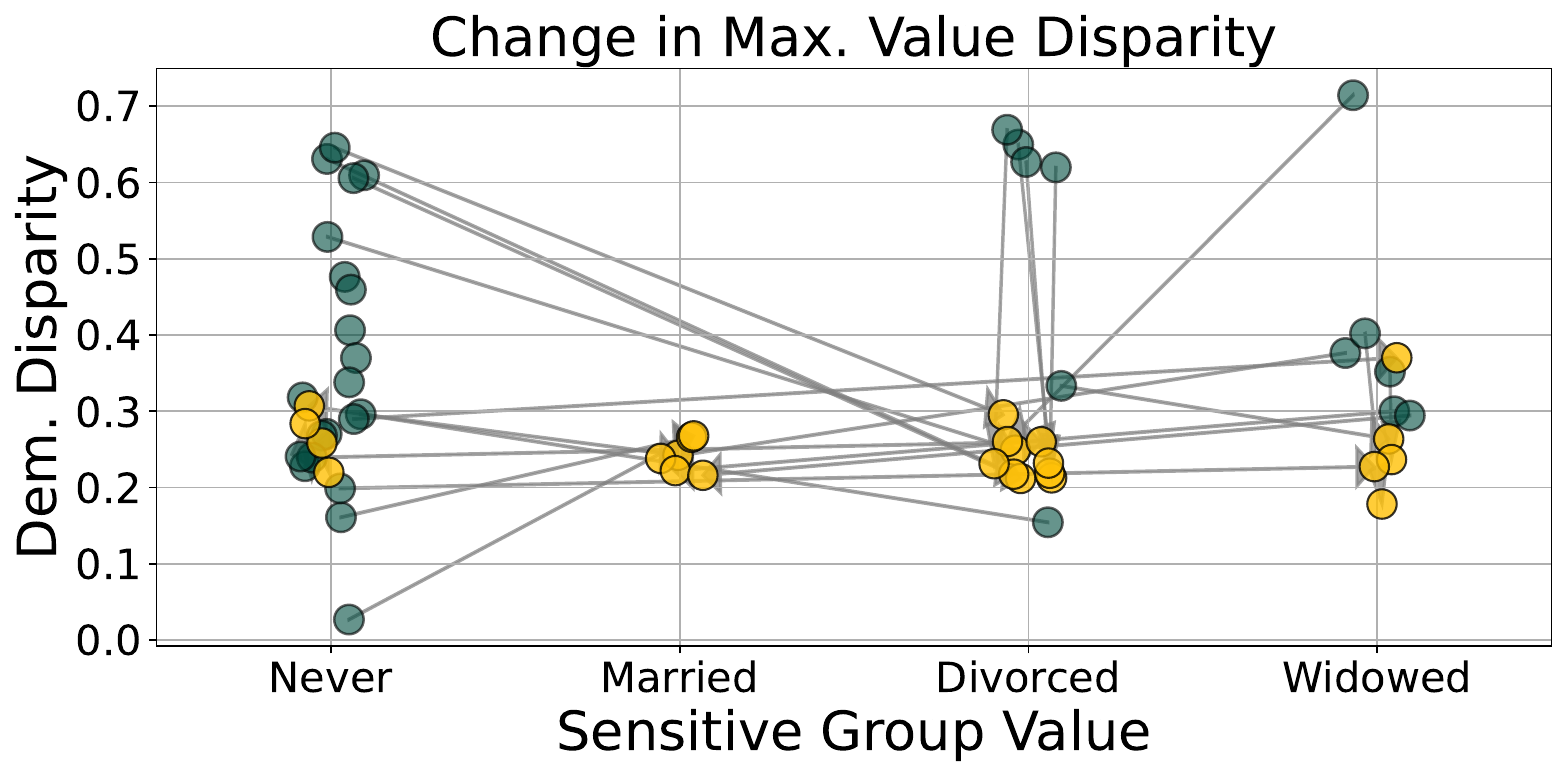}}
  \subfloat[\textbf{Value-silo:} LR vs. PUFFLE]{\includegraphics[width=0.44\textwidth]{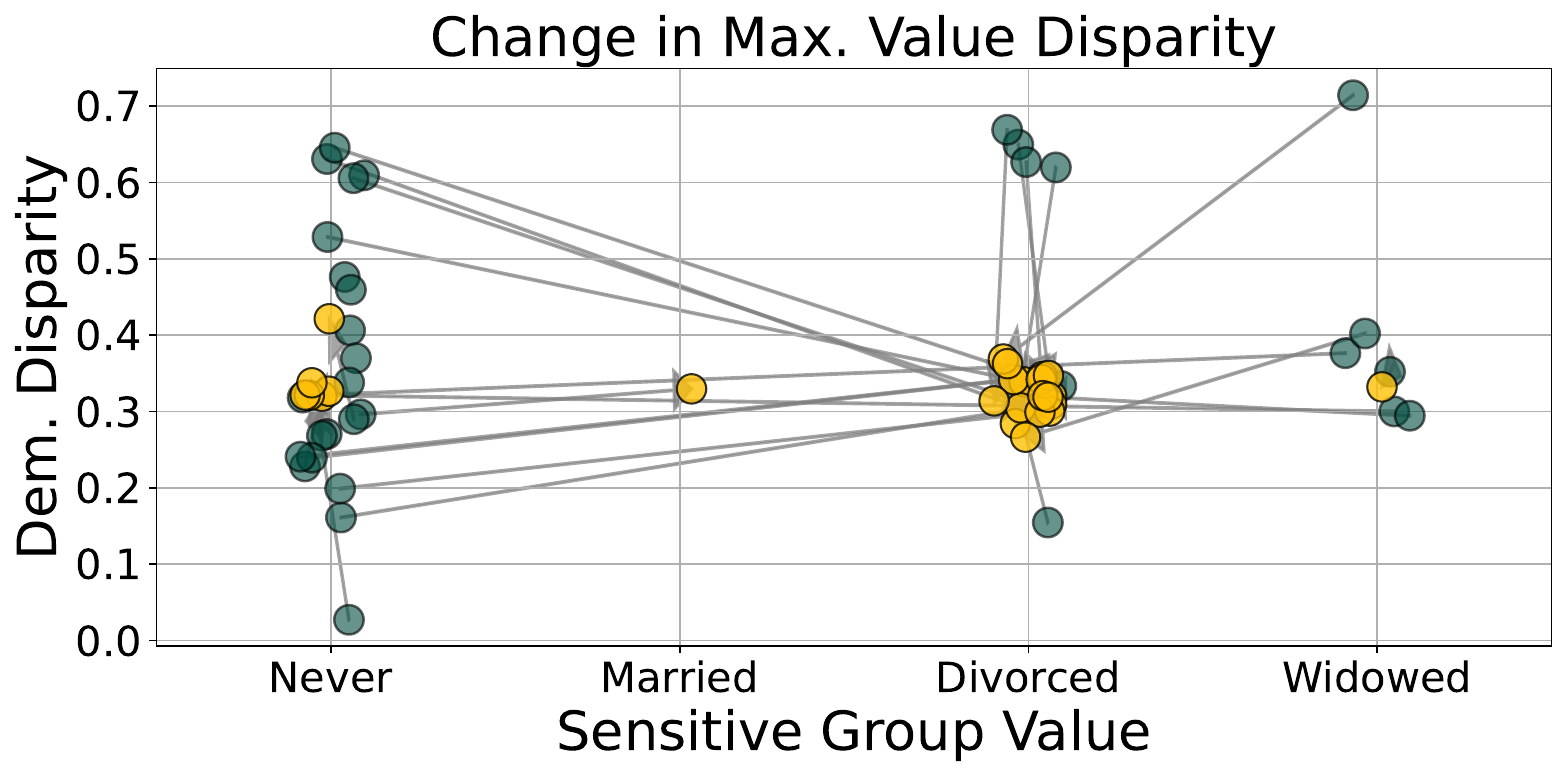}}\\
  \subfloat[\textbf{Value-silo:} LR vs. Reweighing]
  {\includegraphics[width=0.44\textwidth]{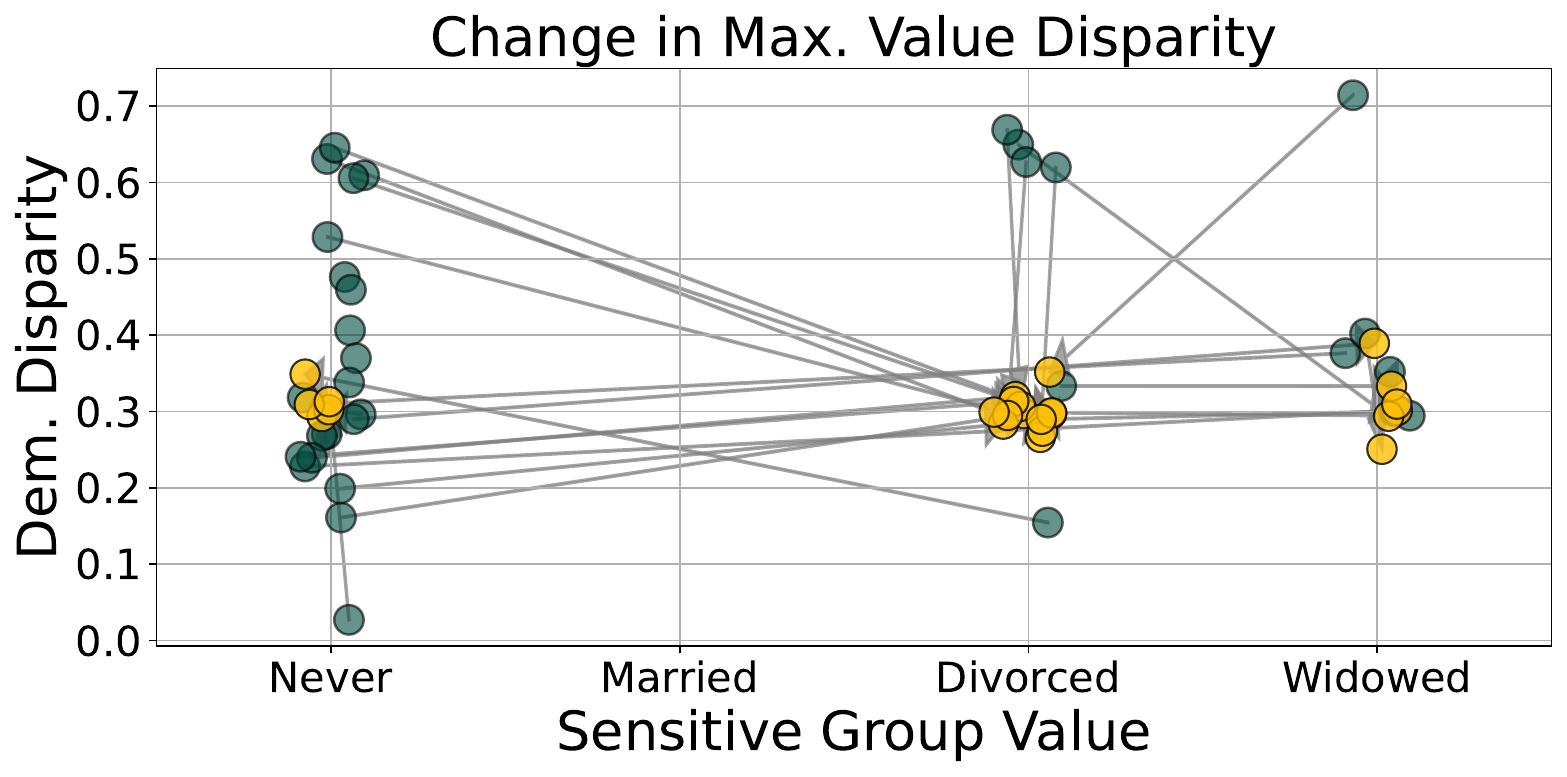}}\\
  \centering\includegraphics[width=0.40\textwidth]{images/legend/value_change_legend.pdf}
  \caption{\val toward \mar as well as value changes measured with Demographic Disparity on the local Logistic Regression (LR) models versus the FedAvg model, the PUFFLE model, and the Reweighing model for the value-silo \dutch datasets.}
  \label{fig:before_after_silo_value_lr_dutch}
\end{figure}

In Figures \ref{fig:before_after_silo_value_xgboost_dutch} and \ref{fig:before_after_silo_value_lr_dutch} we show the results for the \dutch value-silo datasets. These figures show both how DD changes when moving from local training to FL and how the distribution of the attribute value associated with the maximum DD shifts. Firstly, we see that overall DD values are in a tighter range for the FL models. Furthermore, a shift towards \texttt{Divorced} can be noticed for the FL models, which is more pronounced for PUFFLE and Reweighing. Again, \texttt{Divorced} is the most underrepresented group in our dataset (see Figure~\ref{fig:fd_cs_val_dutch}).

\begin{figure}   
\subfloat[\textbf{Value-device:} XGB vs. FedAvg]{\includegraphics[width=0.44\textwidth]{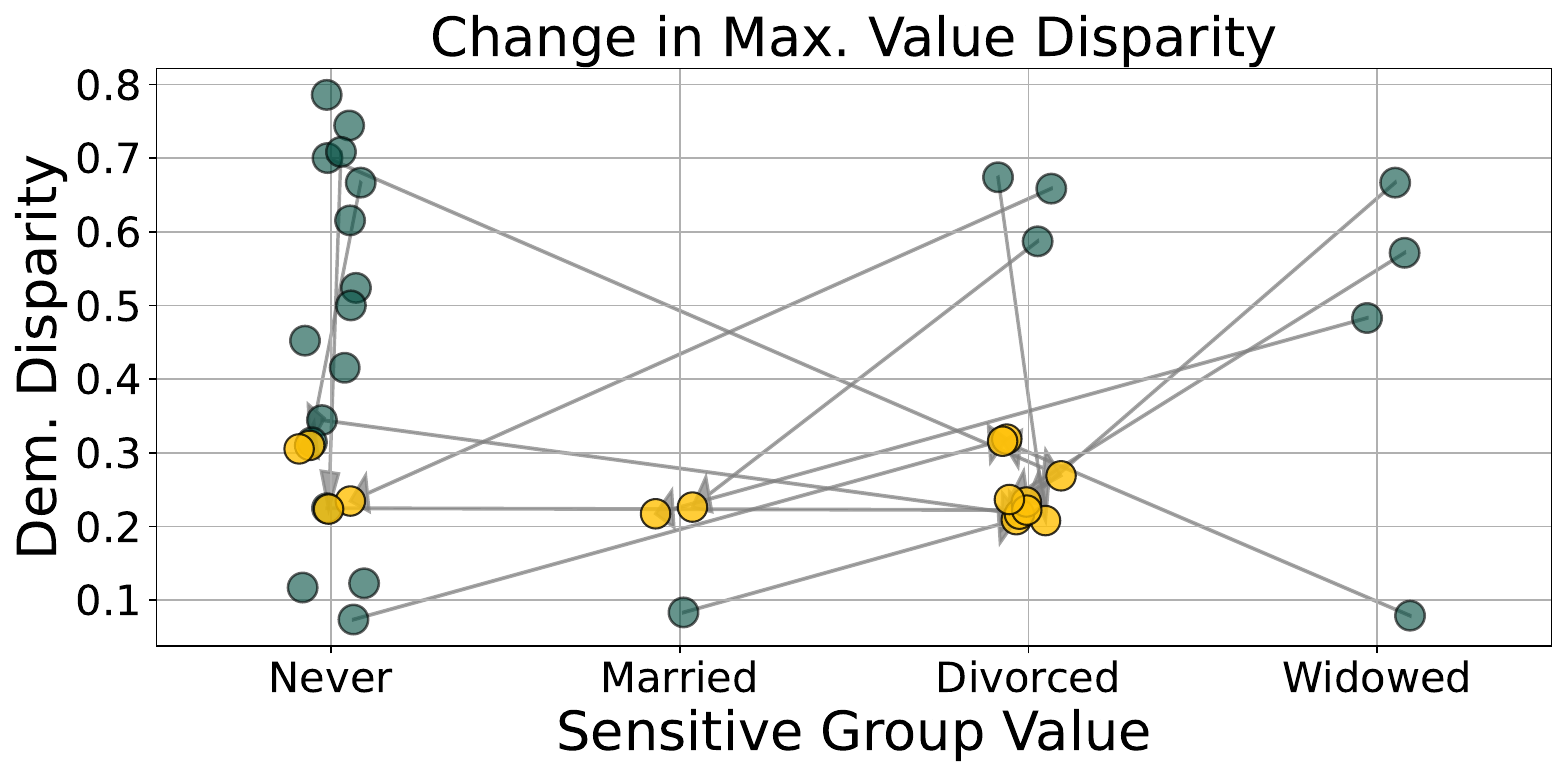}}
  \subfloat[\textbf{Value-device:} XGB vs. PUFFLE]{\includegraphics[width=0.44\textwidth]{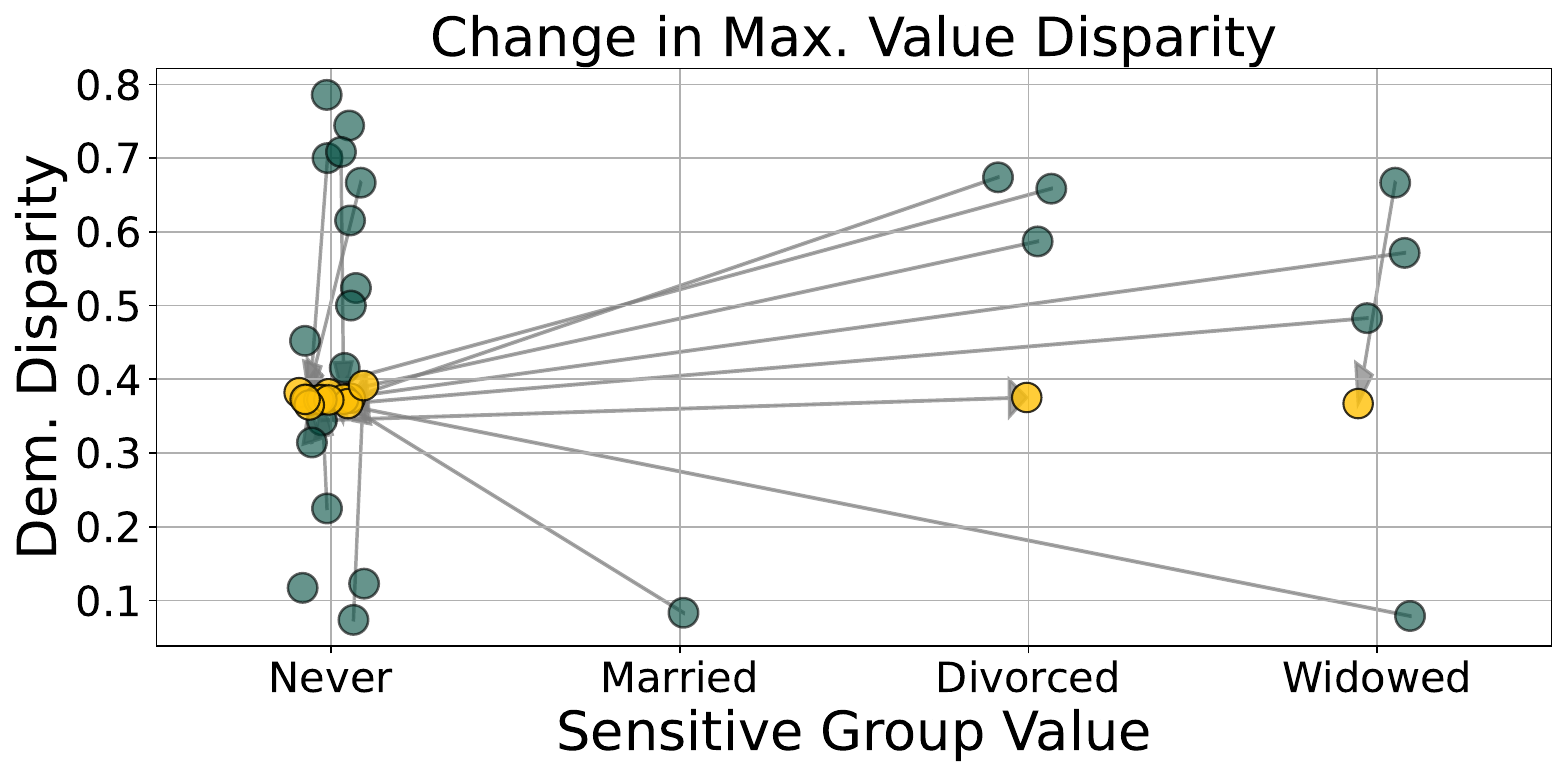}}\\
  \subfloat[\textbf{Value-device:} XGB vs. Reweighing]
  {\includegraphics[width=0.44\textwidth]{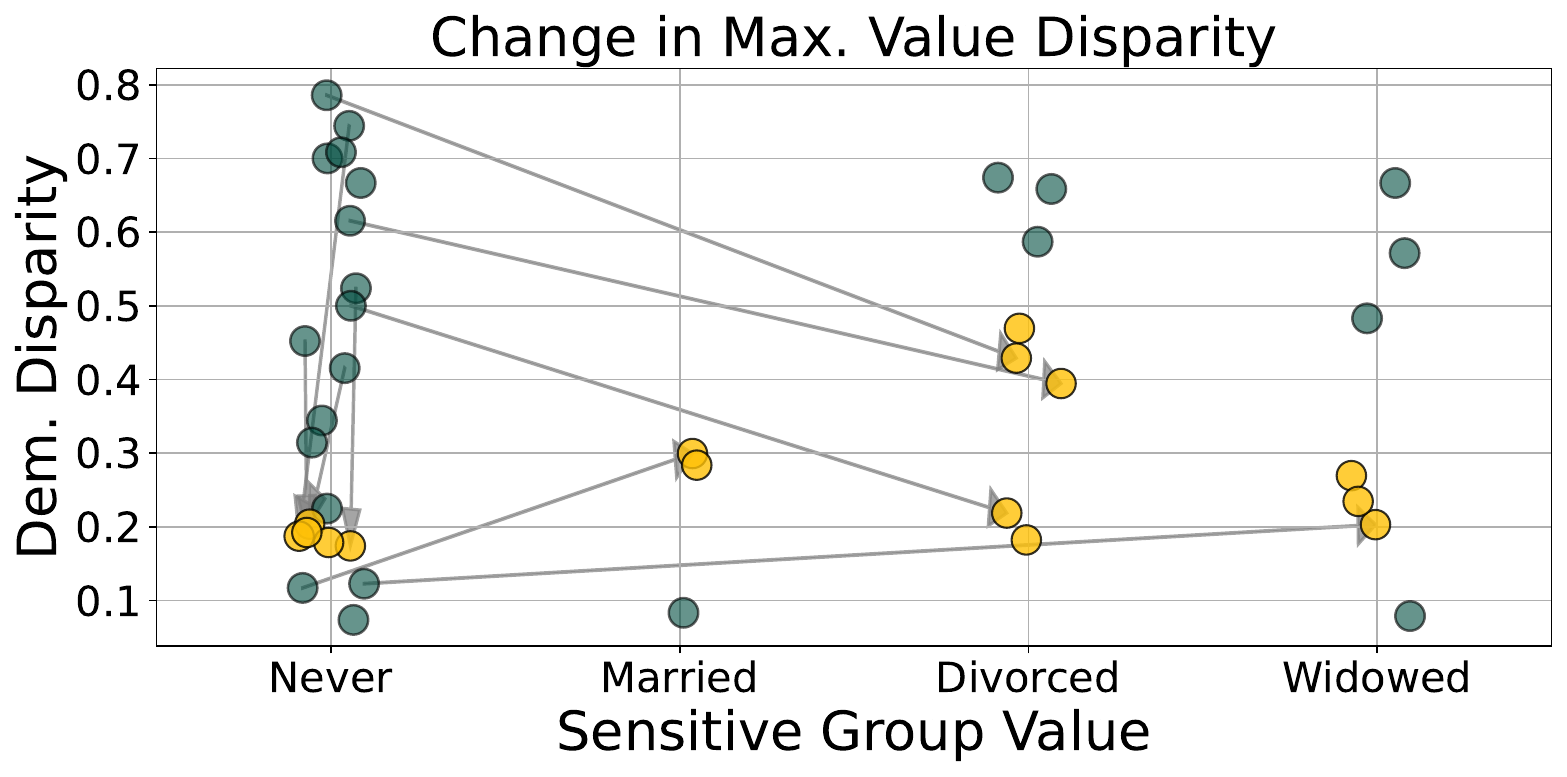}}\\
  \centering\includegraphics[width=0.40\textwidth]{images/legend/value_change_legend_xgboost.pdf}
  \caption{\val toward \mar as well as value changes measured with Demographic Disparity on the local XGBoost models versus the FedAvg model, the PUFFLE model, and the Reweighing model for the value-device \dutch datasets.}
  \label{fig:before_after_device_value_xgboost_dutch}
\end{figure}

\begin{figure}   
\subfloat[\textbf{Value-device:} LR vs. FedAvg]{\includegraphics[width=0.44\textwidth]{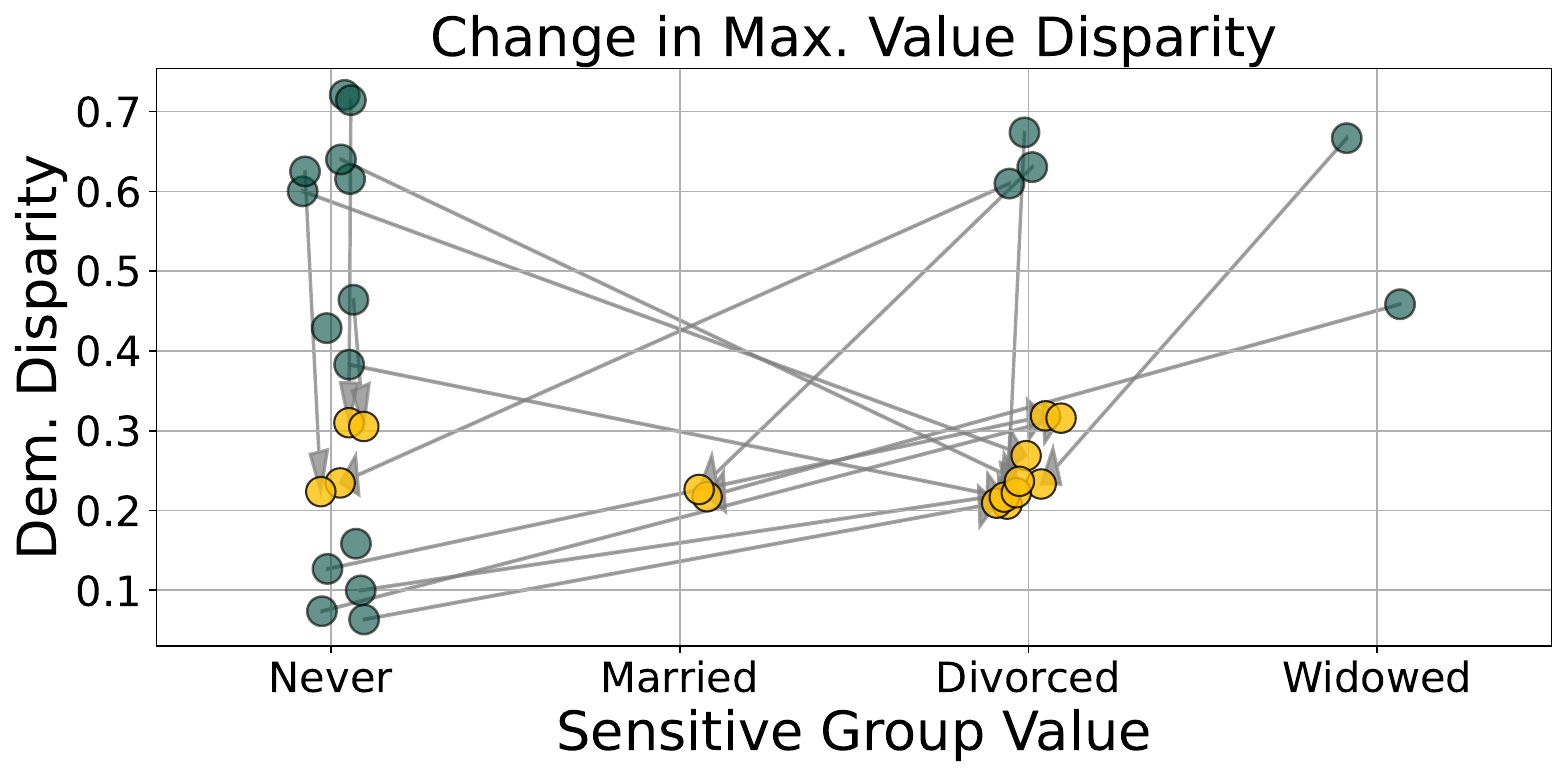}}
  \subfloat[\textbf{Value-device:} LR vs. PUFFLE]{\includegraphics[width=0.44\textwidth]{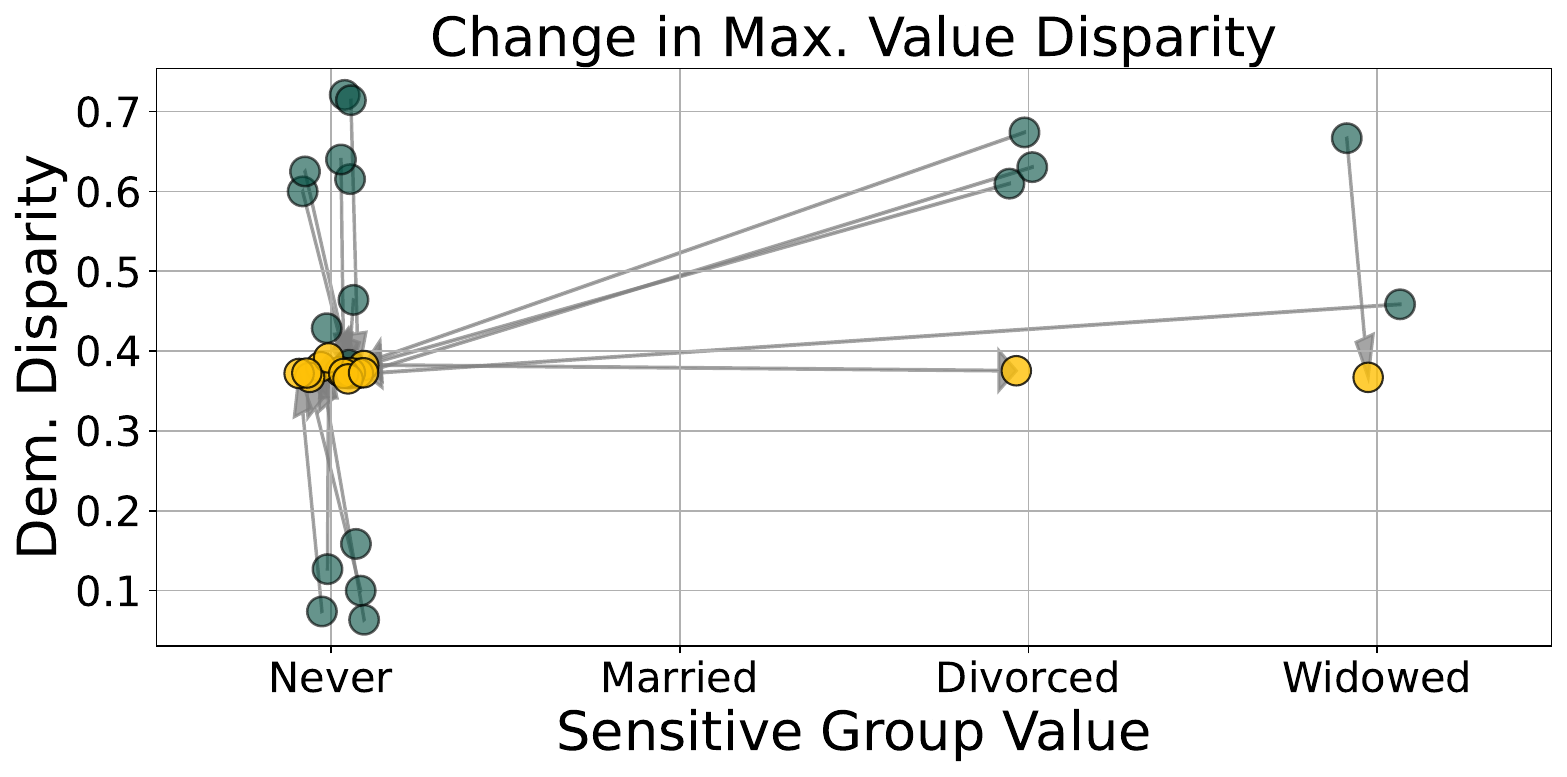}}\\
  \subfloat[\textbf{Value-device:} LR vs. Reweighing]
  {\includegraphics[width=0.44\textwidth]{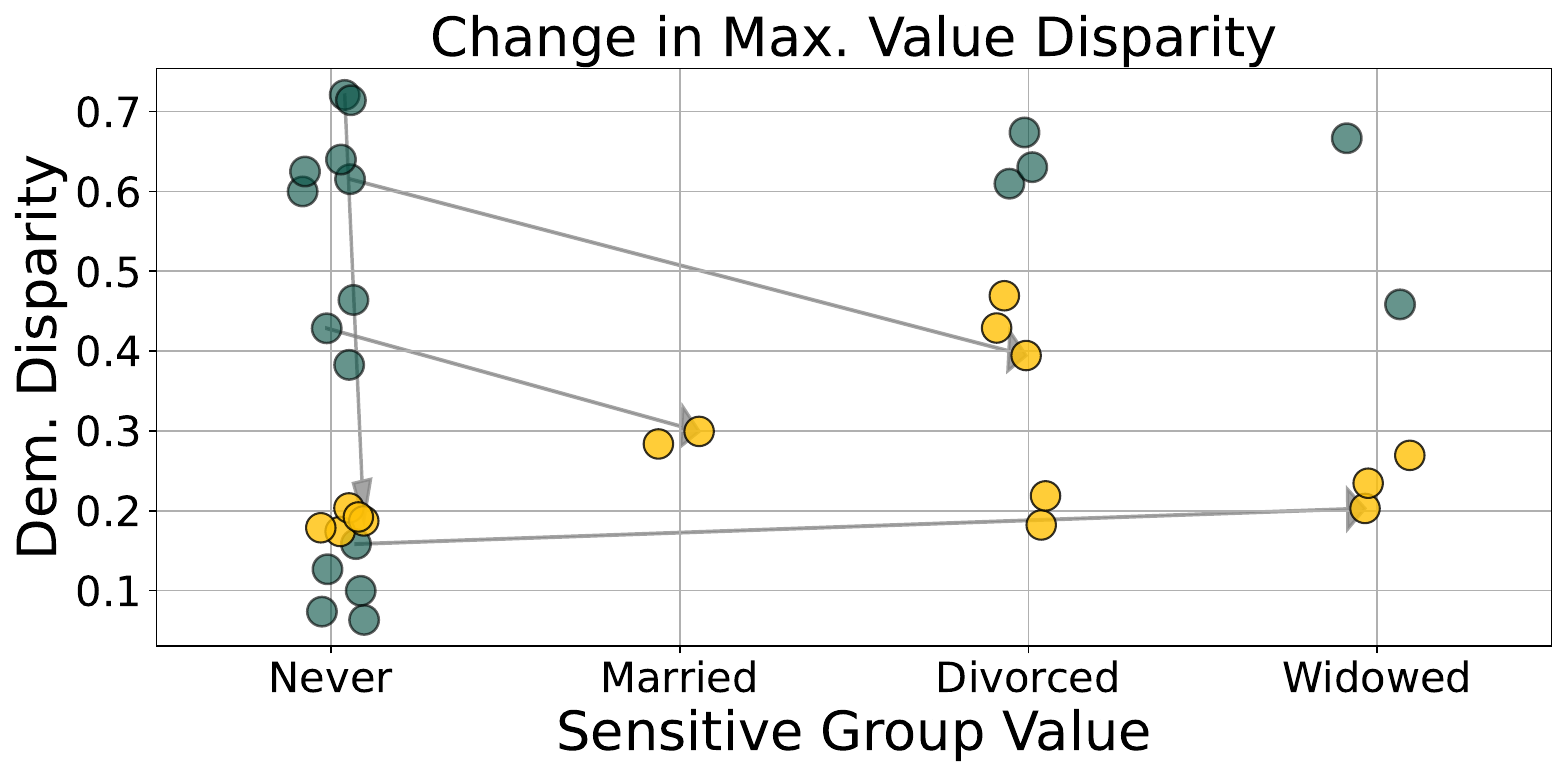}}\\
  \centering\includegraphics[width=0.40\textwidth]{images/legend/value_change_legend.pdf}
  \caption{\val toward \mar as well as value changes measured with Demographic Disparity on the local Logistic Regression (LR) models versus the FedAvg model, the PUFFLE model, and the Reweighing model for the value-device \dutch datasets.}
  \label{fig:before_after_device_value_lr_dutch}
\end{figure}

Figure \ref{fig:before_after_device_value_xgboost_dutch}, and \ref{fig:before_after_device_value_lr_dutch} show the results for the \dutch value-device datasets. Here, we also see that the spread of DD is smaller for FL methods. However, only for FedAvg and Reweighting a shift towards \texttt{Divorced} is visible, while for PUFFLE, a clear shift towards the value \texttt{Never} can be observed.

\subsection{Experiments with \celeba}
\label{app:celeba}

Lastly, to show the compatibility of \fairdataset with image datasets, we also include in our benchmarking suite the four configurations obtained with \celeba as the underlying dataset. 

Due to the higher computation needed to run the Celeba experiments, we did not perform the training of the local models for each of the clients. Therefore, the results that we report are analyses made directly on the federated dataset. 

\begin{figure}
\centering
\includegraphics[width=0.45\linewidth]{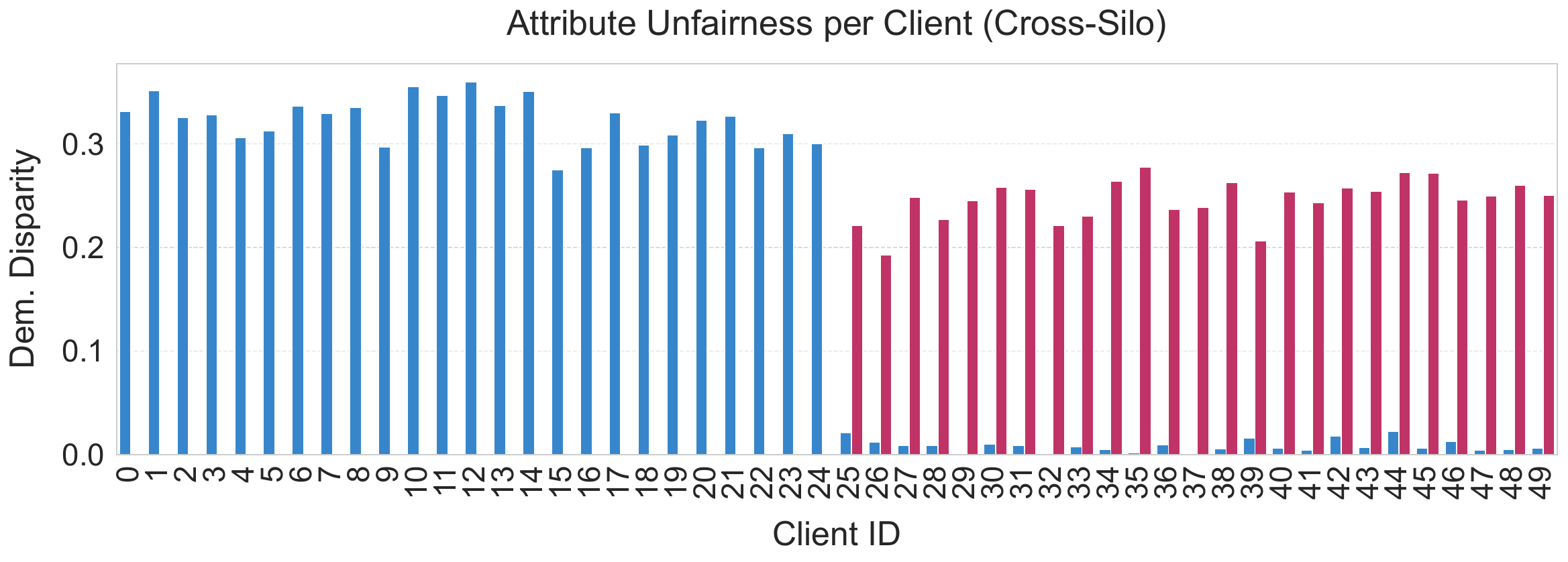}
\includegraphics[width=0.45\linewidth]{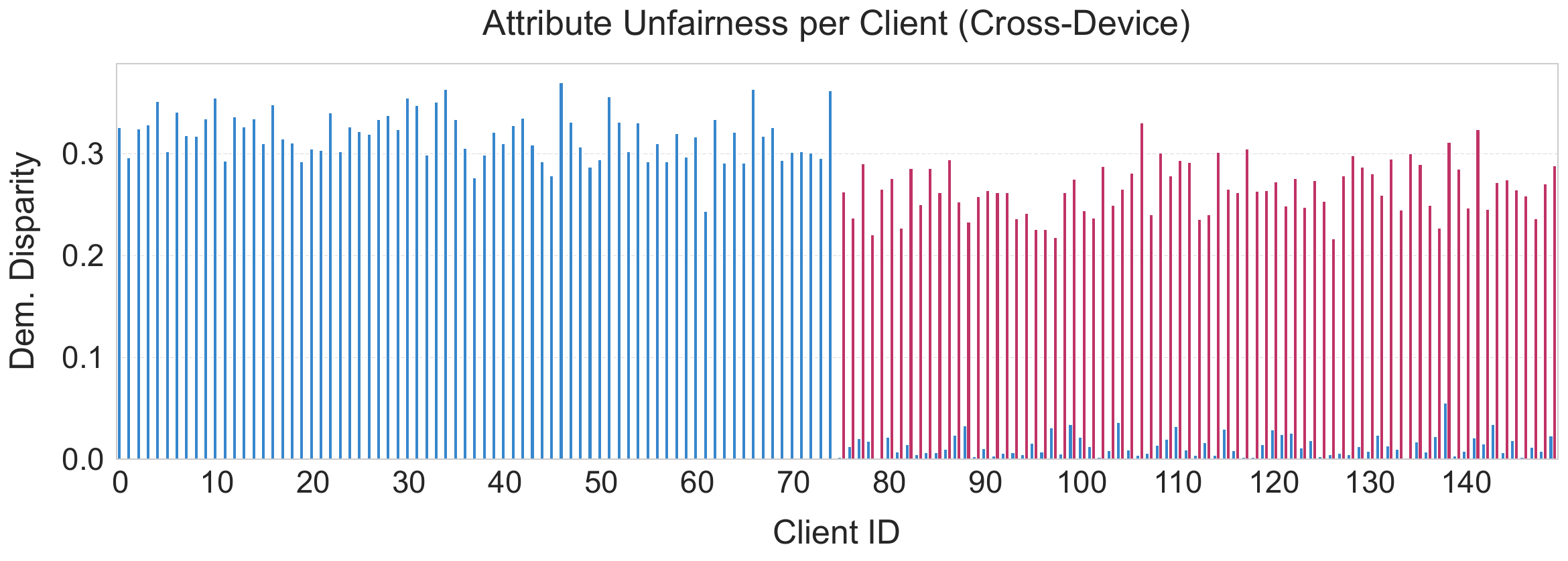}\\
\includegraphics[width=0.2\linewidth]{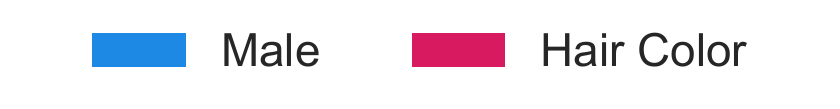}
\caption{Histogram showing the \att distribution, measured with Demographic Disparity on the \celeba dataset in the cross-silo and cross-device scenario. The first half of the clients are unfair toward \sex while the second half toward the attribute \hair.} 
\label{fig:histogram_celeba}
\end{figure}

For the \textbf{attribute-silo} and \textbf{attribute-device} datasets, half of the clients have a higher bias for the \sex sensitive attribute than for \hair; the other half of the clients instead have the opposite setting. This is visualized in Figure~\ref{fig:histogram_celeba}; the clients on the left, in blue, have a higher bias for \sex; however, their bias toward \hair is close to 0. 

\begin{figure}
\centering
\includegraphics[width=0.45\linewidth]{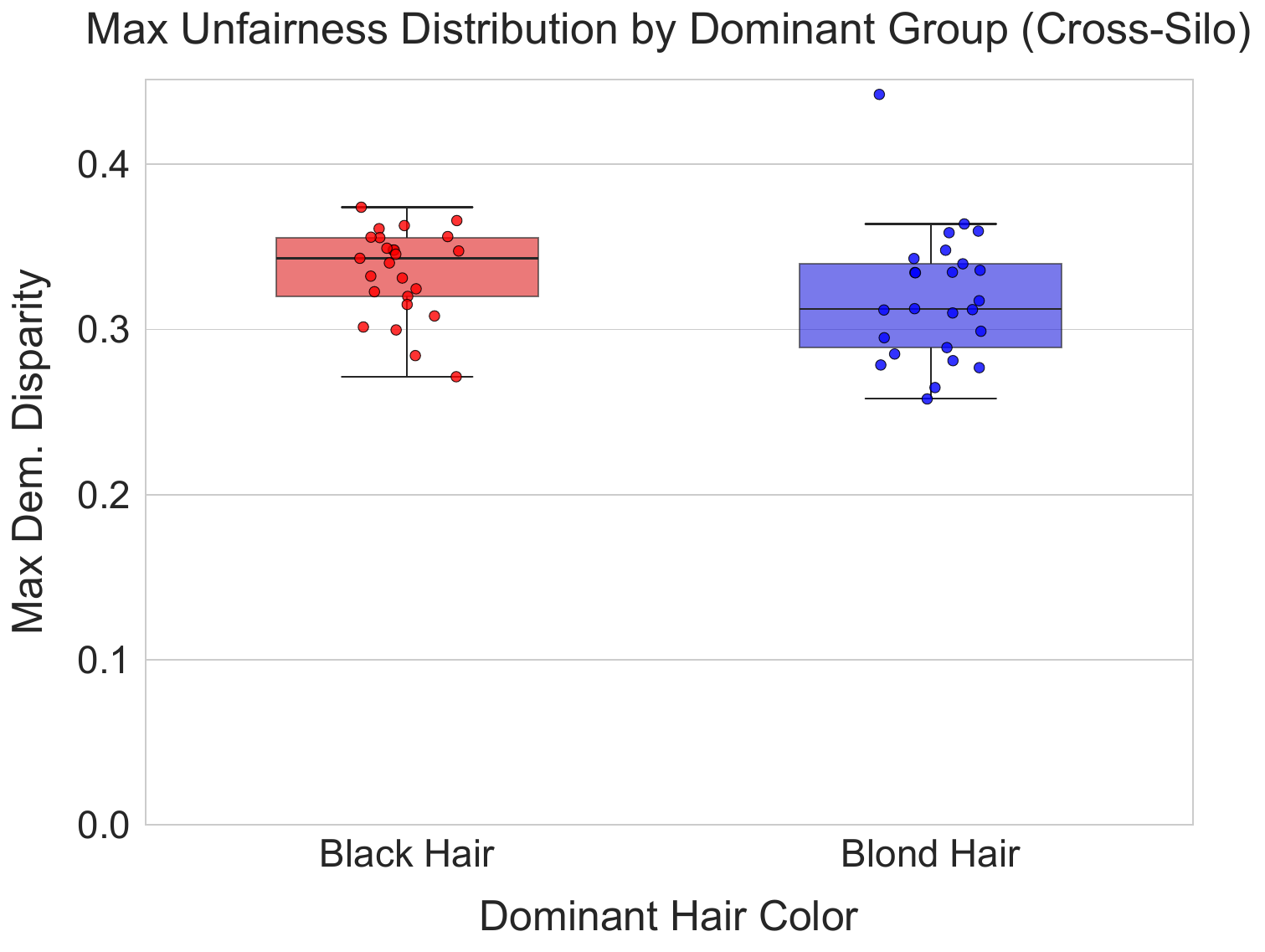}
\includegraphics[width=0.45\linewidth]{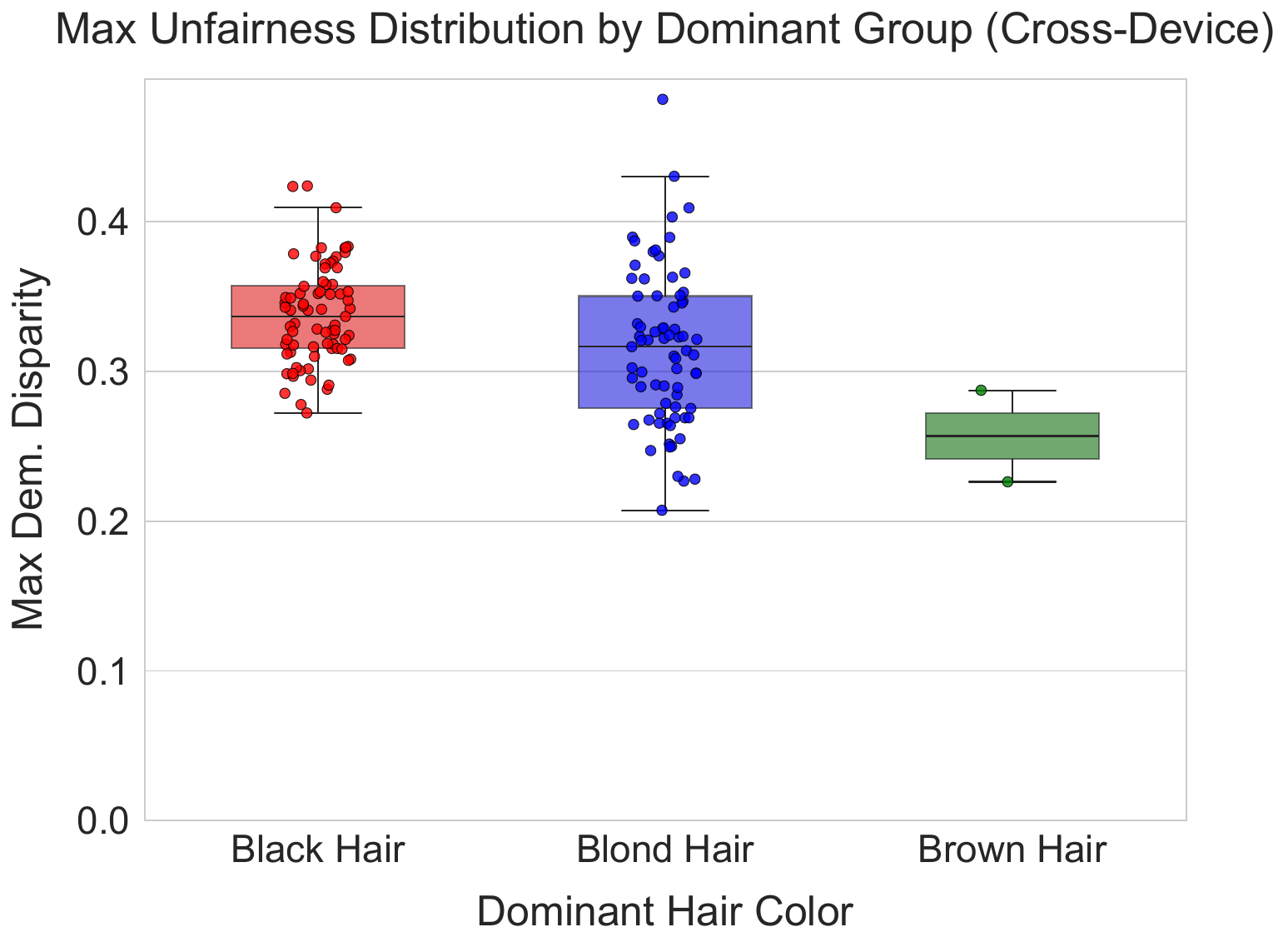}\\
\includegraphics[width=0.4\linewidth]{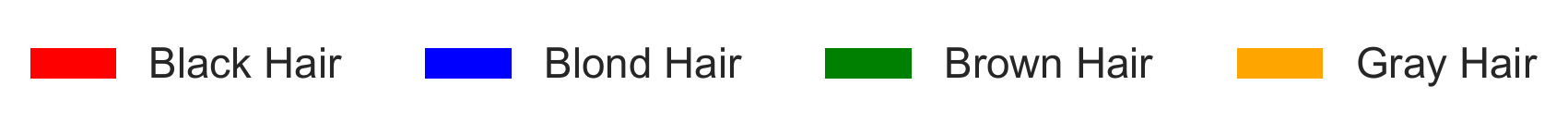}
\caption{Unfairness distribution of the \val experiment in both cross-silo and cross-device on the \celeba dataset.} 
\label{fig:distribution_celeba}
\end{figure}

In Figure~\ref{fig:distribution_celeba}, instead, we report the distribution of the maximum Demographic Disparity in the case of the \textbf{value-silo} and \textbf{value-device} datasets. In this case, we created a scenario in which the maximum Demographic Disparity measured on the clients is, in about half of the cases, the one corresponding to \hairblack and in the other half \hairblond.

\section{Hyperparameter Tuning}\label{app:hyp_tuning}

To perform hyperparameter tuning when training the FedAvg baseline, we performed a Bayesian optimization to maximize the model validation accuracy. The parameters that we optimized are: learning rate, batch size, optimizer, and number of local epochs. 
When training the FL model with PUFFLE\footnote{PUFFLE GitHub repository: \url{https://github.com/lucacorbucci/PUFFLE}}, we followed the suggestions reported by~\citet{puffle}. Therefore, we performed a Bayesian optimization to minimize the model validation accuracy while keeping the model unfairness under $T=0.05$. In this case, we optimized learning rate, batch size, optimizer, number of local epochs, and the value of the $\lambda$ used for the unfairness mitigation.
We trained the FL models for \income and \dutch for 10 FL rounds, while we increased this value to 40 in the case of \celeba. This was needed because learning an image FL model is harder than learning an FL model on tabular data.

\section{Hardware}
\label{app:hardware}

For most of the features offered by \fairdataset, a single CPU is sufficient, and no GPU is necessary. The memory requirements are in line with what is available on most commercial laptops. We believe that most resource-constrained laboratories will be able to create data with \fairdataset. Model fitting on the tabular datasets is similarly cheap.
Moreover, to avoid the need to create datasets, we also provide 12 datasets to download. In the direction of reducing the computational needs and testing the fairness-aware FL in resource-constrained scenarios, it is possible to reduce the size of these fixed datasets by only subsampling a set of client datasets. We provide extensive statistics on each client dataset so that users may make well-founded decisions on which datasets to sample. Additionally, if restrictions on computation exist and practitioners want to create their dataset, it is not necessary to download the full \income datasets, but only, e.g., preselected states.
For all model-fitting experiments presented in this paper involving tabular datasets, we used a Mac Mini with an Apple M4 Processor and 24GB of RAM. In the experiments with images, instead, we used a NVIDIA DGX H100 with 224 CPU Intel(R) Xeon(R) Platinum 8480CL, 8 GPUs Nvidia H100 (80GB), and 2TB of RAM.

\begin{table}[h!]
\centering
\caption{\income attribute-silo, \income attribute-device dataset: Applied bias injections.}
\begin{tabular}{l|c|l|c}
\label{tab:fixds_att_bias_modific}
\textbf{State} & \textbf{$P_{drop}$} & \textbf{Attribute} & \textbf{Value} \\
\midrule
WY & 0.1 & SEX & 2 \\
WI & 0.1 & RAC1P & 2 \\
ND & 0.1 & SEX & 2 \\
CA & 0.1 & RAC1P & 2 \\
MT & 0.1 & RAC1P & 2 \\
LA & 0.1 & SEX & 2 \\
KY & 0.1 & RAC1P & 2 \\
ME & 0.1 & RAC1P & 2 \\
AL & 0.2 & RAC1P & 2 \\
IN & 0.2 & SEX & 2 \\
MS & 0.2 & RAC1P & 2 \\
GA & 0.2 & RAC1P & 2 \\
VT & 0.2 & RAC1P & 2 \\
IL & 0.3 & SEX & 2 \\
WA & 0.3 & SEX & 2 \\
NH & 0.3 & SEX & 2 \\
PA & 0.4 & SEX & 2 \\
WV & 0.4 & SEX & 2 \\
AR & 0.4 & SEX & 2 \\
KS & 0.4 & SEX & 2 \\
OR & 0.4 & RAC1P & 2 \\
TX & 0.4 & SEX & 2 \\
DE & 0.4 & RAC1P & 2 \\
OK & 0.4 & SEX & 2 \\
ID & 0.4 & SEX & 2 \\
MI & 0.5 & SEX & 2 \\
VA & 0.5 & SEX & 2 \\
TN & 0.5 & SEX & 2 \\
OH & 0.5 & SEX & 2 \\
MO & 0.6 & SEX & 2 \\
PR & 0.6 & SEX & 2 \\

\end{tabular}
\end{table}

\begin{table}[h!]
\centering
\caption{\income value-silo, \income value-device dataset: Applied bias injections.}
\begin{tabular}{l|c|l|c}
\label{tab:fixds_value_bias_modific}
\textbf{State} & \textbf{$P_{drop}$} & \textbf{Attribute} & \textbf{Value} \\
\midrule
AZ & 0.1 & RAC1P & 5 \\
OH & 0.1 & RAC1P & 4 \\
AR & 0.2 & RAC1P & 4 \\
MN & 0.2 & RAC1P & 5 \\
OR & 0.2 & RAC1P & 2 \\
WV & 0.2 & RAC1P & 5 \\
DE & 0.3 & RAC1P & 4 \\
LA & 0.3 & RAC1P & 5 \\
NE & 0.3 & RAC1P & 4 \\
AK & 0.5 & RAC1P & 4 \\
MS & 0.5 & RAC1P & 4 \\
PR & 0.6 & RAC1P & 4 \\

\end{tabular}
\end{table}

\begin{table}[h!]
\centering
\caption{\textbf{\income attribute-silo dataset}: Counts and states for which the maximum Demographic Disparity/Equalized Odds Difference across all evaluated models is reached for the stated sensitive attribute. States where bias is distributed the same for both Demographic Disparity/Equalized Odds Difference are marked in bold.}
\label{tab:fixds_cs_att_bias_dis}
\renewcommand{\arraystretch}{1.5}
\begin{tabular}{c|c|p{5cm}|c|p{5cm}}
\multirow{2}{*}{Sensitive att.} & \multicolumn{2}{c|}{\textbf{Demographic Disparity}} & \multicolumn{2}{c}{\textbf{Equalized Odds Difference}} \\
 & Count & States & Count & States \\
\midrule
\sex&21& AR, ID, IL, IN, KS, LA, MI, MO, ND, NH, OH, OK, PA, PR, SD, TN, TX, UT, VA, WA, WV, WY
 &0 \\
\midrule
\race & 29 & \textbf{AK}, \textbf{AL}, \textbf{AZ}, CA, CO, CT, DE, FL, GA, \textbf{HI}, IA, KY, MA, MD, ME, \textbf{MN}, \textbf{MS}, MT,  NC,  \textbf{NE}, NJ, \textbf{NM}, \textbf{NV},  \textbf{NY}, OR, \textbf{RI}, \textbf{SC}, \textbf{VT}, \textbf{WI}
 &17 & \textbf{AK}, \textbf{AL}, \textbf{AZ}, \textbf{HI}, \textbf{MN}, \textbf{MS}, ND, \textbf{NE}, \textbf{NM}, \textbf{NV}, \textbf{NY}, OK, \textbf{RI}, \textbf{SC}, \textbf{VT}, \textbf{WI}, WY
 \\ 
\end{tabular}
\end{table}

\begin{table}[h!]
\centering
\caption{\textbf{\income value-silo dataset}: Counts and states for which the maximum Demographic Disparity/Equalized Odds Difference across all evaluated models is reached for \race and the stated value. States where bias is distributed the same for both Demographic Disparity/Equalized Odds Difference are marked in bold.}
\label{tab:fixds_cs_value_bias_dis}
\renewcommand{\arraystretch}{1.5}
\begin{tabular}{c|c|p{5cm}|c|p{5cm}}
\multirow{2}{*}{Value (\race)} & \multicolumn{2}{c|}{\textbf{Demographic Disparity}} & \multicolumn{2}{c}{\textbf{Equalized Odds Difference}} \\
 & Count & States & Count & States \\
\midrule
1&0& &0& \\
\midrule
2 & 9 & FL, \textbf{ME}, MN, \textbf{MT}, ND, \textbf{NH}, OK, \textbf{SD}, \textbf{VT}
&5&\textbf{ME}, \textbf{MT}, \textbf{NH}, \textbf{SD}, \textbf{VT}
 \\ \midrule
3&2&PR, \textbf{WY}
&2&ND, \textbf{WY}
\\\midrule
4&16&AK, \textbf{AL}, \textbf{CO}, \textbf{CT}, \textbf{DE}, \textbf{GA}, \textbf{HI}, ID, \textbf{KY}, \textbf{MS}, \textbf{NE}, \textbf{NM}, \textbf{OH}, \textbf{OR}, \textbf{PA}, UT
&17&\textbf{AL}, \textbf{CO}, \textbf{CT}, \textbf{DE}, FL, \textbf{GA}, \textbf{HI}, \textbf{KY}, \textbf{MS}, \textbf{NE}, NJ, \textbf{NM}, NY, \textbf{OH}, \textbf{OR}, \textbf{PA}, SC
\\\midrule
5&24&\textbf{AR}, AZ, \textbf{CA}, IA, \textbf{IL}, \textbf{IN}, KS, \textbf{LA}, MA, MD, \textbf{MI}, MO, \textbf{NC}, NJ, \textbf{NV}, NY, \textbf{RI}, SC, TN, \textbf{TX}, VA, WA, \textbf{WI}, \textbf{WV}
&13&\textbf{AR}, \textbf{CA}, ID, \textbf{IL}, \textbf{IN}, \textbf{LA}, \textbf{MI}, \textbf{NC}, \textbf{NV}, \textbf{RI}, \textbf{TX}, \textbf{WI}, \textbf{WV}\\
\end{tabular}
\end{table}

\begin{table}[h!]
\centering
\caption{\textbf{\income attribute-device dataset}: Counts and states for which the maximum Demographic Disparity/Equalized Odds Difference across all evaluated models is reached for the stated sensitive attribute. States where bias is distributed the same for both Demographic Disparity/Equalized Odds Difference are marked in bold.}
\label{tab:fixds_cd_att_bias_dis}

\renewcommand{\arraystretch}{1.5}
\begin{tabular}{c|c|p{5cm}|c|p{5cm}}

\multirow{2}{*}{Sensitive att.} & \multicolumn{2}{c|}{\textbf{Demographic Disparity}} & \multicolumn{2}{c}{\textbf{Equalized Odds Difference}} \\
 & Count & States & Count & States \\\midrule
\sex & 55 &
ID\_0, IL\_2, IL\_3, IL\_4, IN\_0, IN\_2, IN\_3, \textbf{KS\_2}, KS\_4,  LA\_4, MI\_0, MI\_1, MI\_2, MI\_3, MI\_4, MO\_1, MO\_4, ND\_1, \textbf{ND\_4}, NE\_5, NH\_0, NH\_1, \textbf{NH\_4}, NH\_5, OH\_0, OH\_1, OH\_2, OH\_3, OH\_4, OK\_0, OK\_2, OK\_3, PA\_0, PA\_1, PA\_2, PA\_3, PA\_5, TN\_2, TN\_4, TN\_5, TX\_0, TX\_1, TX\_3, TX\_4, TX\_5, UT\_0, UT\_4, UT\_5, VA\_0, VA\_1, VA\_3, VA\_4, WA\_3, WA\_5, WV\_0
&
3 &
 \textbf{KS\_2}, \textbf{ND\_4}, \textbf{NH\_4},
\\\midrule
\race & 56 &
\textbf{AL\_2}, AL\_3, AL\_4, \textbf{AZ\_0}, \textbf{AZ\_5}, CA\_1, \textbf{CO\_0}, \textbf{CO\_3}, \textbf{CT\_2}, CT\_5, FL\_5, \textbf{HI\_0}, \textbf{IA\_0}, \textbf{IA\_1}, \textbf{ID\_2}, LA\_2, MA\_0, \textbf{MA\_4}, MA\_3, \textbf{ME\_1}, \textbf{MN\_0}, \textbf{MN\_3}, \textbf{MN\_4}, MN\_5, \textbf{MS\_4}, \textbf{MS\_5}, NC\_1, \textbf{NE\_3}, \textbf{NE\_4}, NJ\_1, NJ\_3, NJ\_4, \textbf{NM\_1}, \textbf{NM\_5}, NV\_0, \textbf{NV\_1}, NV\_3, NY\_0, \textbf{NY\_1}, NY\_2, \textbf{NY\_3}, NY\_4, \textbf{NY\_5}, OR\_1, \textbf{OR\_3}, \textbf{OR\_4}, \textbf{RI\_2}, \textbf{SC\_0}, SC\_1, SC\_2, SC\_4,  \textbf{SD\_3}, \textbf{UT\_1},  WI\_2, \textbf{WI\_3}, \textbf{WV\_5}
&
39 &
\textbf{AL\_2}, \textbf{AZ\_0}, \textbf{AZ\_5}, \textbf{CO\_0}, \textbf{CO\_3}, \textbf{CT\_2}, \textbf{HI\_0}, \textbf{IA\_0}, \textbf{IA\_1}, \textbf{ID\_2}, \textbf{MA\_4}, \textbf{ME\_1}, MI\_2, \textbf{MN\_0}, \textbf{MN\_3}, \textbf{MN\_4}, \textbf{MS\_4}, \textbf{MS\_5}, \textbf{NE\_3}, \textbf{NE\_4}, NH\_0, NH\_5, \textbf{NM\_1}, \textbf{NM\_5}, \textbf{NV\_1}, \textbf{NY\_1}, \textbf{NY\_3}, \textbf{NY\_5}, OK\_3, \textbf{OR\_3}, \textbf{OR\_4}, \textbf{PA\_3}, \textbf{RI\_2}, \textbf{SC\_0}, \textbf{SD\_3}, TN\_2, UT\_5, \textbf{WI\_3}, \textbf{WV\_5}
\\
\end{tabular}
\end{table}

\begin{table}[h!]
\centering
\caption{\textbf{\income value-device dataset}: Counts and states for which the maximum Demographic Disparity/Equalized Odds Difference across all evaluated models is reached for race and the stated value. States where bias is distributed the same for both Demographic Disparity/Equalized Odds Difference are marked in bold.}
\label{tab:fixds_cd_value_bias_dis}
\renewcommand{\arraystretch}{1.5}
\begin{tabular}{c|c|p{5cm}|c|p{5cm}}
\multirow{2}{*}{Value (\race)} & \multicolumn{2}{c|}{\textbf{Demographic Disparity}} & \multicolumn{2}{c}{\textbf{Equalized Odds Difference}} \\
 & Count & States & Count & States \\
\midrule
1&1& WV\_2&0& \\
\midrule
2 & 15 & \textbf{ID\_1}, \textbf{ME\_2}, MI\_1, MN\_0, MN\_1, \textbf{MN\_2}, \textbf{MT\_2}, \textbf{ND\_0}, \textbf{ND\_1}, \textbf{NH\_2}, \textbf{SD\_1}, \textbf{SD\_2}, \textbf{VT\_0}, WI\_0, \textbf{WY\_0} &17&HI\_2, ID\_0, \textbf{ID\_1}, KY\_0, ME\_1, \textbf{ME\_2}, \textbf{MN\_2}, \textbf{MT\_2}, \textbf{ND\_0}, \textbf{ND\_1}, NE\_0, \textbf{NH\_2}, PA\_0, \textbf{SD\_1}, \textbf{SD\_2}, \textbf{VT\_0}, \textbf{WY\_0}
 \\ \midrule
3&6&AK\_0, \textbf{AR\_2}, \textbf{MT\_0}, MT\_1, NE\_1, \textbf{VT\_1}&4&\textbf{AR\_2}, \textbf{MT\_0}, NE\_2, \textbf{VT\_1}
\\\midrule
4&31&\textbf{AK\_1}, AK\_2, \textbf{CO\_1}, \textbf{CT\_1}, \textbf{CT\_2}, \textbf{DE\_0}, \textbf{DE\_1}, \textbf{GA\_1}, \textbf{GA\_2}, \textbf{HI\_0}, \textbf{HI\_1}, \textbf{ID\_2}, \textbf{IL\_0}, \textbf{IL\_1}, \textbf{KS\_0}, \textbf{KY\_2}, LA\_1, MA\_0, ME\_1, \textbf{MI\_2}, \textbf{MO\_0}, \textbf{MO\_2}, \textbf{MS\_0}, \textbf{MS\_1}, NE\_0, NJ\_2, NM\_0, \textbf{NY\_0}, \textbf{RI\_0}, \textbf{SC\_2}, \textbf{TN\_2}
&35&\textbf{AK\_1}, AZ\_2, \textbf{CO\_1}, CO\_2, \textbf{CT\_1}, \textbf{CT\_2}, \textbf{DE\_0}, \textbf{DE\_1}, GA\_0, \textbf{GA\_1}, \textbf{GA\_2}, \textbf{HI\_0}, \textbf{HI\_1}, \textbf{ID\_2}, \textbf{IL\_0}, \textbf{IL\_1}, IN\_2, \textbf{KS\_0}, \textbf{KY\_2}, \textbf{MI\_2}, \textbf{MO\_0}, \textbf{MO\_2}, \textbf{MS\_0}, \textbf{MS\_1}, NC\_1, \textbf{NY\_0}, NY\_1, OH\_0, \textbf{RI\_0}, \textbf{SC\_2}, TN\_0, \textbf{TN\_2}, VA\_2, WI\_0, WV\_0
\\\midrule
5&47&AL\_0, AL\_2, AR\_0, \textbf{AR\_1}, AZ\_0, \textbf{AZ\_1}, AZ\_2, \textbf{CA\_1}, \textbf{CA\_2}, CO\_2, \textbf{DE\_2}, FL\_0, GA\_0, HI\_2, ID\_0, \textbf{IN\_0}, IN\_2, KS\_2, KY\_0, \textbf{MD\_0}, \textbf{MD\_1}, \textbf{MD\_2}, MO\_1, NC\_0, NC\_1, NC\_2, NE\_2, NJ\_1, NV\_0, \textbf{NV\_2}, NY\_1, NY\_2, OH\_0, PA\_0, PA\_2, RI\_2, \textbf{SC\_1}, TN\_0, TX\_0, \textbf{TX\_1}, \textbf{TX\_2}, VA\_2, WA\_1, WA\_2, \textbf{WI\_1}, \textbf{WI\_2}, WV\_0
&17&\textbf{AR\_1}, \textbf{AZ\_1}, \textbf{CA\_1}, \textbf{CA\_2}, \textbf{DE\_2}, \textbf{IN\_0}, MA\_0, \textbf{MD\_0}, \textbf{MD\_1}, \textbf{MD\_2}, \textbf{NV\_2}, \textbf{SC\_1}, \textbf{TX\_1}, \textbf{TX\_2}, \textbf{WI\_1}, \textbf{WI\_2}, WV\_2
\\

\end{tabular}
\end{table}

\begin{table}[]
\caption{\textbf{\celeba:} Parameter settings for unfairness injection. All configurations set the group parameters of $P_{drop}$ to $\mu=0.3, \sigma=0.05$ and for $P_{flip}$ we set $\sigma=0.02$.}
\label{tab:celeba_settings}
\renewcommand{\arraystretch}{1.3}
\begin{tabular}{llcc}
\toprule
\textbf{Dataset Scenario} & \textbf{Bias Configuration} & \textbf{Primary $P_{flip}$ ($\mu$)} & \textbf{Secondary $P_{flip}$ ($\mu$)} \\
\midrule
\textbf{Attribute-device/attribute-silo} & \sex vs. \hair & 0.28 (\sex) & 0.60 (\hair) \\
\midrule
\textbf{Value-device/value-silo} & \hairblack vs. \hairblond & 0.35 (\hairblack) &0.35 (\hairblond) \\
 
\bottomrule
\end{tabular}
\centering
\end{table}

\begin{figure}[t]
\subfloat[\textbf{Attribute-silo:} LR vs. FedAvg]{\includegraphics[width=0.16\textwidth]{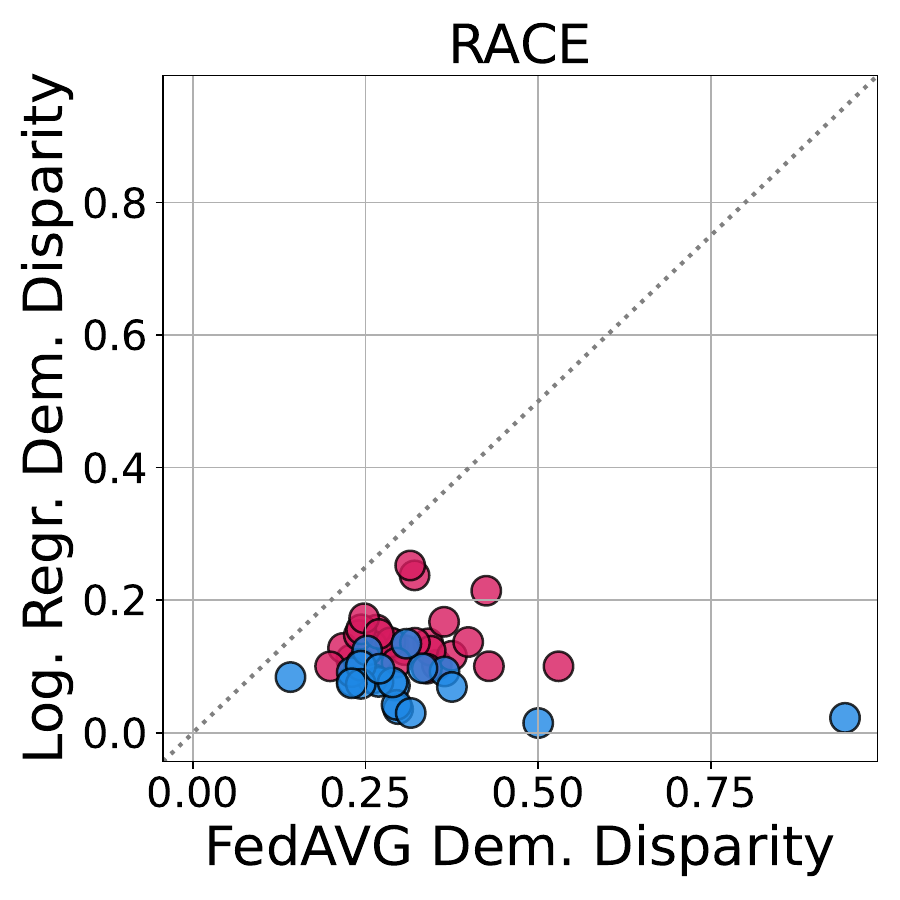}\includegraphics[width=0.16\textwidth]{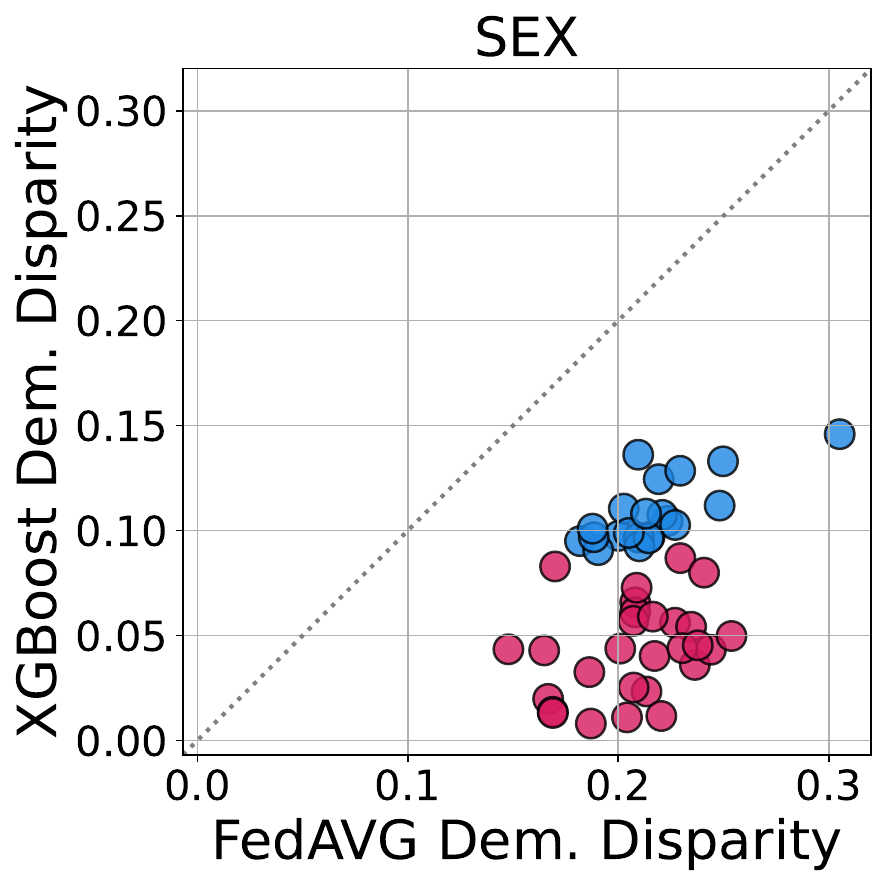}}
  \subfloat[\textbf{Attribute-silo:} LR vs. PUFFLE.]{\includegraphics[width=0.16\textwidth]{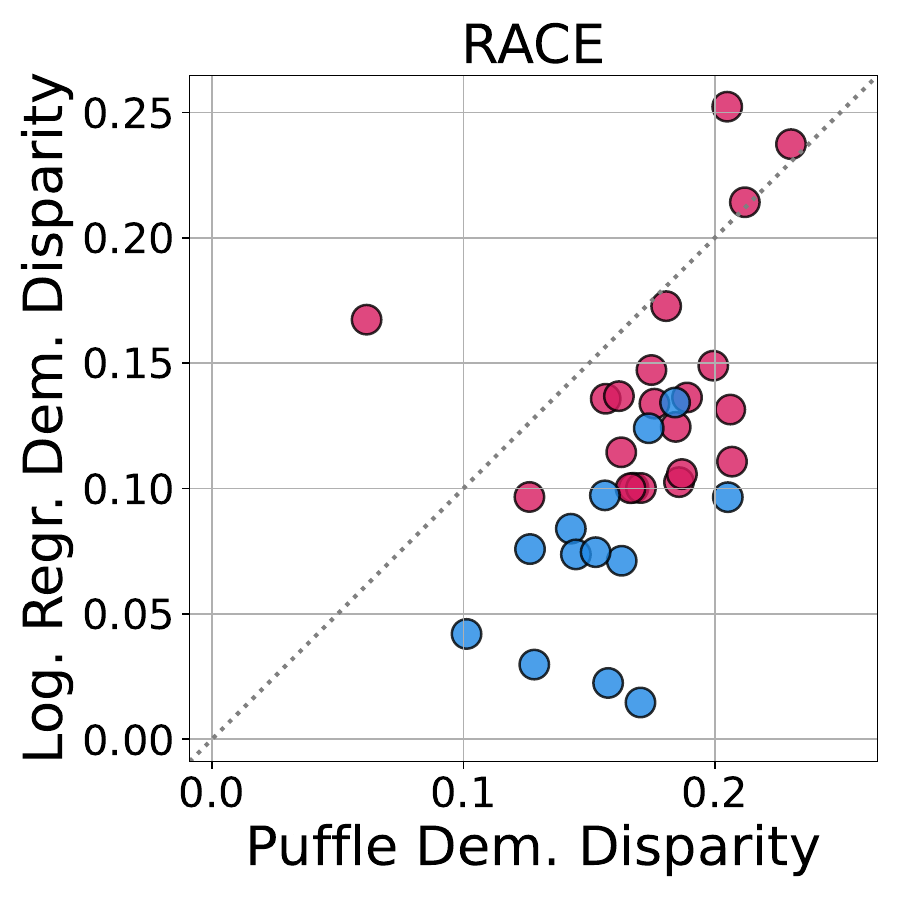}\includegraphics[width=0.16\textwidth]{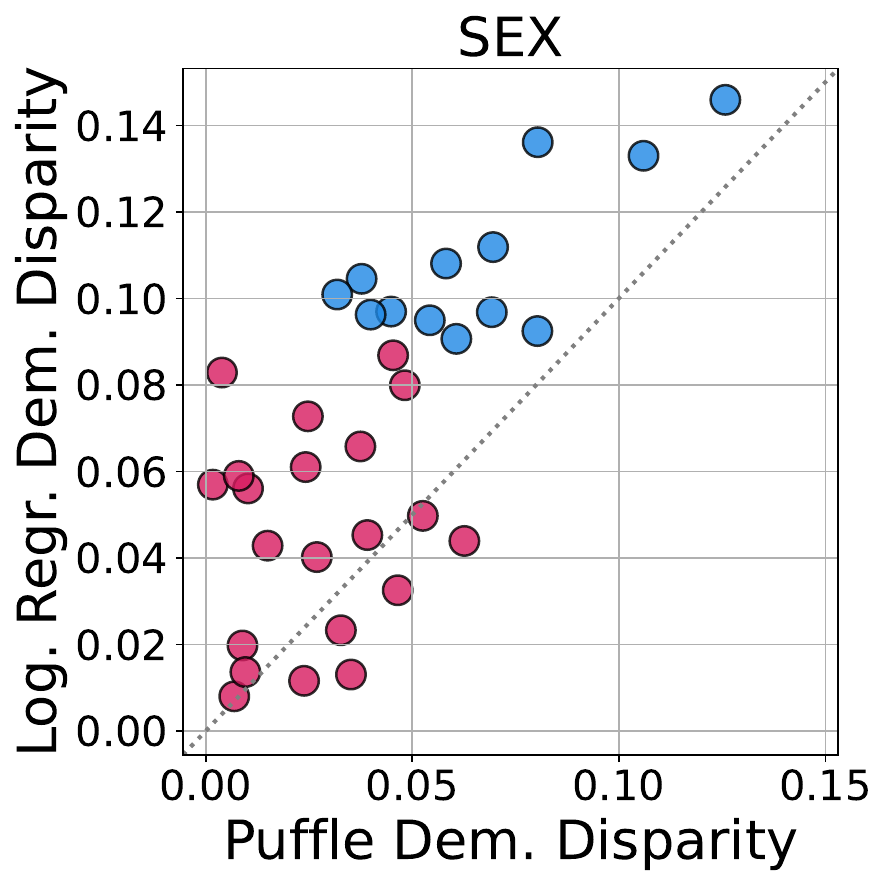}}
  \subfloat[\textbf{Attribute-silo:} LR vs. Reweighing.]
{\includegraphics[width=0.16\textwidth]{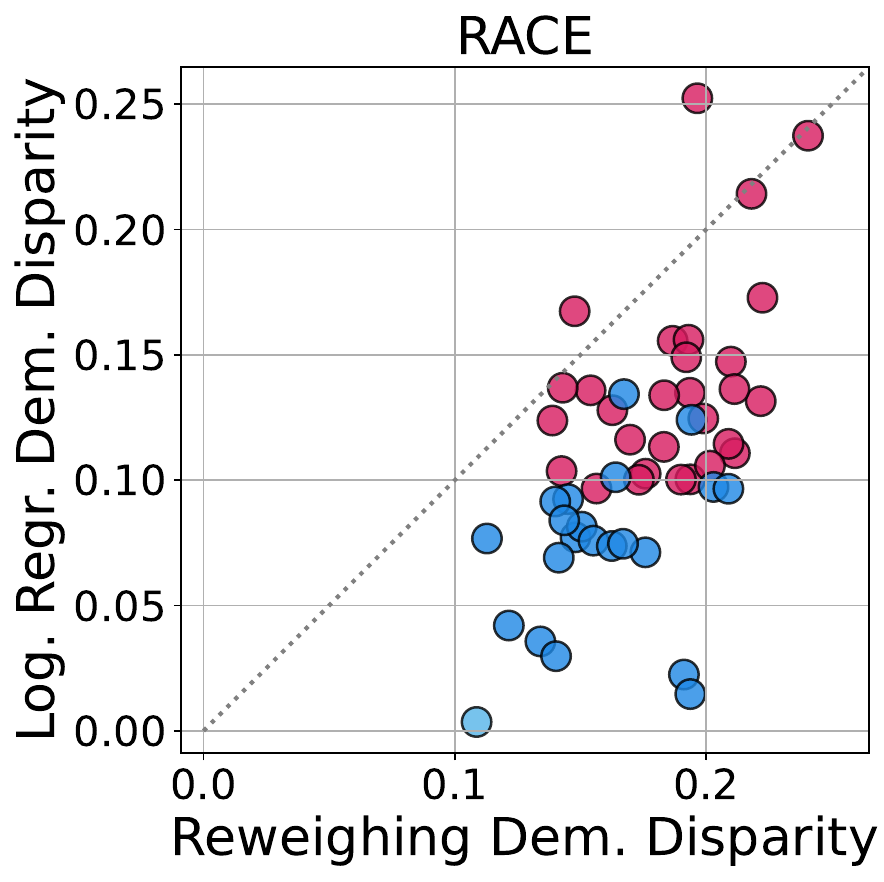}\includegraphics[width=0.16\textwidth]{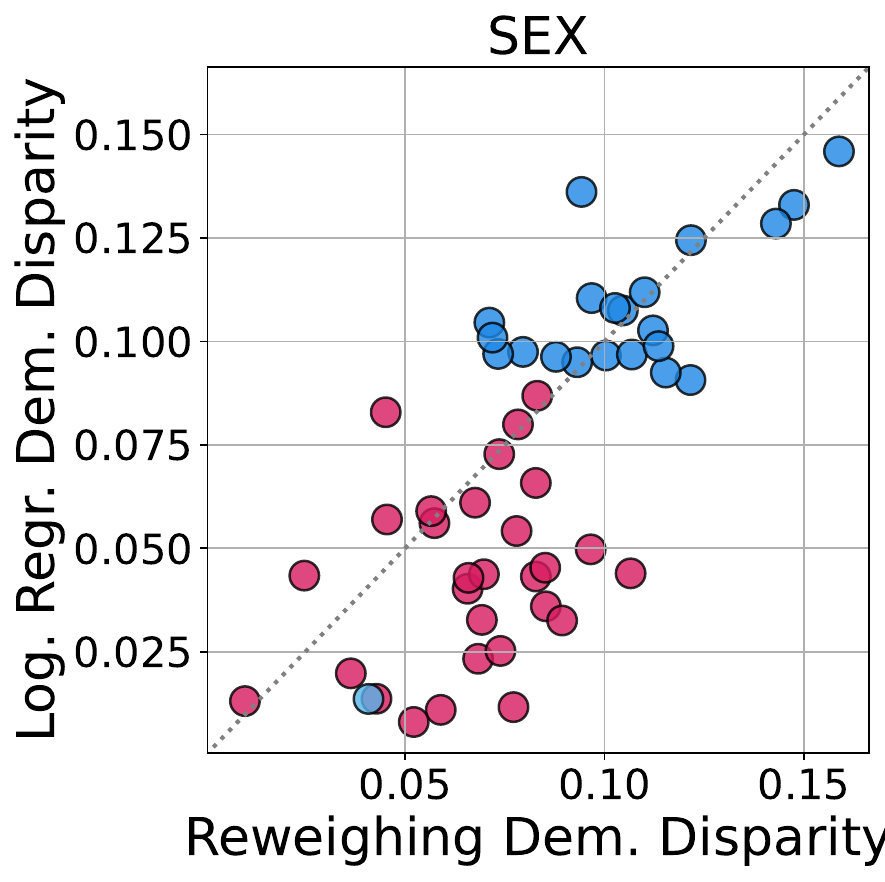}}\\

\centering\includegraphics[width=0.40\textwidth]{images/legend/legend_blue_red.pdf}
  \caption{\att toward \race and \sex measured with Demographic Disparity on the local Logistic Regression (LR) models versus the FedAvg model, the PUFFLE model, and the Reweighing model for the attribute-silo \income dataset.}
  \label{fig:before_after_attribute_silo_lr}
\end{figure}


\begin{figure}   
\subfloat[\textbf{Value-silo:} LR vs. FedAvg]{\includegraphics[width=0.44\textwidth]{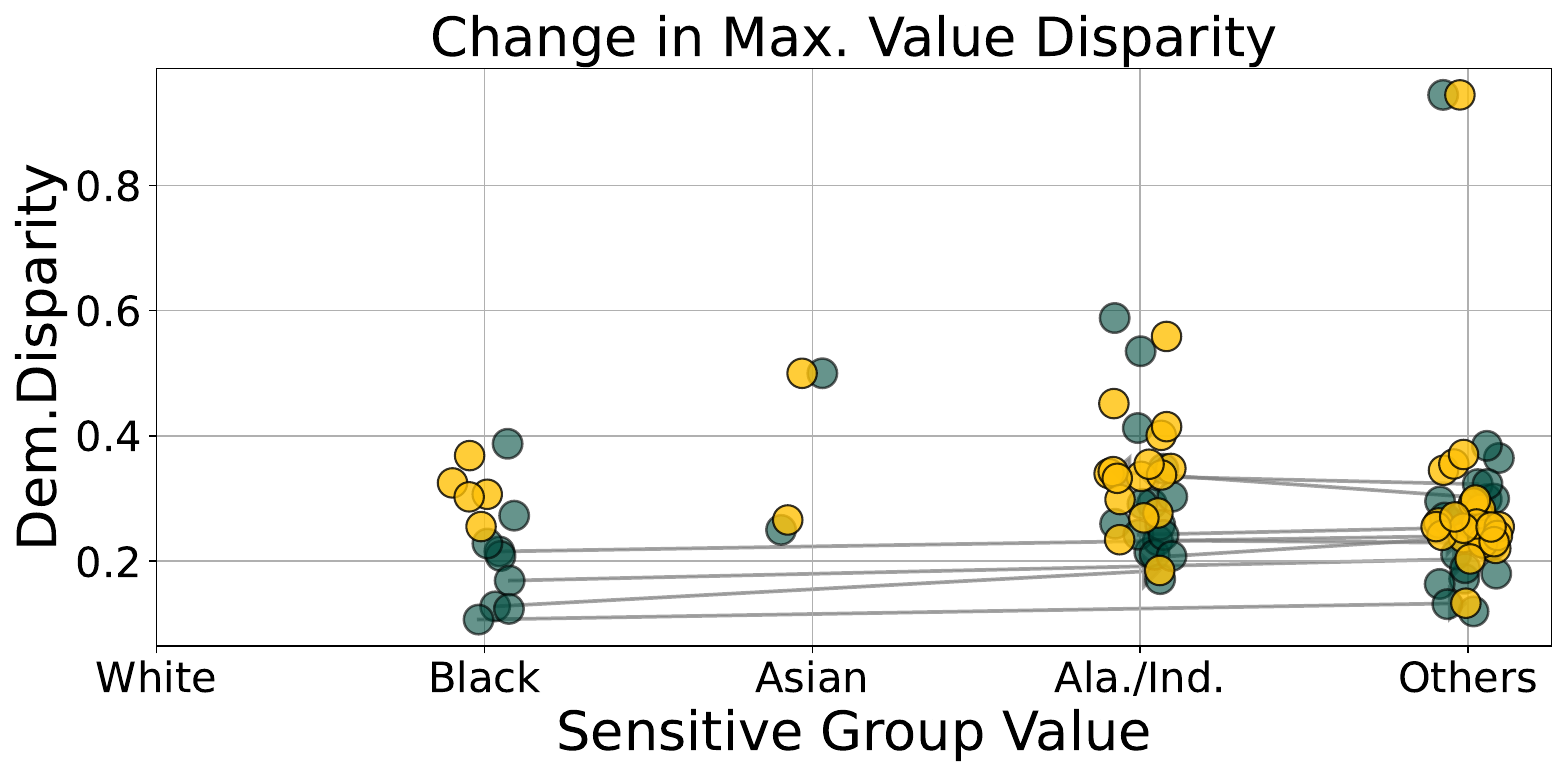}\label{fig:before_after_cross_silo_value}}
  \subfloat[\textbf{Value-silo:} LR vs. PUFFLE]{\includegraphics[width=0.44\textwidth]{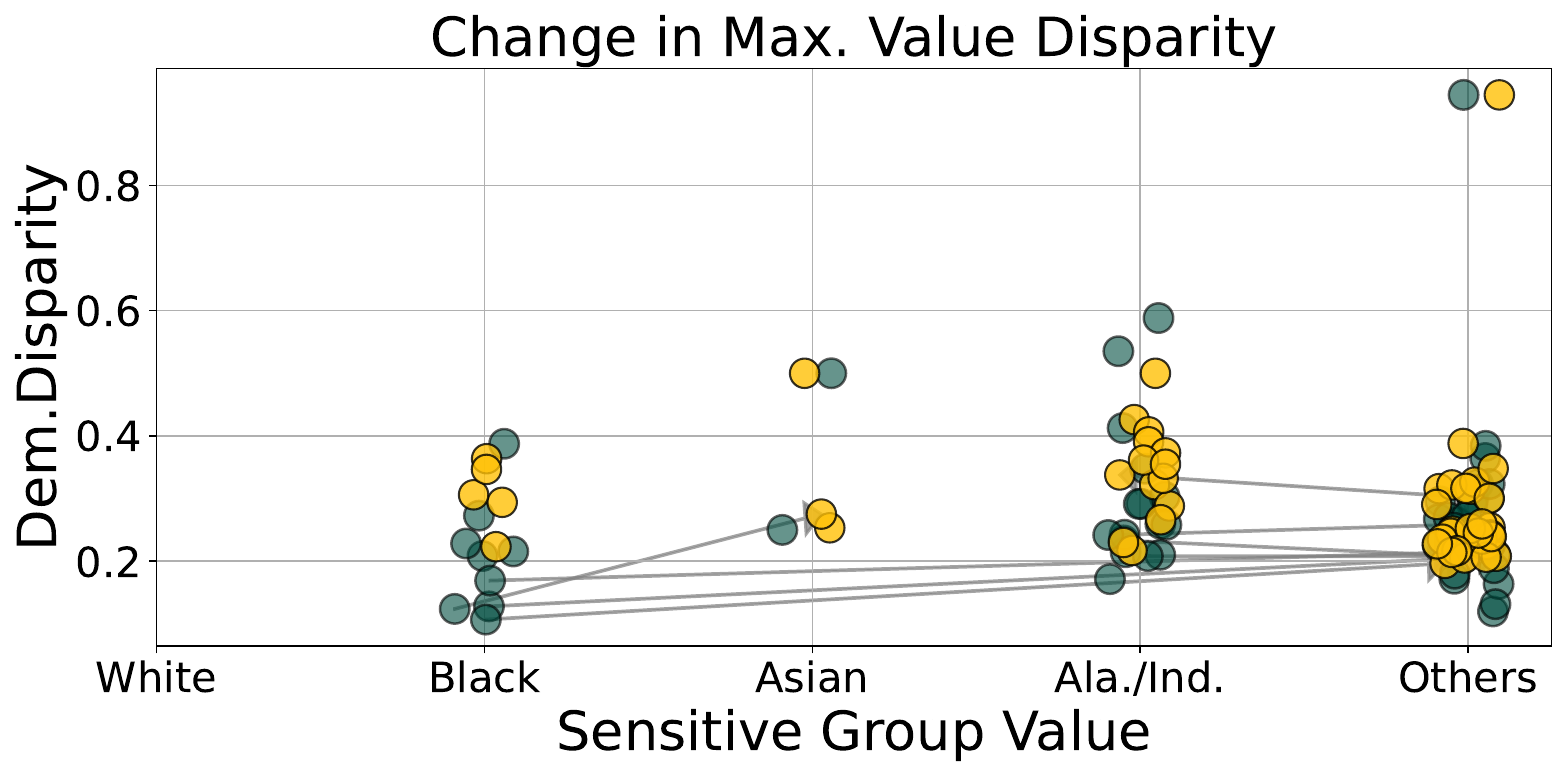}}\\
  \subfloat[\textbf{Value-silo:} LR vs. Reweighing]
  {\includegraphics[width=0.44\textwidth]{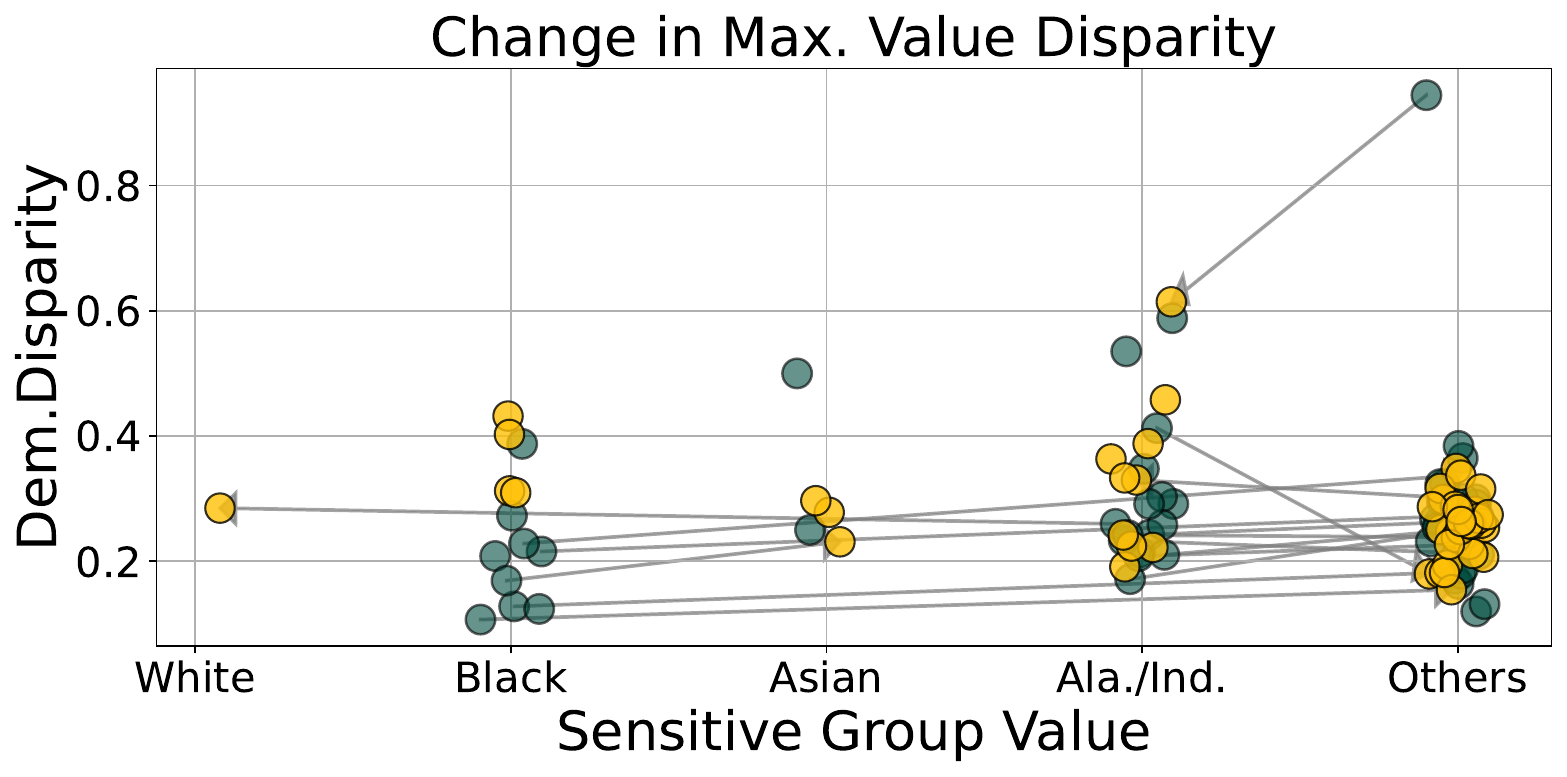}}\\
  \centering\includegraphics[width=0.40\textwidth]{images/legend/value_change_legend.pdf}
  \caption{\val toward \race as well as value changes measured with Demographic Disparity on the local Logistic Regression (LR) models versus the FedAvg model, the PUFFLE model, and the Reweighing model for the value-silo \income datasets.}
  \label{fig:before_after_silo_value_lr}
\end{figure}


\begin{figure}
\subfloat[\textbf{Attribute-silo:} LR vs. FedAvg]{\includegraphics[width=0.49\textwidth]{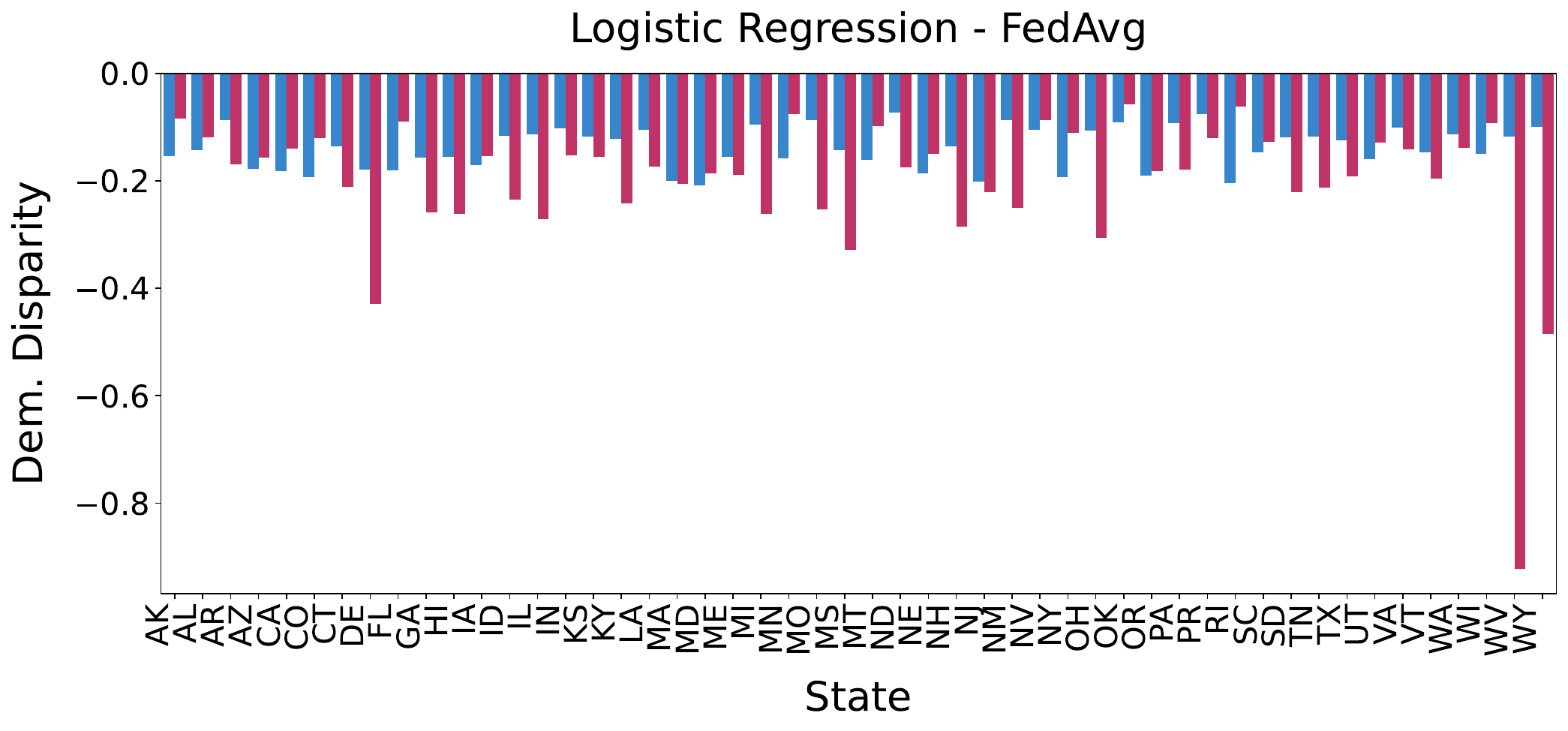}\label{fig:appendix_bar_plot_cross_silo_lr_attribute}}
\subfloat[\textbf{Attribute-silo:} LR vs. PUFFLE]{\includegraphics[width=0.49\textwidth]{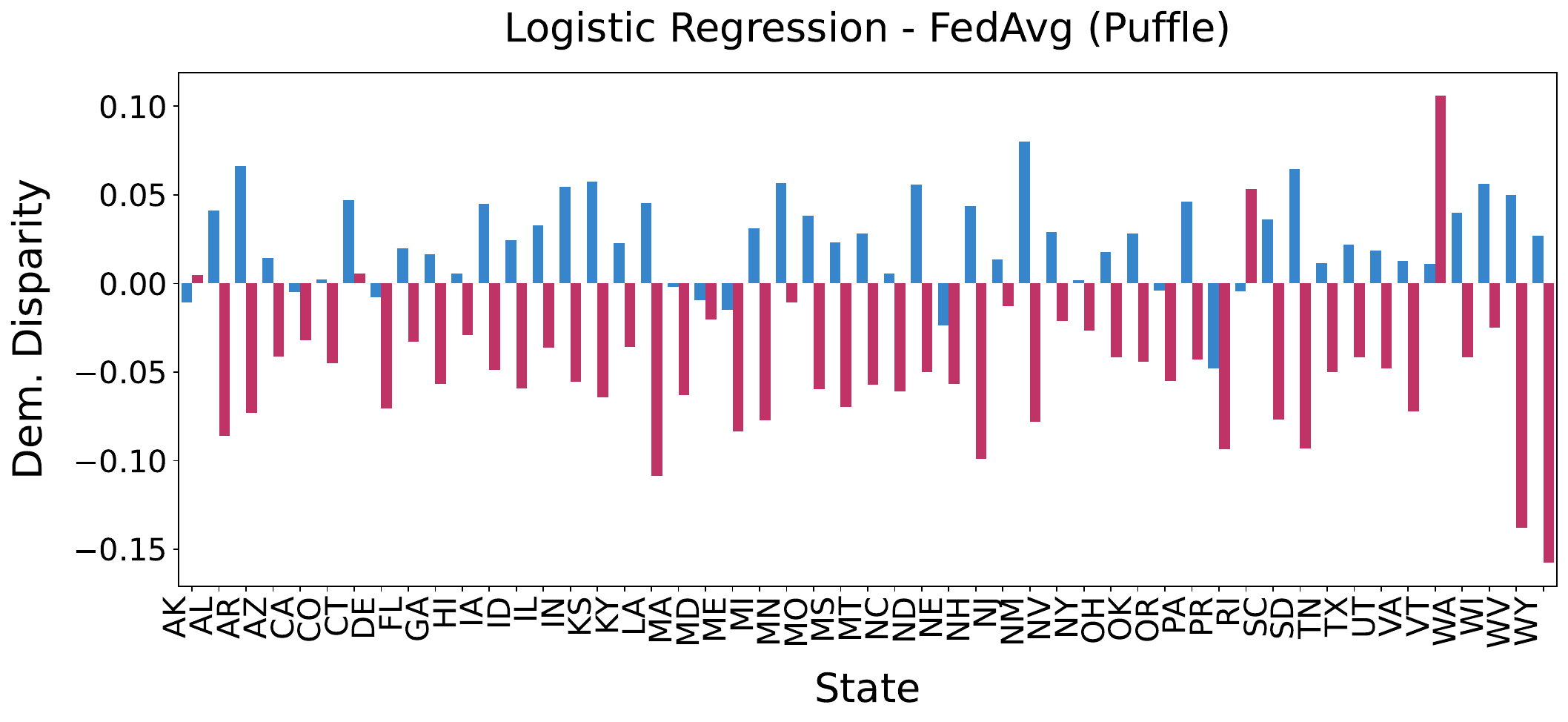}\label{fig:appendix_bar_plot_cross_silo_logistic_attribute_puffle}}\\
\subfloat[\textbf{Attribute-silo} LR vs. Reweighing]{\includegraphics[width=0.49\textwidth]{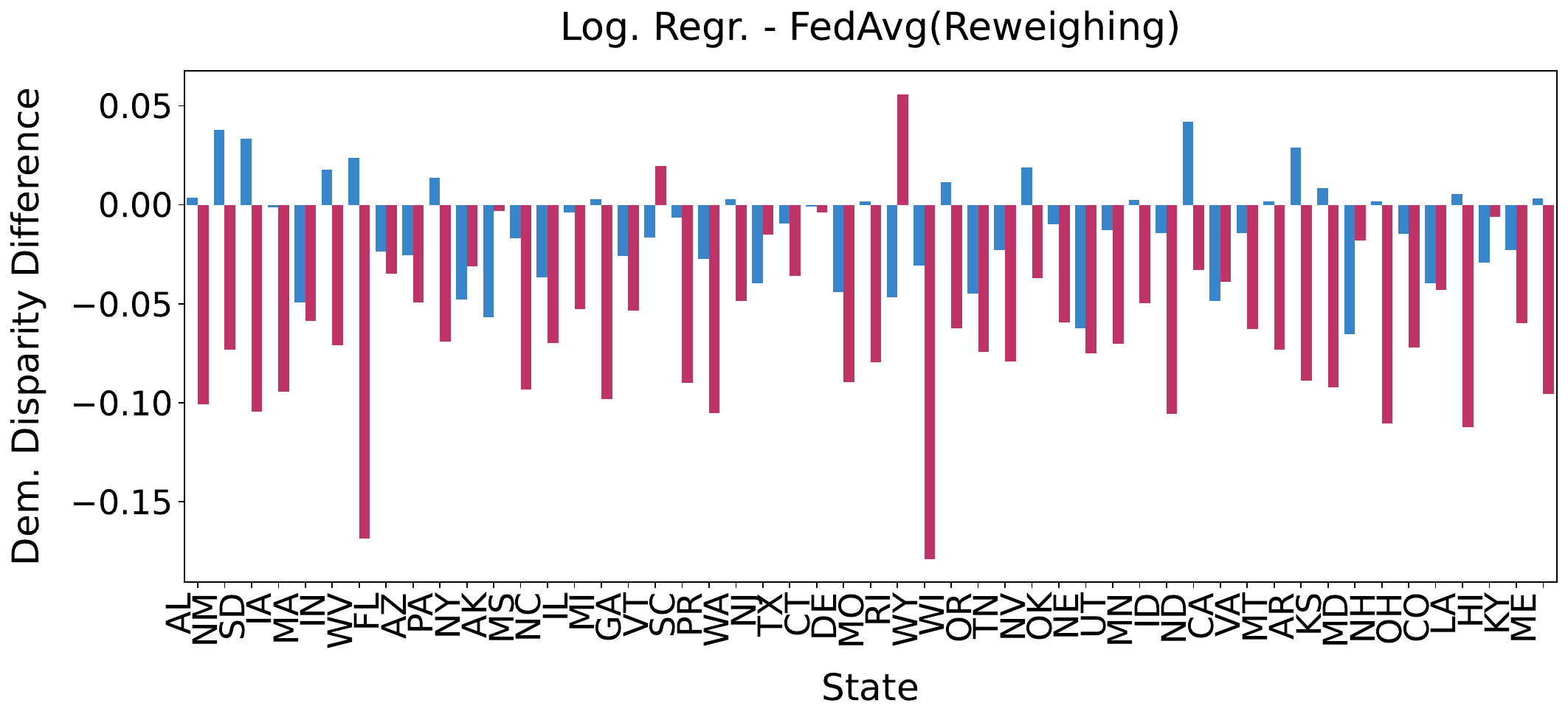}\label{fig:appendix_bar_plot_cross_silo_logistic_attribute_reweighing}}\\
\centering\includegraphics[width=0.4\textwidth]{images/legend/bar_plot_differences_legend.pdf}
  \caption{\income: Individual, per attribute, bias differences in Demographic Disparity between the local Logistic Regression (LR) models versus the FedAvg model, the PUFFLE model, and the Reweighing model.}
  \label{fig:bar_atribute_silo_lr}
\end{figure}

\begin{figure}
\subfloat[\textbf{Value-silo:} XGB vs. FedAvg]{\includegraphics[width=0.49\textwidth]{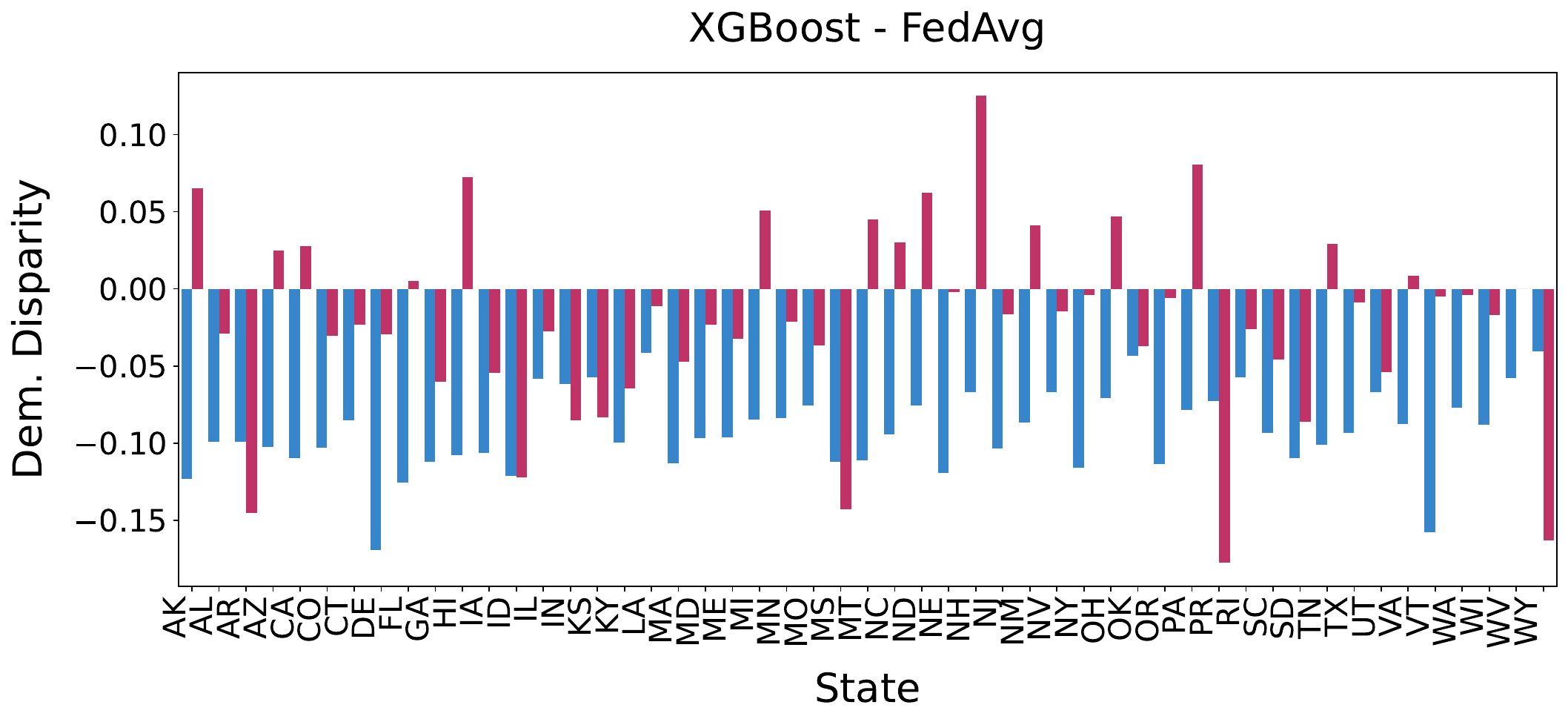}\label{fig:appendix_bar_plot_cross_silo_xgb_value}}
\subfloat[\textbf{Value-silo:} XGB vs. PUFFLE]{\includegraphics[width=0.49\textwidth]{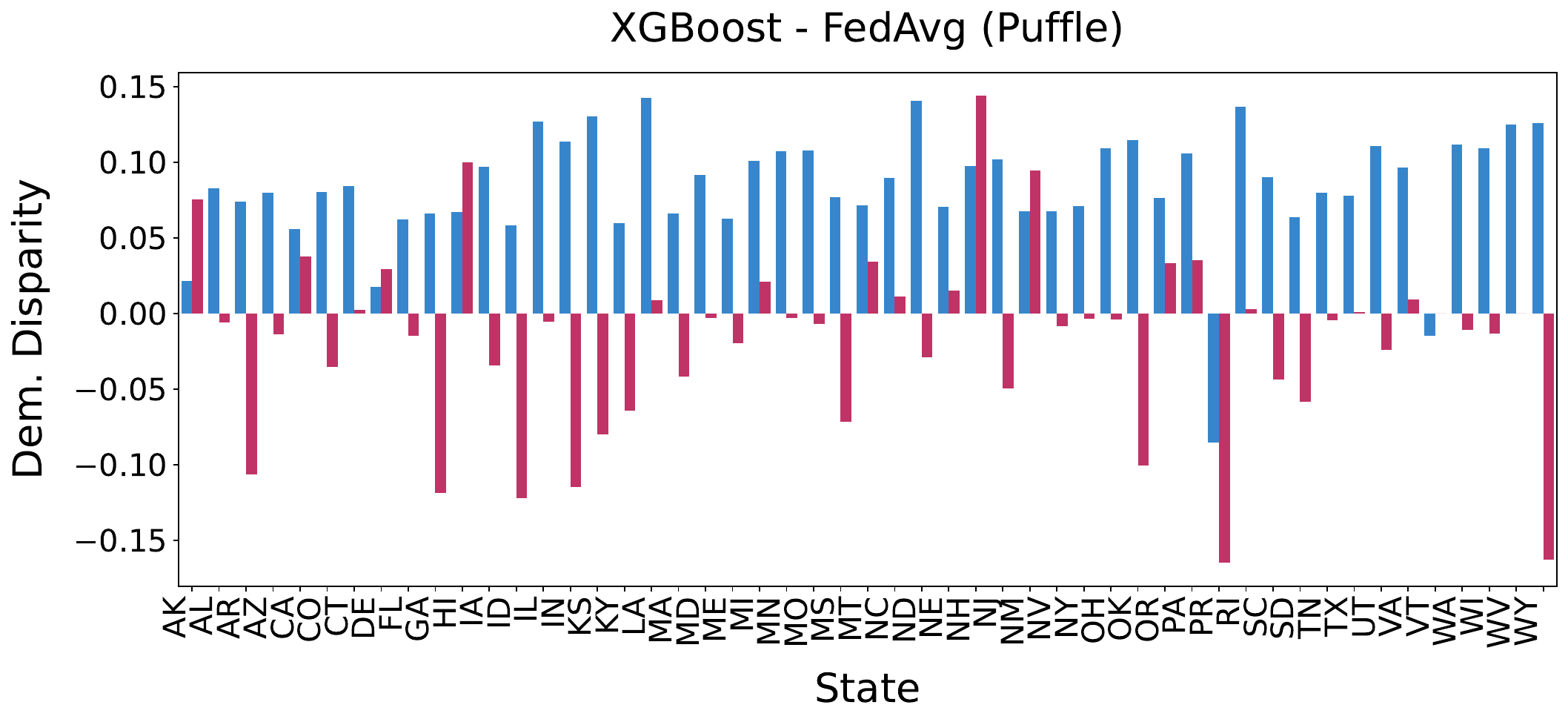}\label{fig:appendix_bar_plot_cross_silo_xgb_attribute_puffle}}\\
\subfloat[\textbf{Value-silo:} XGB vs. Reweighing]{\includegraphics[width=0.49\textwidth]{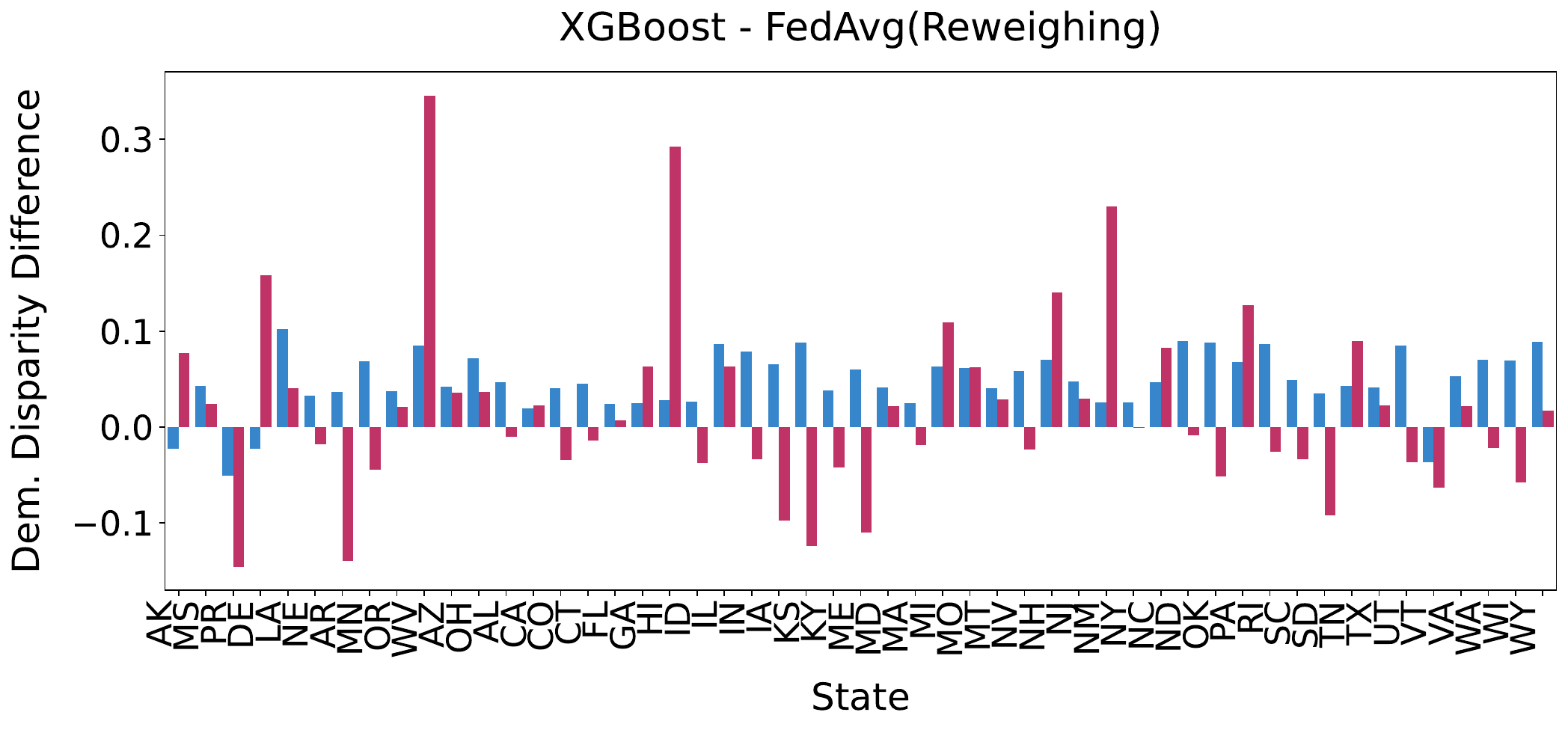}}\\
\centering\includegraphics[width=0.4\textwidth]{images/legend/bar_plot_differences_legend.pdf}
  \caption{Individual, per attribute, bias differences in Demographic Disparity between the local XGBoost models versus the FedAvg model, the PUFFLE model, and the Reweighing model for the value-silo dataset.}
  \label{fig:bar_value_silo}
\end{figure}

\begin{figure}
\subfloat[\textbf{Value-silo} LR vs. FedAvg]{\includegraphics[width=0.49\textwidth]{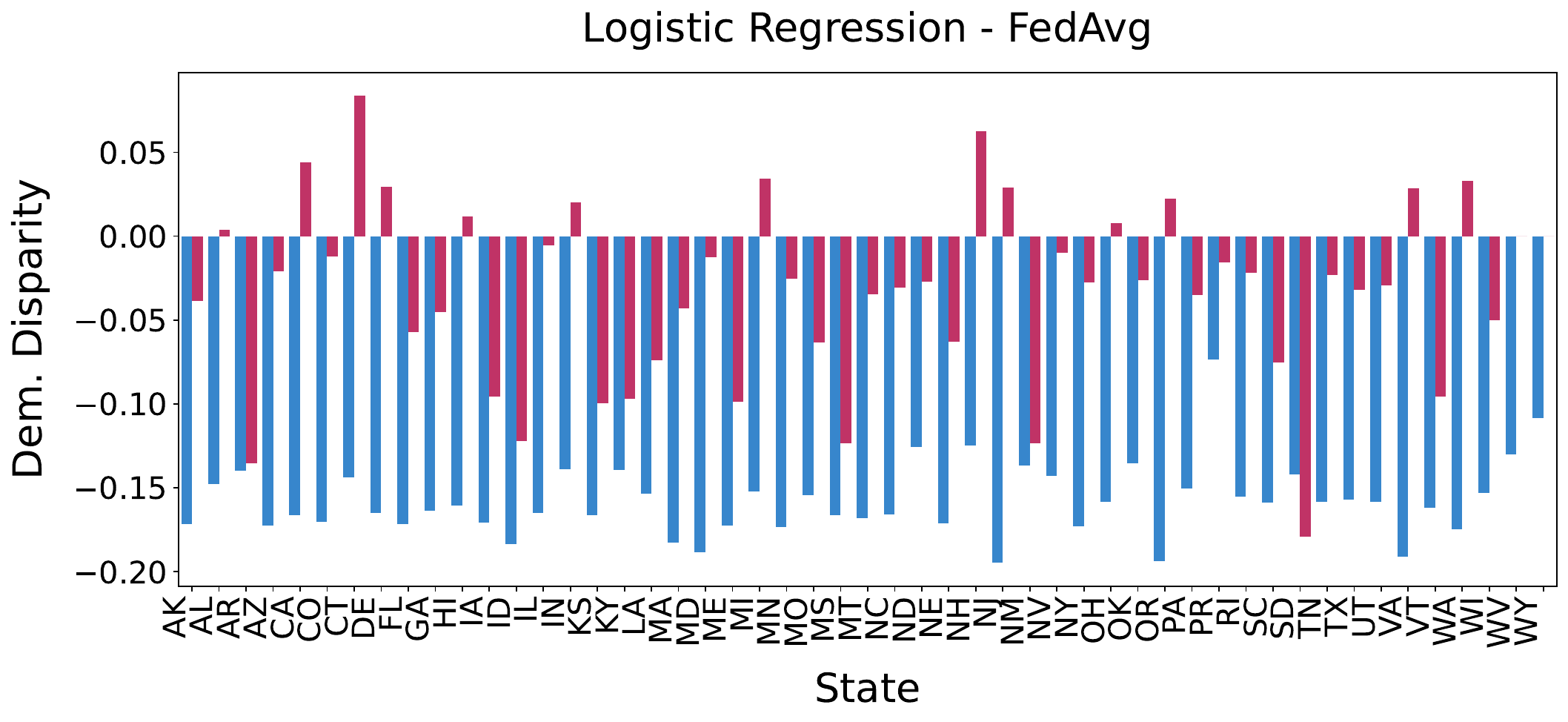}\label{fig:appendix_bar_plot_cross_silo_lr_value}}
\subfloat[\textbf{Value-silo} LR vs. PUFFLE]{\includegraphics[width=0.49\textwidth]{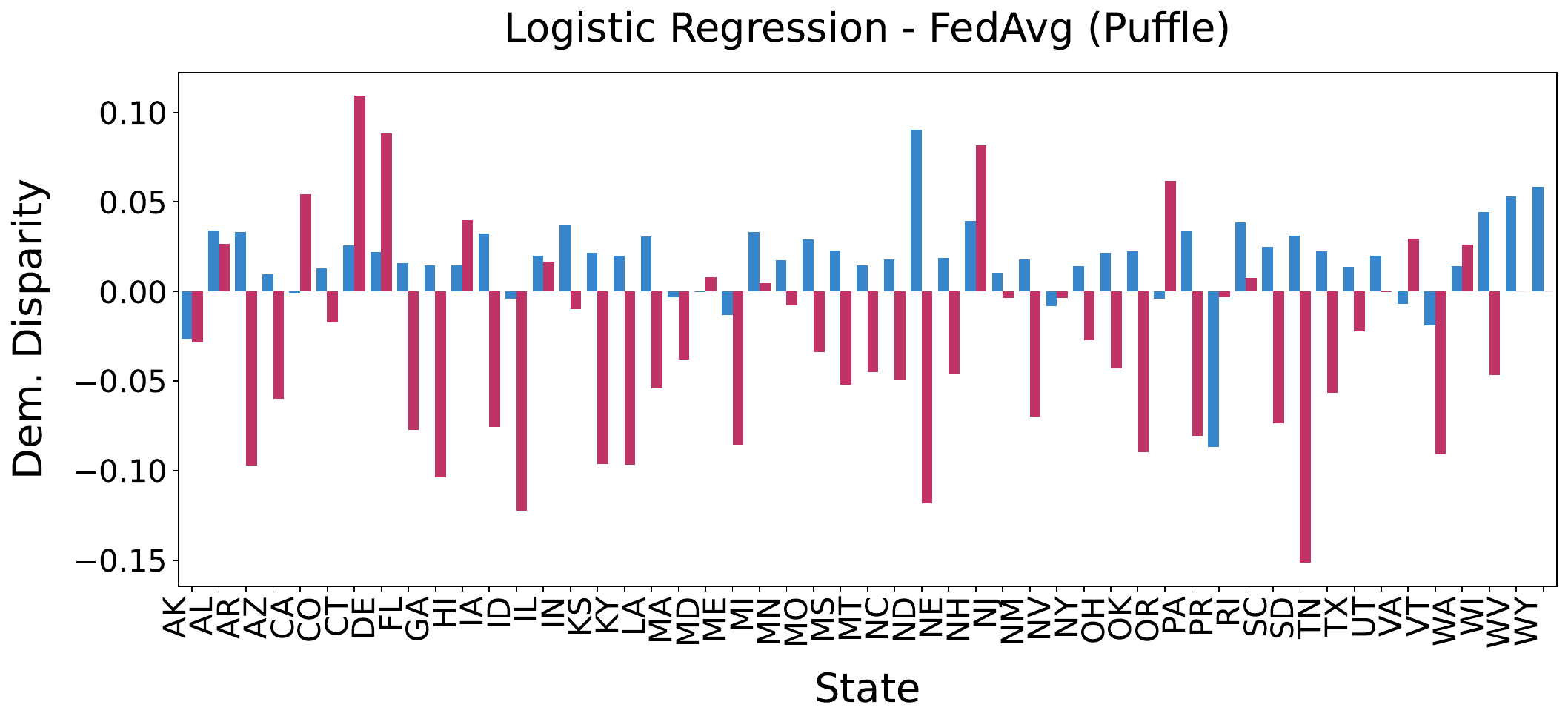}\label{fig:bar_plot_cross_device_xgb_attribute}}\\
\subfloat[\textbf{Value-silo} LR vs.  Reweighing]{\includegraphics[width=0.49\textwidth]{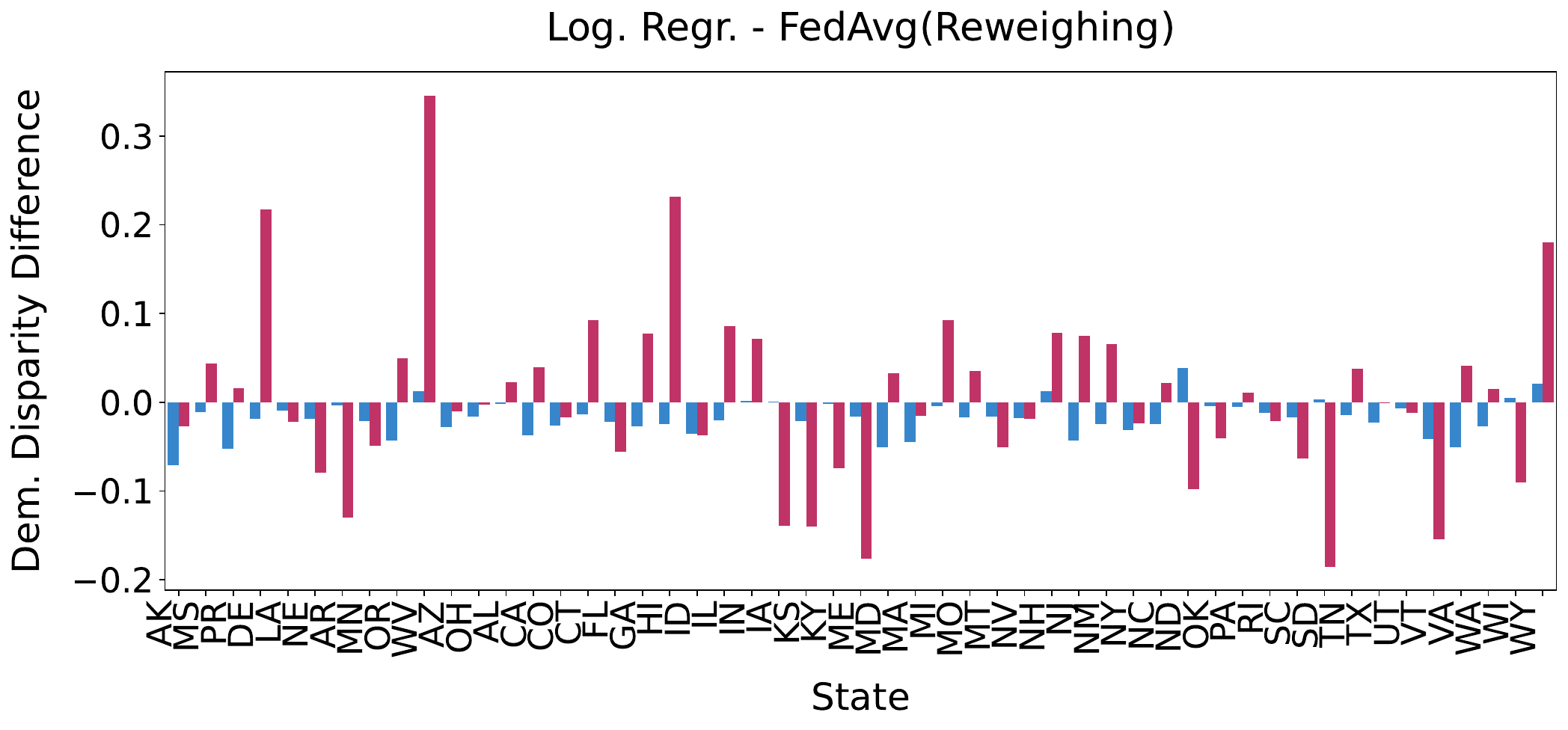}\label{fig:bar_plot_cross_device_xgb_attribute_reweighing}}\\
\centering\includegraphics[width=0.4\textwidth]{images/legend/bar_plot_differences_legend.pdf}
  \caption{Individual, per attribute, bias differences in Demographic Disparity between the local Logistic Regression (LR) models versus the FedAvg model, the PUFFLE model, and the Reweighing model for the value-silo dataset.}
  \label{fig:bar_value_silo_lr}
\end{figure}

\end{document}